\documentclass[journal,twoside]{IEEEtran}
\usepackage{amsmath,amsfonts}
\usepackage{algorithmic}
\usepackage{algorithm}
\usepackage{array}
\usepackage[caption=false,font=normalsize,labelfont=sf,textfont=sf]{subfig}
\usepackage{textcomp}
\usepackage{stfloats}
\usepackage{url}
\usepackage{verbatim}
\usepackage{graphicx}
\usepackage{cite}
\usepackage{booktabs}
\usepackage{multirow}
\usepackage{xcolor}

\usepackage[colorlinks,linkcolor=blue,anchorcolor=blue,citecolor=blue]{hyperref}
\newcommand{\orcid}[1]{\href{https://orcid.org/#1}{\includegraphics[width=10pt]{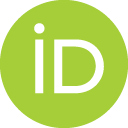}}}

\begin{document}

\title{Super4DR: 4D Radar-centric Self-supervised Odometry and Gaussian-based Map Optimization}

\author{%
Zhiheng Li\,\orcid{0000-0002-1477-2066},
Weihua Wang\,\orcid{0009-0002-7462-8290},
Qiang Shen\,\orcid{0009-0000-9582-6275},
Yichen Zhao\,\orcid{0009-0004-6530-9057},
and Zheng Fang*\,\orcid{0000-0003-3887-3141},~\IEEEmembership{Member,~IEEE}%
\thanks{This work was supported in part by the National Natural Science Foundation of China under Grant 62073066 and in part by the 111 Project under Grant B16009. \textit{(Corresponding author: Zheng Fang.)}}
\thanks{The authors are with the Faculty of Robot Science and Engineering, Northeastern University, Shenyang 110819, China. Zhiheng Li and Zheng Fang are also with the National Frontiers Science Center for Industrial Intelligence and Systems Optimization, Shenyang 110819, China. Zheng Fang is further affiliated with the Key Laboratory of Data Analytics and Optimization for Smart Industry (Northeastern University), Ministry of Education, Shenyang 110819, China (e-mail: fangzheng@mail.neu.edu.cn).}
}

\markboth{IEEE Transactions on Robotics. Preprint Version. Accepted July, 2026}{Li \MakeLowercase{\textit{et al.}}: Super4DR: 4D Radar-centric Self-supervised Odometry and Gaussian-based Map Optimization}

\maketitle

\begin{abstract}
Conventional odometry and mapping methods using visual or LiDAR data often struggle under poor illumination and adverse weather conditions. Although 4D radar is suited for such environments, its sparse and noisy point clouds hinder accurate odometry estimation, while the radar maps suffer from obscure and incomplete structures. Thus, we propose Super4DR, a 4D radar-centric framework for learning-based odometry estimation and gaussian-based map optimization. First, we design a cluster-aware odometry network that incorporates object-level cues from the clustered radar points for inter-frame matching, alongside a hierarchical self-supervision mechanism to overcome outliers through spatio-temporal consistency, knowledge transfer, and feature contrast. Second, we propose using 3D gaussians as an intermediate representation, coupled with a radar-specific growth strategy, selective separation, and multi-view regularization, to recover blurry map areas and those undetected based on image texture. Experiments show that Super4DR achieves a 67\% performance gain over prior self-supervised methods, nearly matches supervised odometry, and narrows the map quality disparity with LiDAR while enabling multi-modal image rendering.
\end{abstract}


\section{Introduction}
\IEEEPARstart{O}{dometry} and mapping are essential capabilities for mobile robots. Over the past decade, LiDAR- and vision-based methods have made notable progress, enabling accurate pose estimation and scene reconstruction in standard operating conditions.
However, LiDAR measurements can be affected by scattering and attenuation in adverse atmospheric conditions, resulting in fewer valid returns and increased noise. Similarly, conventional visible-light cameras usually capture limited appearance information under poor illumination or dense smoke. Together, these limitations hinder reliable odometry and mapping in such challenging environments.

\begin{figure}[t!]
\centering
\includegraphics[width=\linewidth]{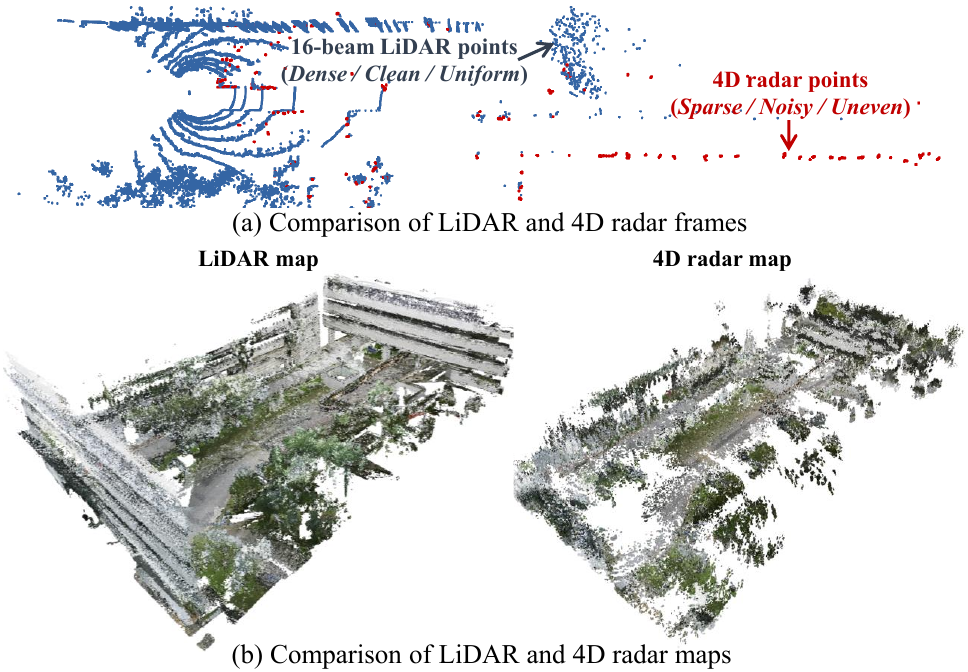}
\vspace{-0.3in}
\caption{Comparison of LiDAR and 4D radar data. Unlike LiDAR, 4D radar points usually suffer from greater sparsity, noise, and uneven density, thereby posing challenges for robust odometry estimation and high-integrity mapping. 
Here, the maps are constructed based on ground-truth poses provided by the NTU4DRadLM~\cite{ntu4dradlm} dataset.
}
\vspace{-0.2in}
\label{fig:Comparison}
\end{figure}

Fortunately, 4D radar has emerged as a promising solution to these issues. Its longer wavelengths provide superior signal penetration and all-weather robustness compared with LiDAR.
Additionally, unlike high-cost scanning radar limited to coarse planar imaging and 3D single-chip automotive radar with low resolution, 4D radar provides point cloud data that can describe the approximate contour of objects. 
Consequently, recent studies have developed geometric and end-to-end frameworks for 4D radar odometry, aiming for reliable pose estimation in challenging weather. 
However, geometry-based algorithms~\cite{4dradarslam,EFEAR-4D,Radar-ICP,Radar4Motion} that rely on dense and precise structural association encounter limitations due to the characteristics of radar points (Fig.~\ref{fig:Comparison}(a)). 
1) \textit{Limited geometric primitives:} The sparser radar data lacks distinct geometric elements (e.g., lines and planes), which act as essential constraints for traditional registration approaches. 
2) \textit{Measurement uncertainty:} More severe positional noise in range and azimuth violates the rigid geometric assumptions of conventional methods. 
In comparison, learning-based methods~\cite{4DRONet,4DRVONet,SelfRONet,RaFlow} are capable of extracting information-rich features from low-quality radar points. 
They further model inter-frame relationships using adaptive matching within high-dimensional feature space to alleviate the impact of noise, thereby pushing the boundaries of 4D radar odometry.

\begin{figure*}[t!]
\centering
\includegraphics[width=\linewidth]{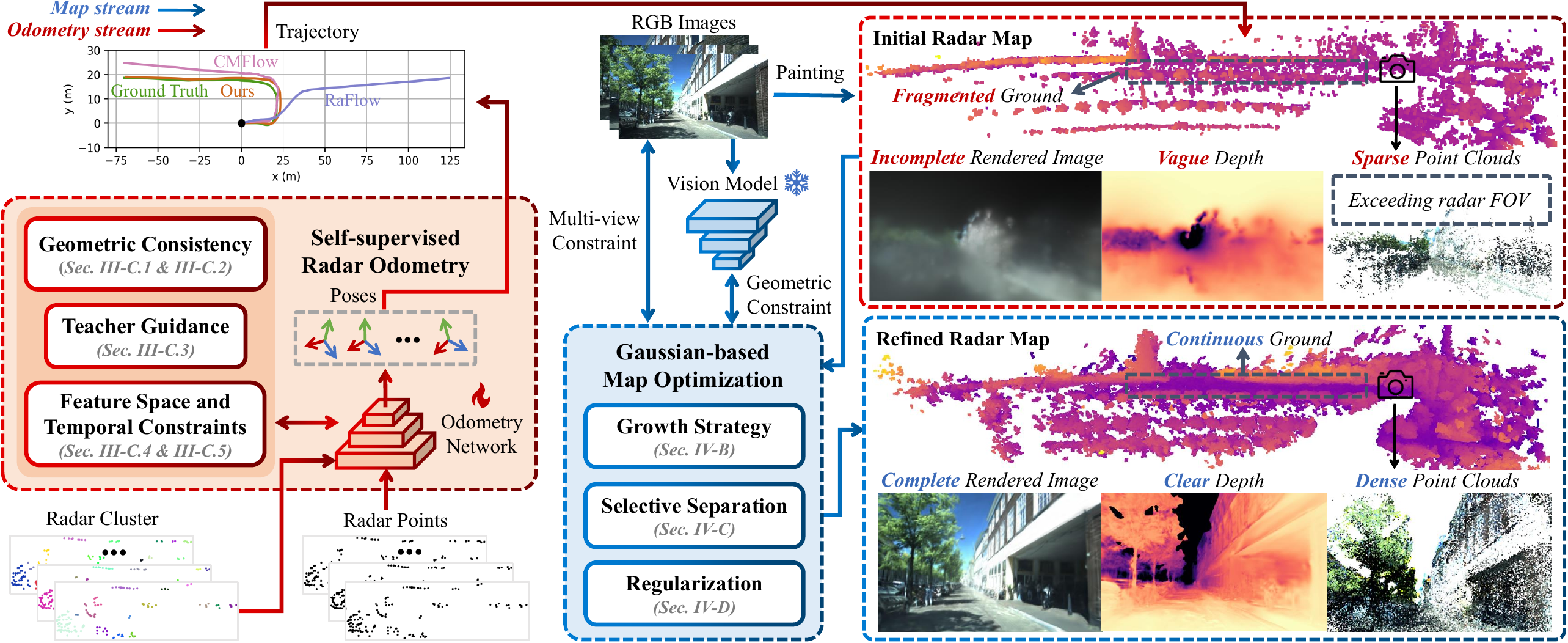}
\vspace{-0.25in}
\caption{Overview of Super4DR framework with end-to-end radar odometry and gaussian-based map optimization. 
First, a cluster-aware network, trained with multi-level self-supervised losses, processes radar points along with cluster grouping to estimate poses. 
Afterwards, an initial map with incomplete structures is constructed based on the trajectory and painted through image pixels. 
We further propose a map optimizer adopting gaussians as an intermediate representation, with a radar-specific growth strategy, selective separation and multi-view regularization for complete map reconstruction and detailed image rendering.
}
\vspace{-0.15in}
\label{fig:framework}
\end{figure*}

However, supervised learning depends on expensive ground truths that are difficult to obtain in harsh conditions. 
Although self-supervised algorithms~\cite{SelfRONet,RaFlow} are label-free, their accuracy lags notably behind supervised approaches due to over-reliance on rigid geometric constraints that are easily disrupted by radar noise and measurement errors.
Meanwhile, they inherit point-level matching originally designed for uniform LiDAR points, which may be suboptimal for the blocky distribution of radar points and fail to exploit latent object cues.
Beyond odometry, the maps built from 4D radar points are too loose and indistinct (Fig.~\ref{fig:Comparison}(b)). 
While utilizing diffusion networks~\cite{RDiffusion, RDiff} is a way for point cloud completion, they struggle to reconstruct objects outside radar's field of view (e.g., inferring a complete tree from a single trunk without rich geometric priors) and face a critical bottleneck in scene generalization.
Furthermore, most existing works focus on enhancing radar frames independently, leaving the optimization of the entire map as a challenging and open topic.
Thus, in this article, we propose a 4D radar-centric framework called \textit{Super4DR}, as illustrated in Fig.~\ref{fig:framework}, to address the radar limitations in accurate pose estimation and complete map reconstruction.

For radar odometry, building upon radar properties, we propose a cluster-aware network with multi-level self-supervision, moving beyond sole reliance on geometric constraints. 
Specifically, we use clustering information derived from point distribution to enable the network to learn inter-frame relationships at the object level while suppressing the point-level noise. 
We further introduce a \textit{cluster-weighted distance} loss to supervise geometric alignment, with particular emphasis on large, stable point clusters.
Next, we present a \textit{column occupancy} loss that represents points as grid occupation, allowing for tolerance of noisy positions and ensuring large-scale consistency through column-wise similarity.
Beyond rigid spatial consistency rules, a geometry-based approach serves as a \textit{teacher} to produce soft pose labels based on a secondary refinement and selection mechanism, distilling meaningful knowledge into the network.
Additionally, we adopt a \textit{feature contrast} loss to maximize the feature difference of non-corresponding point pairs for reliable matching and design a constant-acceleration \textit{motion model} for trajectory smoothing.

For map optimization, encouraged by 3D gaussian splatting (3DGS~\cite{3dgs}), we believe that converting the 4D radar map into gaussian representation and optimizing it could enhance map quality, grounded in several capabilities of 3DGS framework. 
\textbf{First}, benefiting from splitting and cloning mechanisms, radar gaussians can multiply adaptively based on gradients to occupy structural voids, while their positions are adjusted to generate dense geometry under image supervision.
\textbf{Second}, because of the image's larger vertical FOV, gaussians are forced to expand and recover undetected regions of radar to guarantee that their rendered output aligns with image pixels correctly, thereby improving map integrity.
\textbf{Third}, with the help of cross-modal and full-map optimization without neural network, it can avoid generalization problems and ensure greater global consistency than frame-wise point refinement.
Once optimization finishes, refined gaussians can be reverted into points according to their centroids, producing a complete and structurally coherent map.

Nevertheless, directly using 3DGS~\cite{3dgs} is ``insufficient" as it prioritizes rendering quality over geometric accuracy, while gaussians initialized from radar maps with uneven point distribution and noise cause challenges for subsequent optimization.
Thus, we propose a \textit{growth strategy} that first creates synthetic gaussians using depth priors derived from a visual foundation model to solve the ground fragmentation issue unique to radar maps.
Unlike the densification manner that only meets image rendering requirements in most GS-based methods, we adopt a geometry-aware strategy that splits and interpolates gaussians based on local geometry to generate denser structures.
Besides, \textit{selective separation} decouples sky gaussians by maintaining a mask to decrease floaters and saves valuable large gaussians from periodic pruning for subsequent completion.
Rather than optimizing gaussians based on a single view with ambiguous spatial information one by one, we introduce a \textit{multi-view regularization} that leverages overlapping regions to jointly refine noisy radar gaussians into deterministic states. 
Furthermore, geometric priors from visual foundation models are utilized as extra supervision to improve the geometric accuracy of maps.

In summary, the contributions of this paper are as follows:
\begin{itemize}
    \item We present Super4DR as a two-stage framework that targets radar odometry and map optimization, 
    enabling pose prediction from low-quality radar points and structural completion of initially sparse and indistinct radar maps. 
    \item We propose a cluster-aware odometry network trained by a multi-level self-supervision strategy, including cluster-weighted distance, column occupancy, teacher guidance, and the constraints on feature-space and temporal aspects, to mitigate the impact of outliers on the learning process.
    \item We innovatively employ 3D gaussian representation as an intermediary to enhance map quality, developing gaussian growth strategy with depth-guided ground completion and geometry-aware densification, complemented by selective separation and multi-view overlap regularization, thereby recovering blurry structures and radar-undetected areas. 
    \item Experiments on public and self-collected datasets indicate that our radar odometry achieves state-of-the-art results among self-supervised algorithms while remarkably closing the gap to recent supervised methods. Super4DR also exceeds previous GS-based works in radar map and rendered image quality, validating that gaussian splatting can be a viable new approach for 4D radar map optimization. 
    \item We additionally explore the potential of 4D radar-thermal gaussian splatting under challenging conditions and show the effectiveness of our algorithm across diverse imaging modalities, including RGB and thermal cameras. To facilitate research in relevant domains, we will release our code and a multi-sensor dataset collected from handheld and mobile robotic platforms at \url{https://github.com/NEU-REAL/Super4DR}.
\end{itemize}

\section{Related Work}
\subsection{LiDAR-based Odometry}
The purpose of odometry is to calculate pose transformation between adjacent frames. 
As a pioneering method, the iterative closest point (ICP)~\cite{P2P-ICP} is widely adopted for point cloud registration by minimizing point-to-point distance, but it exhibits sensitivity to noise. 
Its variants, like point-to-line and point-to-plane ICP, exploit local geometry to enhance robustness, while GICP~\cite{GICP} merges their advantages.
KISS-ICP~\cite{Kiss-icp} unleashes the potential of point-to-point matching by employing adaptive thresholds for data association and outlier rejection. 
NDT~\cite{NDT} forms a voxel-based probability distribution from point clouds, bypassing point correspondences.
For efficiency, LOAM~\cite{loam} extracts edge and plane features to perform matching. Building on it, Lego-LOAM~\cite{lego-loam} incorporates ground segmentation for noise filtering, then F-LOAM~\cite{F-loam} reduces computation burden through non-iterative two-step compensation.

With the rise of deep learning, many studies are dedicated to end-to-end odometry methods, achieving competitive results. 
PWCLO-Net~\cite{Pwclo} and EfficientLO~\cite{EfficientLO} introduce a coarse-to-fine framework to iteratively match point clouds and leverage masks to suppress the effect of dynamic objects. TransLO~\cite{Translo} employs window-based attention to extract global features for reliable registration, and DSLO~\cite{DSLO} utilizes historical features as hidden states to raise prediction continuity.
While effective, these supervised learning methods require costly labels. Thus, DeLORA~\cite{DeLORA} uses a KD-Tree to search for target pairs based on predicted pose and adopts point-to-plane and plane-to-plane error metrics to optimize model. HPPLO-Net~\cite{HPPLO-Net} then boosts network robustness using hierarchical point-to-plane distances as supervision.
RSLO~\cite{RSLO} partitions point clouds into multiple sub-regions, estimates ego-motion by regional voting, and uses uncertainty-aware geometric loss for self-supervised training. 
Although these LiDAR-based algorithms work well in normal conditions, their performance markedly drops with point cloud degradation from smoke and adverse weather.

\subsection{Radar-based Odometry}
To achieve odometry estimation in challenging scenes, early works utilize scanning radar as the primary sensor, employing point-to-line metric minimization~\cite{CFEAR}, feature point extraction and matching~\cite{ORORA} or improved NDT with the outlier filter~\cite{SDRP} to estimate ego-motion. However, the blurred 2D imaging of the sensor limits odometry precision and map quality. Its large volume and high cost also hinder the practical application.
By contrast, compact, low-cost and higher-precision 4D radar has attracted research attention. As an initial study, APDGICP~\cite{4dradarslam} extends GICP with probabilistic distributions to improve scan matching. 
EFEAR-4D~\cite{EFEAR-4D} applies a scan-to-map mechanism to form denser spatial correspondences for pose solving.
Besides, a doppler-based ICP~\cite{Radar-ICP} is presented to harness radar velocity information, and RCS-weighted registration~\cite{Radar4Motion} is proposed to handle inherent sparsity and measurement noise in radar data. 

Recently, end-to-end methods have demonstrated strong performance. For example, 4DRONet~\cite{4DRONet}  separately encodes radar information with different properties and constructs two-stage pose refinement. CMFlow~\cite{CMFlow} leverages cross-modal supervision from multi-sensor data to jointly train a network for scene flow and ego-motion estimation. Then, CAO-RONet~\cite{CAO-RONet} uses local completion to supply denser constraints for matching and adopts multi-level resilient registration with feature similarity to suppress noise influence. Besides, 4DRVO-Net~\cite{4DRVONet} presents a multi-modal method that integrates rich image information to assist sparse point matching.
To remove reliance on labels, RaFlow~\cite{RaFlow} computes relative poses from predicted scene flow, subsequently optimizing the model based on chamfer distance. SelfRO~\cite{SelfRONet} further applies spherical reprojection and velocity-aware loss to achieve notable improvement, but there is still a significant gap against supervised methods.
More importantly, most methods focus on the pose accuracy but neglect the map quality. Thus, the constructed radar map remains noisy, vague and incomplete as a result of the low-quality 4D radar points.

\subsection{3D Gaussian Splatting}
Unlike Neural Radiance Fields (NeRFs)~\cite{Nerf} that use MLP to learn implicit neural scenes, 3DGS~\cite{3dgs} represents scenarios as explicit gaussian ellipsoids whose parameters are optimized by differentiable rasterization and image reconstruction losses, enabling faster convergence and rendering speed. 
On this basis, GaussianPro~\cite{Gaussianpro} proposes a propagation operation to generate reliable rendered depths and normals, which are further applied to yield new gaussians to populate under-reconstructed regions and enhance image details.
To promote geometric optimization in sparse views, DNGaussian~\cite{Dngaussian} presents hard and soft depth regularization to drive the movement of gaussians and reshape space structure. 
As 3DGS struggles with precise surface fitting, 2DGS~\cite{2dgs} proposes planar 2D gaussians with view-consistent geometry to exactly model thin surfaces. To reconstruct scenes at real scale, some works~\cite{LiV-GS, HGS-Mapping, Drivinggaussian} integrate LiDAR points to initialize gaussians. LiV-GS~\cite{LiV-GS} uses a conditional constraint to guide gaussian optimization in areas lacking LiDAR depth. HGS-Mapping~\cite{HGS-Mapping} optimizes hybrid gaussians with different classes, while DrivingGaussian~\cite{Drivinggaussian} represents dynamic scenes by decoupling gaussians for static and moving objects. Recently, RadarSplat~\cite{kung2025radarsplat} integrates 3D gaussian splatting with radar noise modeling for scanning radar, enabling radar image rendering and occupancy estimation.
Despite these advances, enhancing incomplete maps built from 4D radar points has received limited attention.
This is a distinct topic because, unlike dense LiDAR point clouds or 2D scanning radar data, 4D radar points are naturally sparse and non-uniformly distributed in 3D space, leaving many spatial regions uninitialized and hindering subsequent optimization owing to the lack of geometric priors.

\begin{figure*}[t!]
\centering
\includegraphics[width=\linewidth]{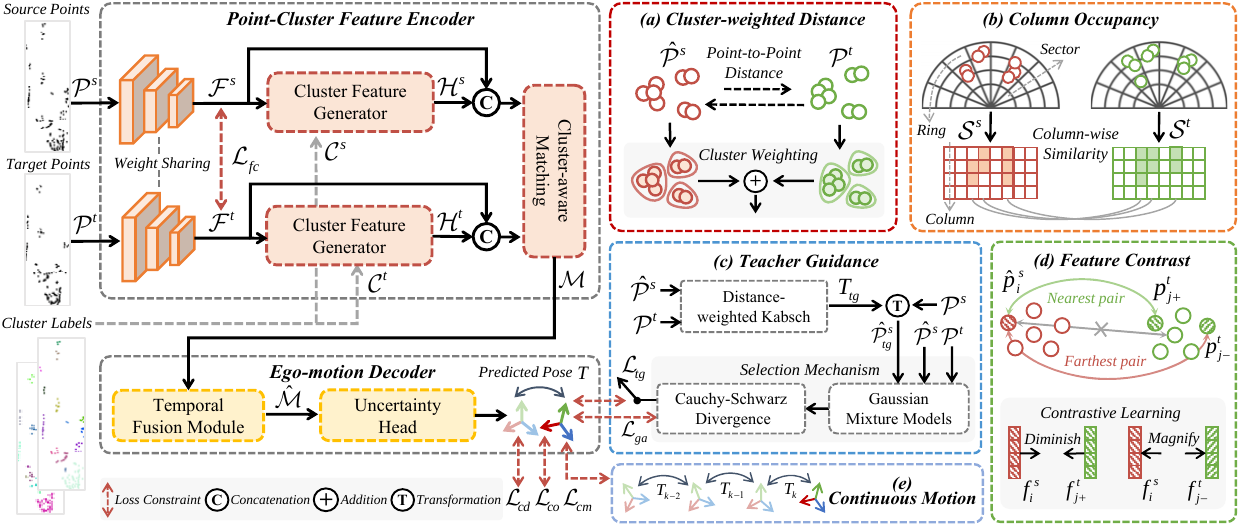}
\vspace{-0.3in}
\caption{Framework of end-to-end radar odometry with multi-level constraints. The odometry network consists of a point-cluster feature encoder and an ego-motion decoder. For self-supervised training, the loss functions first include cluster-weighted distance and column-wise occupancy comparison, which account for the distribution and noise properties of radar points. Network learning is further selectively guided via soft labels generated by a geometry-based algorithm. We also facilitate discriminative feature extraction using feature contrast, while enforcing motion smoothness through a constant-acceleration assumption.}
\vspace{-0.2in}
\label{fig:odom_pip}
\end{figure*} 

\section{Self-supervised Radar Odometry with Multi-level Constraints}
\subsection{Problem Definition}
Odometry estimation aims to calculate pose transformation $T=\{R, t\}$ between consecutive timesteps using sensor data, where $R$ and $t$ mean rotation and translation components. 
The source and target points $\mathcal{P} \in \{\mathcal{P}^s, \mathcal{P}^t\}$ are expressed as $\mathcal{P} = \{p_i = \{x_i, f_i\}\}^{N}_{i=1}$, where $N$ is the number of points, $x \in \mathbb{R}^3$ means 3D coordinates, and $f \in \mathbb{R}^2$ denotes point feature with relative radial velocity (RRV) and radar cross section (RCS).
For end-to-end odometry, a neural network $\Gamma$ is used to encode inter-frame relationship and decode pose $T$. 
Unlike supervised learning using ground-truth pose $(R^{gt}, t^{gt})$ as a constraint to train model parameters $\theta$, i.e., $\min_{\theta} \mathcal{L}(\Gamma_{\theta}(\mathcal{P}^s, \mathcal{P}^t), (R^{gt}, t^{gt}))$, we propose a hierarchical self-supervised loss with geometric ($\mathcal{L}_{cd}$, $\mathcal{L}_{co}$), temporal $\mathcal{L}_{cm}$ and feature-level $\mathcal{L}_{fc}$ terms, along with teacher guidance $\mathcal{L}_{tg}$ and  $\mathcal{L}_{ga}$, to generate comprehensive supervision signals. 
Our optimization objective is as follows:
\begin{equation}
\begin{aligned}
\min_{\theta} ( 
& \mathcal{L}_{cd}(\hat{\mathcal{P}}^s, \mathcal{P}^t) + \mathcal{L}_{co}(\hat{\mathcal{P}}^s, \mathcal{P}^t) + \mathcal{L}_{tg}(T, T_{tg}) \ + \\ 
& \mathcal{L}_{ga}(\hat{\mathcal{P}}^s, \mathcal{P}^t) + \mathcal{L}_{fc}(\mathcal{F}^s, \mathcal{F}^t) + \mathcal{L}_{cm}(T_{k-2:k})), 
\end{aligned}
\end{equation}
where $\hat{\mathcal{P}}^s = \mathcal{T}(\mathcal{P}^s, T)$, and $\mathcal{T}$ is the transformation operation. $T_{tg}$ and $T_{k-2:k}$ are soft labels and sequential predicted poses.

\subsection{Cluster-aware Network Architecture}
Even though radar data is much sparser than LiDAR points, it possesses a block-like spatial distribution, making it easier to segment into distinct objects or regions based on geometric distances. 
These grouping labels then enable the integration of instance features for matching, which reduces search space for potential matches and enhances outlier rejection by enforcing consistency within similar clusters. 
Motivated by this insight, we design a cluster-aware network structured as follows:
\subsubsection{Point-cluster Feature Encoder}
In Fig.~\ref{fig:odom_pip}, we first employ DBSCAN~\cite{DBSCAN} to cluster raw points $\mathcal{P} \in \{\mathcal{P}^s, \mathcal{P}^t\}$ and assign cluster labels. Then, we utilize farthest point sampling (FPS) to unify two frames into the same number $N$ of points and obtain the corresponding labels $\mathcal{C}^s, \mathcal{C}^t \in \mathbb{R}^{N}$.
The set convolution layer~\cite{Flownet3d} is further used to extract multi-scale features from downsampled points and combine them into $\mathcal{F}^s, \mathcal{F}^t \in \mathbb{R}^{N \times D}$ along channel dimension.

\begin{figure}[t!]
\centering
\includegraphics[width=\linewidth]{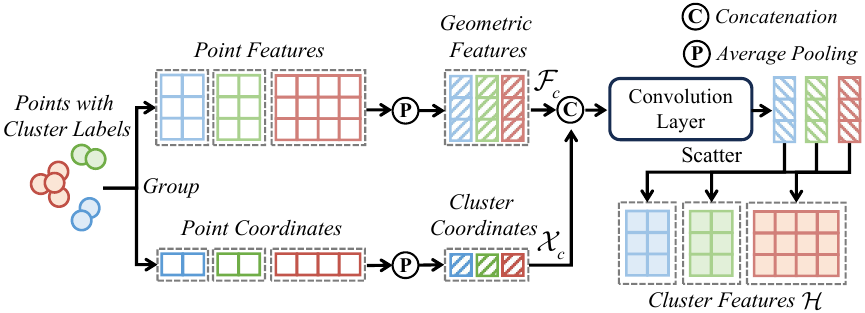}
\vspace{-0.25in}
\caption{Details of cluster feature generator. It takes point features with cluster labels as input to produce per-point cluster features.}
\vspace{-0.2in}
\label{fig:Cluster}
\end{figure} 

To acquire cluster information, we propose a cluster feature generator that exploits cluster labels to integrate point features and coordinates belonging to the same group to produce local geometric features $\mathcal{F}_c$ and centroids $\mathcal{X}_c$ of clusters. Both are concatenated and processed through two convolutional layers to obtain cluster features, which are then redistributed to their corresponding points, generating per-point cluster features $\mathcal{H}$. 
This process is illustrated in Fig.~\ref{fig:Cluster} and can be formulated as:
\begin{equation}
    \mathcal{F}_c = \frac{1}{| \mathcal{C}_c |} \sum_{i \in \mathcal{C}_c} \mathcal{F}_i, \ \mathcal{X}_c=\frac{1}{| \mathcal{C}_c |} \sum_{i \in \mathcal{C}_c} \mathcal{X}_i, \ \forall c \in \{1,\dots,L\}
\end{equation}
\begin{equation}
    \mathcal{H}_i = \text{Conv}(\mathcal{F}_{c(i)} \oplus \mathcal{X}_{c(i)}), \ \forall i \in \{1,\dots,N\}
\end{equation}
where $\mathcal{C}_c$ represents the set of points assigned to the cluster $c$, and $L$ signifies the total number of clusters, while $\mathcal{F}_i$ and $\mathcal{X}_i$ denote the feature and coordinate of the $i$-th point from $\mathcal{P}$. 
Next, we leverage cost volume layers~\cite{Pointpwc-net} to conduct feature correlation, which compares point pairs and learns the nonlinear relationship by MLP. 
However, rather than relying solely on point features, we employ cluster features $\mathcal{H}$ to implement cluster-aware matching as described in Eq.~\ref{eq:cost}.
Since cluster features of the same object are similar in adjacent frames, this cue strengthens recognition of specific-object relationships and suppresses mismatches with irrelevant objects.
We further aggregate cost features in a patch-to-patch manner~\cite{Pointpwc-net}, getting an inter-frame motion feature $\mathcal{M}$.
\begin{equation}
\text{Cost}(p_i^s, p_j^t) = \text{MLP}(f_i^s \oplus f_j^t \oplus h_i^s \oplus h_j^t \oplus (x_i^s  \mathbin{-} x_j^t))
\label{eq:cost}
\end{equation}

\subsubsection{Ego-motion Decoder}
To ensure the continuity of motion prediction, we first employ max-pooling to motion feature $\mathcal{M}$, obtaining a global motion representation $\mathcal{M}_g \in \mathbb{R}^{D}$.
It is then merged with historical global feature using GRU~\cite{gru}, thereby constraining the current prediction using temporal consistency from adjacent states implicitly. The updated global feature is further combined with $\mathcal{M}$ to form $\hat{\mathcal{M}} \in \mathbb{R}^{N \times 2D}$. 

Unlike previous methods~\cite{SelfRONet,CAO-RONet,4DRONet} that directly average motion features for pose prediction, we argue that focusing on critical points rather than all is necessary, as noisy points and dynamic objects cause errors in results. Thus, we calculate the confidence score of each point based on its geometric feature $\mathcal{F}$ and motion feature $\mathcal{M}$, which implies density condition and movement relative to neighbours. 
The score-weighted feature is then used to estimate rotation $R$ and translation $t$ as follows:  
\begin{equation}
   W = \text{SoftMax}(\text{Conv}(\mathcal{F} \oplus \mathcal{M}))
\end{equation}
\begin{equation}
    R = \text{Conv}\sum_{i=1}^N (W_i \cdot \hat{\mathcal{M}}_i), \ t = \text{Conv}\sum_{i=1}^N (W_i \cdot \hat{\mathcal{M}}_i)
\end{equation}

\subsection{Multi-level Self-supervised Signals}
\subsubsection{Cluster-weighted Distance Loss}  
Since radar points lack distinct structures (e.g., lines and planes), employing point relationship constraints for network learning is a basic approach.
Specifically, the source points $\mathcal{P}^s$ are transformed by $T$ to $\hat{\mathcal{P}}^s$, where each $\hat{p}_i^s$ is paired with its nearest neighbor $p_j^t \in \mathcal{P}^t$, and their distances serve as supervision.
To mitigate the impact of incorrect pairs caused by isolated and noisy points, we discard outliers using local density and define inter-frame distance as:
\begin{equation}
d_i^{s\rightarrow t} = \mathbb{I}(\rho(\hat{p}_i^s) > \delta) \text{max}(\min_{p_j^t \in \mathcal{P}^t} \|\hat{p}_i^s - p_j^t\|_2^2 - \epsilon, 0)
\end{equation}
where $\rho(\cdot)$ is a density function, while $\delta$ and $\epsilon$ mean the density and distance thresholds. Although averaging all distances $d_i$ is convenient, this overlooks the contribution of different regions.
For example, larger point clusters typically exhibit more stable and complete geometries across frames, enabling more precise consistency assessment than the fragmented clusters. Thus, we enhance point distances in large clusters and suppress those in small clusters through cluster labels $\mathcal{C}_c$ as weighting factors:
\begin{equation}
\mathcal{L}_{cd}^{s\rightarrow t} = \sum_{c=0}^{L-1} (\frac{n_c}{N} \cdot \frac{1}{n_c} \sum_{i \in \mathcal{C}_c} d_i^{s\rightarrow t}), \ \mathcal{L}_{cd} = \mathcal{L}_{cd}^{s\rightarrow t} + \mathcal{L}_{cd}^{t\rightarrow s}
\end{equation}
where $n_c$ indicates the number of points in the cluster labeled $c$. The bidirectional matching strategy is also used to eliminate ambiguity in unidirectional matching, as displayed in Fig.~\ref{fig:odom_pip}(a).

\subsubsection{Column Occupancy Loss}
While the point-to-point constraint facilitates network convergence, it is inherently difficult for matched pairs to correspond to identical physical locations in 3D space, resulting in geometric inconsistency. Meanwhile, distance uncertainty of point pairs caused by inaccurate radar measurement amplifies this issue.
Thus, we propose a column occupancy loss in Fig.~\ref{fig:odom_pip}(b), whose core idea is to transform point clouds into voxel occupancy representation with a fixed resolution that has tolerance for position noise and fuzzy space correspondences. 
Subsequently, column-wise comparison enables a more reliable consistency evaluation by operating over a larger range, which effectively reduces the influence of local noise that often plagues small-scale point-pair comparisons.

Specifically, given a point cloud $\mathcal{P} \in \{\hat{\mathcal{P}}^s, \mathcal{P}^t\}$, we first map each point $p_i = (x_i, y_i, z_i)$ into a 2D polar representation by computing its azimuth angle $\theta_i$ and radial distance $r_i$:
\begin{equation}
\theta_i = 90 + \text{atan2}(y_i, x_i) \cdot \frac{180}{\pi}, \ r_i = \sqrt{x_i^2 + y_i^2}
\end{equation}
The angle $\theta_i$ and distance $r_i$ are discretized into sectors and rings with the resolutions controlled by $\Delta_\theta = \frac{360^\circ}{N_s}$ and $\Delta_r = \frac{r_{max}}{N_r}$, where $N_s$ and $N_r$ mean the number of sectors and rings.
We then generate binary occupancy matrices $\mathcal{S} \in \{0, 1\}^{N_r \times N_s}$ for $\hat{\mathcal{P}}^s$ and $\mathcal{P}^t$, respectively. The element $\mathcal{S}(k,l)$ indicates the presence of at least one point in the $k$-th ring and $l$-th sector:
\begin{equation}
\mathcal{S}(k,l) = \mathbb{I} \left( 
    \exists p_i \text{ s.t. } 
    \left\lfloor \frac{r_i}{\Delta_r} \right\rfloor = k 
    \text{ and } 
    \left\lfloor \frac{\theta_i}{\Delta_\theta} \right\rfloor = l 
\right)
\end{equation}
For the occupancy matrices from two frames, we calculate the cosine similarity between corresponding columns $\mathcal{S}^s[:,l]$ and $\mathcal{S}^t[:,l]$ to signify structural differences within a large receptive field. Columns where all elements are equal to 0 due to point sparsity and limited FOV are removed. 
The column occupancy loss is expressed as the average dissimilarity across all valid columns in Eq.~\ref{eq:column occupancy loss}, where $\mathbb{L}$ is the non-empty column indices.
\begin{equation}
\mathcal{L}_{co} = 1 - \frac{1}{|\mathbb{L}|} \sum_{l \in \mathbb{L}} \frac{\mathcal{S}^s[:,l] \cdot \mathcal{S}^t[:,l]}{\|\mathcal{S}^s[:,l]\|_2 \cdot \|\mathcal{S}^t[:,l]\|_2}
\label{eq:column occupancy loss}
\end{equation}

\subsubsection{Teacher Guidance Loss} 
Beyond the above losses, which propel model learning by rigid geometric alignment, we think that creating pseudo labels as soft supervision for the network can transfer some meaningful knowledge based on distillation learning theory~\cite{HintonDistilling, Apprentice}. 
Meanwhile, unlike common distillation methods (e.g., \cite{Seed}) that use a large network to generate pseudo labels for training a lightweight model, we propose a teacher guidance loss (Fig.~\ref{fig:odom_pip}(c)) that regards a geometry-based method as the teacher. It uses the network’s predicted pose as an initial guess for the pose refinement.
A selective distillation is then adopted to identify and discard unreliable poses, while using trustworthy ones as soft labels to drive model learning towards better geometric solutions.

Concretely, based on the transformed point cloud $\hat{\mathcal{P}}^s$ by $T$, we apply the distance-weighted Kabsch algorithm to solve the pose adjustment $\Delta T = \{\Delta R, \Delta t\}$ by minimizing the nearest-neighbor distance between $\hat{\mathcal{P}}^s$ and $\mathcal{P}^t$, defined as:
\begin{equation}
d_{\max} = \max_i ( \min_{p_j^t \in \mathcal{P}^t} \|\hat{p}_i^s - p_j^t\|_2), \ w_i = \frac{d_{\max} - d_i}{d_{\max}}
\end{equation}
\begin{equation}
\Delta T = \mathop{\mathrm{argmin}} \sum_{i=1}^{N} w_i \| \Delta R \hat{p}_i^s + \Delta t - \mathrm{NN}(\hat{p}_i^s, \mathcal{P}^t) \|^2_2
\end{equation}
where $w_i$ and $\mathrm{NN}(\cdot)$ are the distance weight and nearest neighbor search, and the updated pose is $T_{tg} = \Delta T \cdot T = \{R_{tg}, t_{tg}\}$.
Despite using $T$ as an initial guess, the optimized pose $T_{tg}$ may fail to improve on $T$ because of the challenge posed by radar point sparsity to the Kabsch algorithm. 
Therefore, a selection mechanism is adopted to map $\mathcal{P}^s$ to $\hat{P}_{tg}^s$ by $T_{tg}$, and to convert $\{\hat{P}^s, \hat{P}_{tg}^s, P^t\}$ into gaussian mixture models (GMMs)~\cite{PDF-Flow} and evaluate the alignment of GMMs through Cauchy-Schwarz (CS) divergence. 
When $T_{tg}$ produces a lower CS divergence between the projected and target points $\mathcal{P}^t$ compared to $T$, it can be regarded as a valid refined pose and applied to constrain the model training as follows:
\begin{equation}
M = \mathbb{I} (g(\mathcal{G}(\hat{\mathcal{P}}^s),\mathcal{G}(\mathcal{P}^t)) > g(\mathcal{G}(\hat{\mathcal{P}}_{tg}^s),\mathcal{G}(\mathcal{P}^t)))
\end{equation}
\begin{equation}
\mathcal{L}_{tg} = M \cdot (\| R - R_{tg} \|_1 + \| t - t_{tg} \|_1)
\end{equation}
where $g$ and $\mathcal{G}$ mean functions of CS divergence and gaussian mixture. As a result, selective guidance avoids forcing model to imitate erroneous poses.
In addition, we use CS divergence of $\mathcal{G}(\hat{\mathcal{P}}^s)$ and $\mathcal{G}(\mathcal{P}^t)$ as a loss term $\mathcal{L}_{ga}$ to drive model learning when there are no reliable soft labels available (i.e. $M=0$).

\begin{figure*}[t!]
\centering
\includegraphics[width=\linewidth]{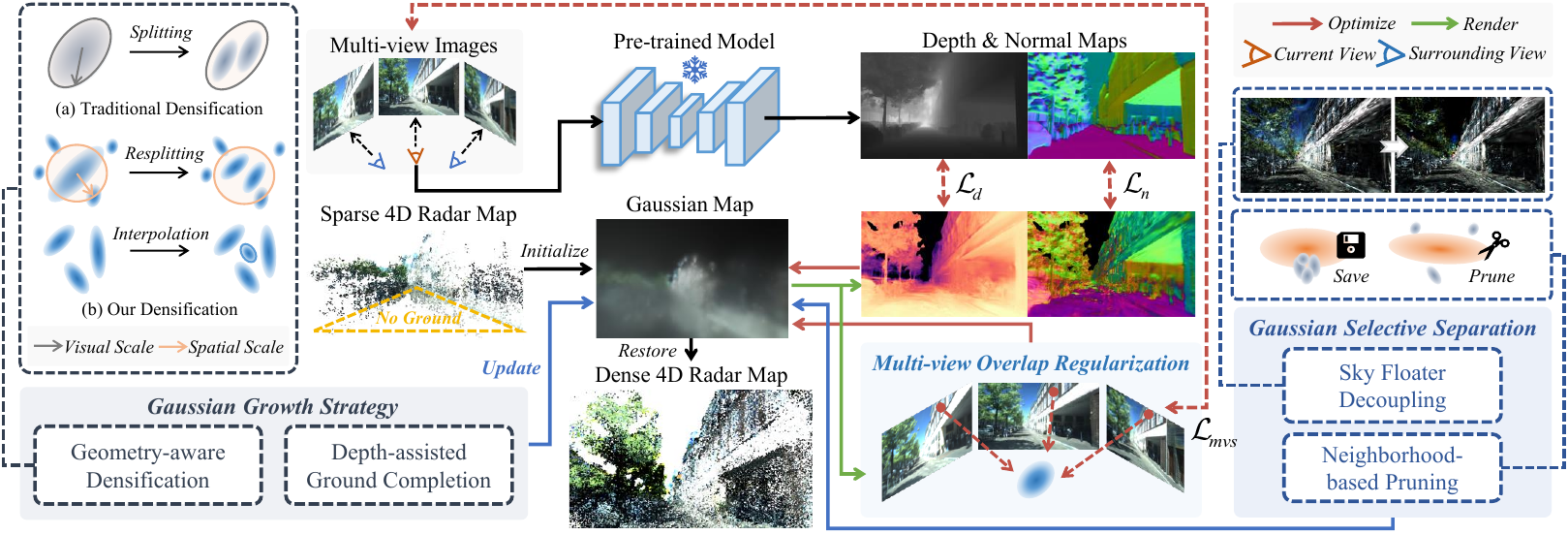}
\vspace{-0.3in}
\caption{Overview of gaussian-based map optimization. We first initialize a sparse 4D radar map as gaussians and perform attribute optimization by gradients derived from multi-view image rendering with the depth and normal maps. During optimization, we adopt ground completion and geometry-aware densification to grow the number of gaussians to reconstruct missing structure, while employing selective separation to decouple sky floaters and avoid excessive pruning. Finally, the optimized gaussians are restored into a dense radar map.}
\vspace{-0.15in}
\label{fig:optimized_map}
\end{figure*} 

\subsubsection{Feature Contrast Loss}
Apart from the result-level constraints, the feature-level regularization is also critical, as pose estimation relies on inter-frame feature correlation. 
Thus, we introduce a feature contrast loss in Fig.~\ref{fig:odom_pip}(d) to enhance feature discrimination among points and help perceive subtle motion between frames. 

We initially consider the backbone's outputs $\mathcal{F}^s$ and $\mathcal{F}^t$ as feature sources.
For each point $\hat{p}_i^s = (\hat{x}_i^s, f_i^s) \in \hat{\mathcal{P}}^s$, we define its nearest point $p_{j+}^t = (x_{j+}^t, f_{j+}^t) \in \mathcal{P}^t$ as a positive sample, whereas the farthest point $p_{j-}^t = (x_{j-}^t, f_{j-}^t)$ is selected as the negative sample instead of neighboring points. 
This is because the neighbors' features are usually similar to $\hat{p}_i^s$, causing weak contrastive signals, which also may generate a false perception that neighbors exhibit larger discrepancies from $\hat{p}_i^s$ than non-neighboring points. Thus, using the farthest point from a global perspective as a negative sample is more reasonable. Next, we use information noise contrastive estimation as a loss function:
\begin{equation}
j^+ = \underset{j \in [1,N]}{\text{argmin}} \| \hat{x}_i^s - x_j^t \|_2, \ j^- = \underset{j \in [1,N]}{\text{argmax}} \| \hat{x}_i^s - x_j^t \|_2
\end{equation}
\begin{equation}
\mathcal{L}_{fc} = -\frac{1}{N} \sum_{i=1}^{N} \log \frac{e^{s(f_i, f_{j^+})/\tau}}{e^{s(f_i, f_{j^+})/\tau} + e^{s(f_i, f_{j^-})/\tau}}
\end{equation}
where $j^+$ and $j^-$ stand for the positive and negative sample indices. $s$ represents the cosine similarity of features between point $\hat{p}_i^s$ and its positive and negative samples. $\tau$ denotes the temperature coefficient.

\subsubsection{Continuous Motion Loss} 
Supervised learning methods such as~\cite{CAO-RONet} implicitly model temporal motion by a network trained with continuous ground truth, which enforces predicted poses to conform to short-term motion patterns.
However, for self-supervised learning, only relying on feature-level temporal fusion is still insufficient owing to missing explicit consecutive constraints.
To address this issue, we apply the assumption of constant acceleration to ensure that the pose predictions maintain physical rationality.
To be specific, we divide the training sequence into $L$-length frame segments and store the historical results in a buffer. For each predicted pose $T_k (2 \leq k < L)$ in the segment, we retrieve poses $\{T_{k-2}, T_{k-1}\}$ from the buffer and optimize the acceleration consistency for prediction stability utilizing $\mathcal{L}_{cm}$:
\begin{equation}
\alpha^R_k = (R_{k} - R_{k-1}), \ \alpha^t_k = (t_{k} - t_{k-1}), \ T_k = [R_k | t_k]
\end{equation}
\begin{equation}
\begin{aligned}
\mathcal{L}_{cm} &= \lambda_{cm}(\max(\| \alpha^R_k - \alpha^R_{k-1} \|_1 - \epsilon^R, 0) \\
&\quad + \max(\| \alpha^t_k - \alpha^{t}_{k-1} \|_1 - \epsilon^t, 0))
\end{aligned}
\end{equation}
where the relative rotation $R_k$ is parameterized in Euler angles. To account for accelerated motion in real-world driving, we apply thresholds $\epsilon^R$ and $\epsilon^t$ to avoid too tight constraints, while enabling a gradual decay of the loss weight $\lambda_{cm}$ for $\mathcal{L}_{cm}$.

\section{Gaussian-based Map Optimization}
\subsection{Preliminary}
The radar map is transformed into a representation signified by anisotropic gaussian ellipsoids $G$. Each gaussian contains its center $\mu \in \mathbb{R}^3$ in the world coordinate, opacity $o \in \mathbb{R}$, scale $S \in \mathbb{R}^3$, rotation $R \in \mathbb{R}^3$, and spherical harmonics $SH \in \mathbb{R}^3$.
Here, $\mu$ corresponds to radar point coordinates. $SH$ is obtained by mapping the RGB pixel values. $S$ is initialized by distances to neighbors. The covariance matrix $\Sigma$ is calculated to define the gaussian shape. Thus, the gaussian distribution is given by:
\begin{equation}
\Sigma = RSS^{T}R^{T}, \ G(x) = e ^ {\left( - \frac{1}{2} (x - \mu)^T \Sigma^{-1} (x - \mu) \right)}
\end{equation}
Afterwards, given the transformation matrix $T_W^C=\{R_W^C,t_W^C\}$ between the camera and world coordinate systems, 3D gaussian ellipsoids $(\mu, \Sigma)$ in the world frame are projected onto the image plane as 2D gaussian $(\mu', \Sigma')$ by the following equation:
\begin{equation}
\mu'=\pi(T_W^C \cdot \mu), \Sigma'=J R_W^C \Sigma (R_W^C)^T (J)^T
\end{equation}
where $R_W^C$ and $t_W^C$ are the rotation and translation components. $J \in \mathbb{R}^{2 \times 3}$ is the Jacobian of the affine approximation of $T_W^C$. $\pi$ means the 3D-to-2D projection operation. The influence of 2D gaussian on pixel $\rho=[u,v]^T$ is determined by weight $\alpha$:
\begin{equation}
\alpha = o \cdot e^{(- \frac{1}{2} (\mu' - \rho)^T {\Sigma'}^{-1} (\mu' - \rho))}
\end{equation}
Gaussians are then sorted by depth and rendered by volumetric $\alpha$-blending to calculate the color of pixel $\rho$, as expressed by:
\begin{equation}
C(\rho) = \sum_{i=1}^n c_i \alpha_i \prod_{j=1}^{i-1} (1 - \alpha_j)
\end{equation}
where $c$ denotes the RGB color from spherical harmonics $SH$.

\subsection{Gaussian Growth Strategy}
\subsubsection{Depth-assisted Ground Completion}
\begin{figure}[t!]
\centering
\includegraphics[width=\linewidth]{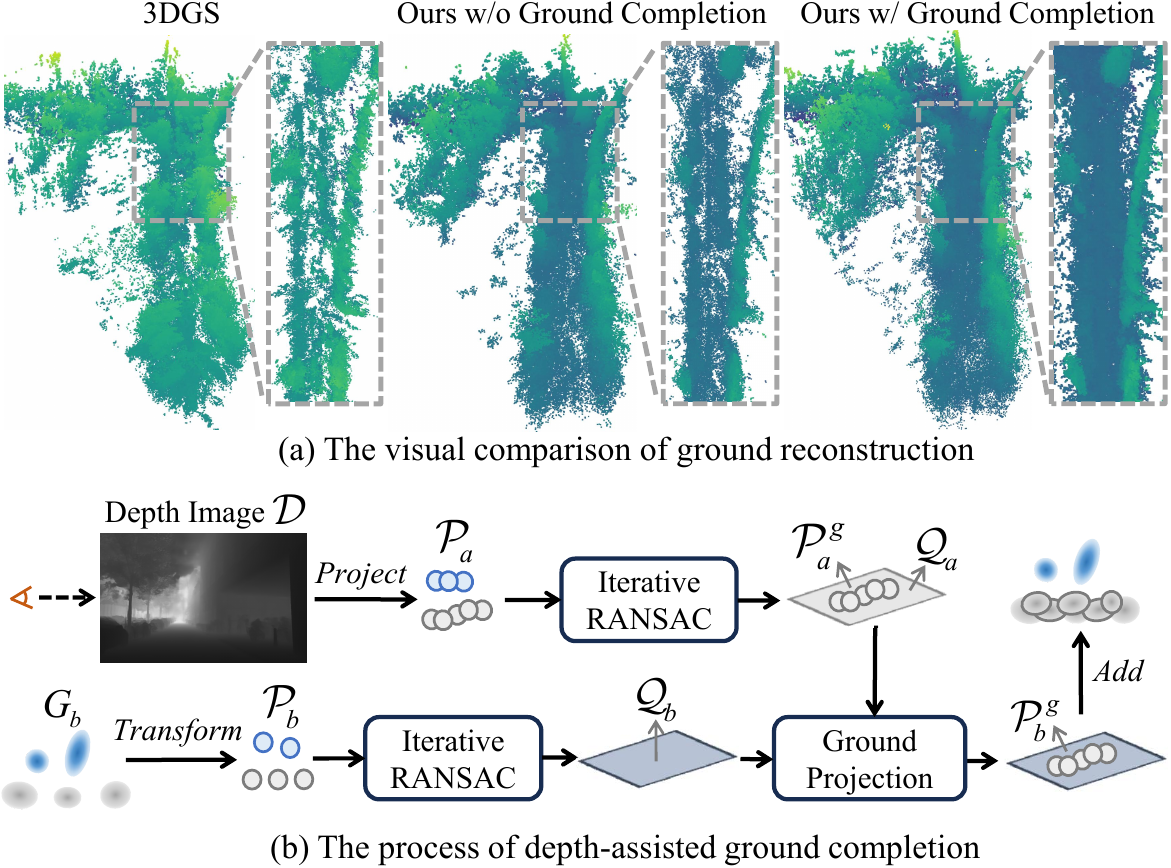}
\vspace{-0.2in}
\caption{The detail of depth-assisted ground completion and visual comparison.}
\vspace{-0.20in}
\label{fig:Ground}
\end{figure}

Caused by specular reflection, the original radar map typically has fewer points on the ground compared to other areas (Fig.~\ref{fig:optimized_map}), posing a challenge for ground reconstruction with poor gaussian initialization.
At the same time, standard gaussian densification in 3DGS\cite{3dgs} only generates a few ground gaussians for low-texture ground rendering in the image, leading to discontinuous ground in the optimized map (Fig.~\ref{fig:Ground}(a)).
Thus, we propose a depth-assisted ground completion mechanism as described in Fig.~\ref{fig:Ground}(b), which integrates synthetic ground gaussians into optimization process through the depth estimation of the visual foundation model.
The positions of these gaussians are adjusted through gradients from the differentiable rasterization to form a realistic ground in the final radar map.
For optimization within a specific view, the ground completion consists of three steps:

\textit{a) Virtual Points Generation}: DepthAnythingV2~\cite{Depthanythingv2} is first utilized to estimate a metric depth map $\mathcal{D}$ of image $\mathcal{I}$ from the current view, which is further mapped to the world coordinate by a transformation matrix $T_W^C$, generating virtual points $\mathcal{P}_a$.

\textit{b) Ground Fitting}: 
We transform gaussians $G_b \subset G$ within the current view into points $\mathcal{P}_b$ according to their center $\mu_b$. 
Owing to the sparsity of ground points, we design an iterative RANSAC with normal-based filtering to fit an initial ground plane to $\mathcal{P}_b$. 
If the plane's normal departs from vertical, nearby outlier points are removed, and the plane is refitted until the normal approaches vertical direction. 
Afterwards, the resulting ground plane $\mathcal{Q}_b=[n_b,d_b]$ is determined, where $n_b$ and $d_b$ are the normal vector and distance from the plane to the origin. 
For the virtual points $\mathcal{P}_a$, we apply the same approach to calculate $\mathcal{Q}_a=[n_a,d_a]$ and nearby ground points $\mathcal{P}_a^g$.

\textit{c) Ground Projection}: We obtain the rotation matrix $R$ and translation $t$ between planes via Rodrigues' rotation formula:
\begin{equation}
v = \hat{n}_a \times \hat{n}_b, \ \hat{n}_a = \frac{n_a}{\| n_a \|_2}, \ \hat{n}_b = \frac{n_b}{\| n_b \|_2}
\end{equation}
\begin{equation}
R = I + [v]_\times + [v]_\times^2 \frac{1 - \hat{n}_a \cdot \hat{n}_b}{\|v\|^2_2}, \ t = (d_a - d_b) \cdot \hat{n}_b
\end{equation}
where $[v]_\times$ means the skew-symmetric matrix of $v$. Then, we incorporate new ground gaussians that are initialized from the projected points $\mathcal{P}_b^g = R \cdot \mathcal{P}_a^g + t$ into the original gaussian set $G$.
Finally, by iteratively applying this process across multiple views, we achieve a dense reconstruction of ground structure.

\subsubsection{Geometry-aware Densification}
\label{sec:geometry-aware densification}
Due to the limited FOV and uneven distribution of radar points, initial gaussians only cover partial image regions, which forces them to be enlarged before being split into tile vacant regions. 
However, the common binary splitting (in Fig.~\ref{fig:optimized_map}(a)) struggles to efficiently deal with large-scale gaussians for rapid densification. 
Meanwhile, post-split positions are sampled from the rendering-optimized visual scales rather than physical spatial scales.
This means the splitting results are tailored for image rendering but may fail to accurately restore the fine 3D structure.
Thus, we present a geometry-aware splitting and interpolation scheme that yields gaussians adhering to local distributions and improves structural integrity.

For efficiency and a trade-off between image rendering and map quality, we extend initial splitting in~\cite{3dgs} with a geometry-aware resplitting stage as illustrated in Fig.~\ref{fig:optimized_map}(b).
This process operates on a subset $G_s$ of $G$, which comprises $I$ gaussians that still retain large scales $S$ after the first splitting.
For each $G_s^i$, we find its nearest gaussian $G_s^j$ and use their distance as a spatial scale $\hat{S}_i$.
Then, we sample the $M$ offsets $\sigma^m_i$ from a normal distribution with a mean of zero and variance of $\hat{S}_i$:
\begin{equation}
\hat{\mathcal{S}}_i = \| \mu_i - \mu_j \|_2, \ \{\sigma_i^m\}_{m=1}^M \sim \mathcal{N}(0, \hat{\mathcal{S}}_i)
\end{equation}
Afterwards, the center $\mu_i$ of gaussian $G_s^i$ is transformed using offset $\sigma_i^m$, and its scale $S_i$ is decayed by ratio $\alpha$. This parallel process generates $M\times I$ new gaussians from $G_s$, collectively denoted as $\hat{G}_s = \{\hat{G}_s^k\}_{k=1}^{M\times I}$.
Upon creation of $\hat{G}_s$, the original set $G_s$ is removed from $G$ to eliminate redundancy, getting an updated gaussian set $G'$. This procedure is denoted as follows:
\begin{equation}
\hat{\mu}_k = \mu_i + R_i \sigma_i^m, \ \hat{S}_k = \alpha \cdot S_i, \ {G}^{\prime} = (G \setminus G_{s}) \cup \hat{G}_{s}
\end{equation}

Next, we adopt an interpolation strategy to improve the local density of gaussians. The gaussians $G_v \subset G$ with high opacity are first selected. For each $G_v^i$, a $k$-nearest neighbor set $\mathcal{N}_i$ is constructed, and distant gaussians are discarded to produce $\hat{\mathcal{N}}_i$. Within the neighborhood, new ellipsoid centers $\hat{\mu}_i$ and colors $\hat{c}_i$ are obtained through the following interpolation operation:
\begin{equation}
\hat{\mathcal{N}}_i = \left\{ G_v^k \in \mathcal{N}_i \mid \| \mu_i - \mu_k\|_2 \leq d_{\max} \right\}
\end{equation}
\begin{equation}
\hat{\mu}_i = \frac{1}{|\hat{\mathcal{N}}_i|} \sum_{G_v^j \in \hat{\mathcal{N}}_i} \mu_j, \ \hat{c}_i = \sum_{G_v^j \in \hat{\mathcal{N}}_i} \frac{\sum \| \mu_i - \mu_j \|_2}{\| \mu_i - \mu_j \|_2} \cdot c_{j} 
\end{equation}
Furthermore, the scale of each interpolated gaussian is derived from the mean distance to its neighboring gaussian centers, while its opacity is initialized to 0.1 and its rotation is set to the identity quaternion before optimization, following 3DGS~\cite{3dgs}. 
In the end, we incorporate new gaussians $\hat{G}_v$ into the original gaussians $G$ to fill the void of local structures.

\begin{figure}[t!]
\centering
\includegraphics[width=\linewidth]{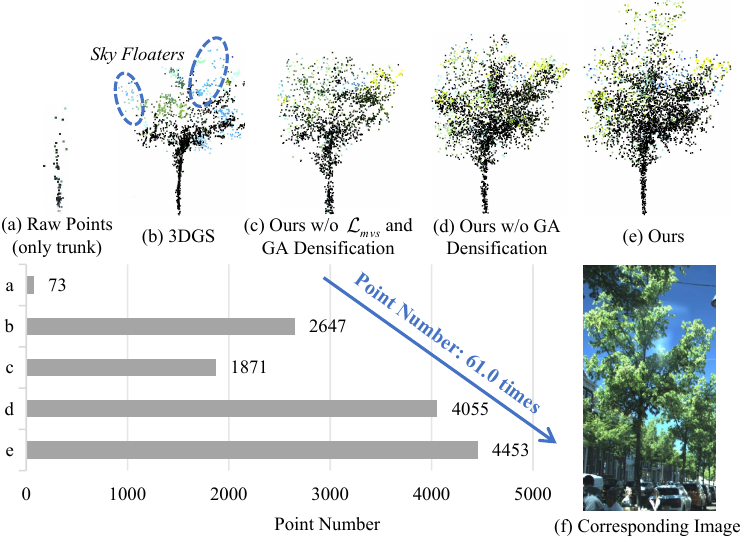}
\vspace{-0.22in}
\caption{Comparison of a tree reconstruction result by 3DGS and our method. We also display the total number of points to reveal the impact of our modules on structural integrity and density. GA is the abbreviation for geometry-aware.}
\vspace{-0.13in}
\label{fig:Tree}
\end{figure} 

\subsection{Gaussian Selective Separation}
\subsubsection{Sky Floater Decoupling}
\begin{figure}[t!]
\centering
\includegraphics[width=\linewidth]{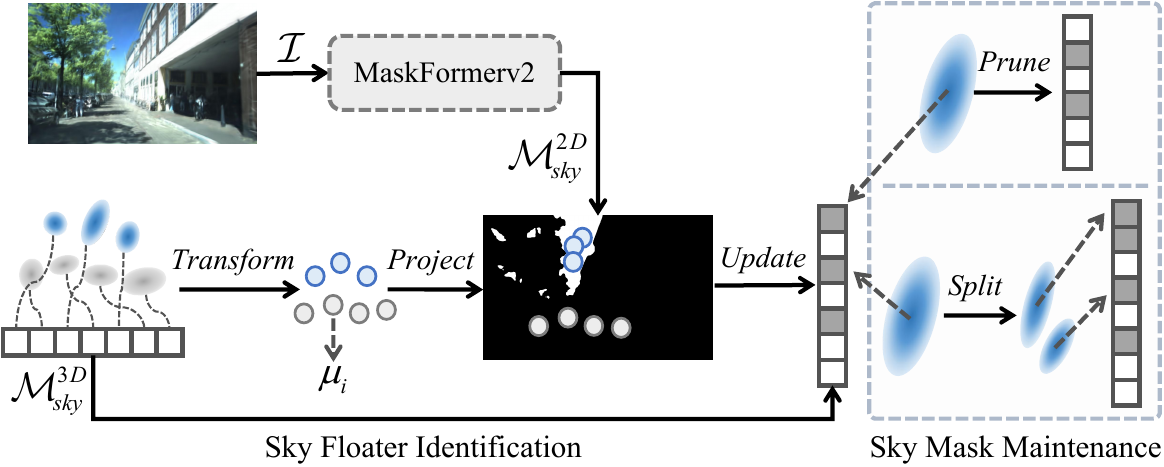}
\vspace{-0.25in}
\caption{The procedures of sky floater identification and sky mask maintenance.}
\vspace{-0.16in}
\label{fig:Sky_mask}
\end{figure}

Radar map inherently contains virtual structures formed by noise, where the gaussians initialized from these artifacts are usually projected into sky regions and undergo splitting, resulting in numerous floaters near the actual structures. In addition, as shown in Fig.~\ref{fig:Tree}(b, f), the area that intersects with the sky in the image also tends to produce sky floaters.
To solve these problems, we maintain a sky mask $\mathcal{M}_\text{sky}^\text{3D} \in \{0, 1\}^N$ (in Fig.~\ref{fig:Sky_mask}), which is updated gradually as new sky gaussians appear by splitting or existing ones are pruned during the optimization process.
To determine which gaussian ellipsoids belong to the sky, we feed the current image $\mathcal{I}$ into MaskFormerv2~\cite{Mask2Former} to yield a 2D mask $\mathcal{M}_\text{sky}^\text{2D} \in \{0, 1\}^{H \times W}$, where $H$ and $W$ are the image height and width. 
We then project the center $\mu_i$ of each gaussian to the image plane, and update the mask $\mathcal{M}_\text{sky}^\text{3D}$ by the corresponding values in $\mathcal{M}_\text{sky}^\text{2D}$:
\begin{equation}
(u_i, v_i) = \Pi(\mu_i), \ \mathcal{M}_\text{sky}^\text{3D}(i) = \mathcal{M}_\text{sky}^\text{2D}( \lfloor u_i \rfloor, \lfloor v_i \rfloor)
\end{equation}
where $\Pi(\cdot)$ and $(u_i, v_i)$ mean the projection operator and pixel coordinates. As a result, unlike directly deleting sky gaussians, our decoupling strategy using a mask not only reduces floaters in the final radar map but also ensures that sky information is not lost during image rendering.

\subsubsection{Neighborhood-based Pruning}
As noted in Sec.~\ref{sec:geometry-aware densification}, uneven distribution of radar points causes oversized gaussians during the initial phase, later shrunk through splitting.
However, most approaches directly remove all large gaussians at a fixed interval, which will mistakenly prune gaussians that are waiting to be split to recover the missing structure beyond the radar FOV. 
Fortunately, we notice that valuable large gaussians usually extend from existing structures with denser neighbors.
Building upon this, we present a neighborhood-based pruning criterion $\mathcal{M}_{p}$ that determines whether the gaussians should be removed by incorporating spatial relationships. Specifically, a gaussian is discarded when its average center distance relative to neighbors exceeds a threshold $\tau_d$, and either its scale $S_i$ or projected 2D radius $r_i^\text{2D}$ surpasses thresholds $\tau_s$ or $\tau_r$:
\begin{equation}
\mathcal{M}_{p} = \mathbb{I}(d(\mu_i) > \tau_d) \land (\mathbb{I}(S_i > \tau_s) \lor \mathbb{I}(r_i^\text{2D} > \tau_r))
\end{equation}
where $d(\cdot)$ denotes the KNN-distance function. In this way, the large gaussians utilized for subsequent completion are avoided from being deleted incorrectly.

\subsection{Multi-view Overlap Regularization}
Existing methods~\cite{2dgs,3dgs,Gaussianpro} usually optimize gaussians by minimizing the photometric loss between the single-view rendered image $\hat{\mathcal{I}}_t$ and the ground truth $\mathcal{I}_t$.
However, geometric ambiguity in single-view observations makes it challenging for both the initial gaussians (with radar noise) and the post-split ones to shift quickly to accurate spatial locations, while risking overfitting in their distribution. 
We thus present a multi-view overlap regularization that renders images $\hat{\mathcal{I}}_{t+w} \ (w \in \{-2L,-L,L,2L\})$ from nearby timesteps and compares them with $\mathcal{I}_{t+w}$ as additional supervision signals for gaussians in the current view. 
Since co-visible regions $\mathcal{A}_{t+w}$ typically exist between $\hat{\mathcal{I}}_{t+w}$ and $\hat{\mathcal{I}}_t$, and satisfy an overlap relationship $\bigcup \mathcal{A}_{t+w} = \hat{\mathcal{I}}_t$ in most cases, this means that the corresponding gaussians for $\hat{\mathcal{I}}_t$ are constrained by at least two perspectives, as shown in Fig.~\ref{fig:optimized_map}.
Thus, gaussian positions and other attributes can be optimized toward unambiguous states more easily. The multi-view overlap regularization is defined by the following functions:
\begin{equation}
\mathcal{L}_c = \frac{1-\lambda}{N} \sum_{i=1}^N |\hat{\mathcal{I}}_t^i - \mathcal{I}_t^i|^\gamma + \lambda \mathcal{L}_\text{SSIM}(\hat{\mathcal{I}}_{t}, \mathcal{I}_{t})
\end{equation}
\begin{equation}
\mathcal{L}_{mvc} = \frac{\lambda_{mvc}}{4} \sum_{i=1}^4 \mathcal{L}_c(\hat{\mathcal{I}}_{t+w_i}, \mathcal{I}_{t+w_i}) + \mathcal{L}_c(\hat{\mathcal{I}}_{t}, \mathcal{I}_{t})
\label{eq:mvc}
\end{equation}
where $\lambda=0.1, \lambda_{mvc}=3.0$ indicate the loss weights, and we adopt L1 loss with exponential weight to strengthen the fitting on difficult samples. We also intuitively exhibit the influence of multi-view overlap regularization in Fig.~\ref{fig:Tree}(c, d).

\begin{figure*}[t!]
\centering
\includegraphics[width=0.99\linewidth]{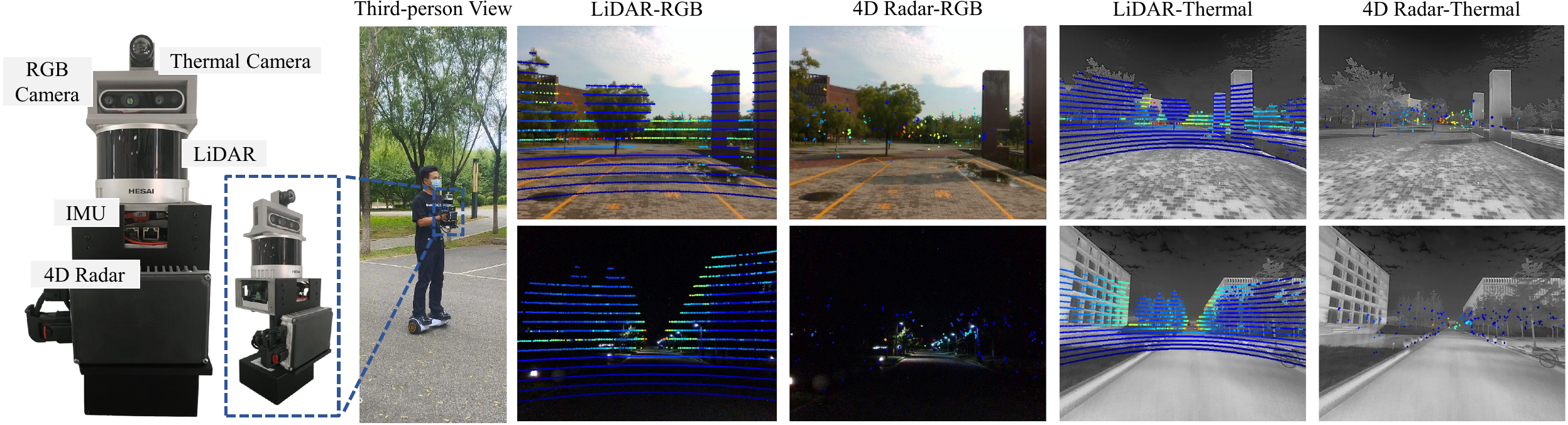}
\vspace{-0.15in}
\caption{The visualization of our handheld equipment and the projection results among different sensors. Multi-sensor data are collected across various campus scenes under both daytime and nighttime conditions.}
\vspace{-0.2in}
\label{fig:Hardware}
\end{figure*} 

\subsection{Loss Functions}
Except for multi-view regularization, we apply prior knowledge derived from visual fundamental models~\cite{Depthanythingv2, Metric3D} to optimize the geometric distribution of gaussians. We first render depths $\hat{\mathcal{D}}$ and normals $\hat{\mathcal{N}}$ through Eq.~\ref{eq:render_dn}. 
We then supervise depths $\hat{\mathcal{D}}$ with L1 loss against the predicted depths $\mathcal{D}$ obtained from DepthAnythingv2~\cite{Depthanythingv2}, while adopting Metric3D~\cite{Metric3D} to get normals $\mathcal{N}$ for regularizing $\hat{\mathcal{N}}$ by normal similarity loss. Both losses are defined in Eq.~\ref{eq:loss_dn}.
\begin{equation}
\hat{\mathcal{D}} = \sum_{i=1}^n d_i \alpha_i \prod_{j=1}^{i-1} (1 - \alpha_j), \ \hat{\mathcal{N}} = \sum_{i=1}^n n_i \alpha_i \prod_{j=1}^{i-1} (1 - \alpha_j)
\label{eq:render_dn}
\end{equation}
\begin{equation}
\mathcal{L}_d = \| (\mathcal{D} - \hat{\mathcal{D}}) \|_1, \ \mathcal{L}_n = 1-\mathcal{N} \cdot \hat{\mathcal{N}}
\label{eq:loss_dn}
\end{equation}
To sum up, the overall loss function is formulated as follows:
\begin{equation}
\mathcal{L} = \mathcal{L}_{mvc} + \mathcal{L}_n + \mathcal{L}_d
\end{equation}
where $\mathcal{L}_{mvc}$ means the multi-view overlap regularization loss (Eq.~\ref{eq:mvc}), enforcing consistency of gaussian attributes across multiple views to resolve spatial ambiguity. 
Then, $\mathcal{L}_n$ and $\mathcal{L}_d$ incorporate monocular geometric priors to guide the gaussians towards forming accurate local structures and smooth surfaces.

\section{Experiment}
\subsection{Datasets}
To thoroughly evaluate our method, we conduct comprehensive experiments not only on public datasets, including View-of-Delft~\cite{VoD} and NTU4DRadLM~\cite{ntu4dradlm}, but also on our self-collected campus dataset to assess both odometry estimation accuracy and map optimization performance in various scenes. 
\subsubsection{View-of-Delft} 
The VoD dataset~\cite{VoD} provides synchronized multi-sensor data, consisting of measurements from 64-beam LiDAR, RGB camera, and ZF FRGen21 4D radar, captured in complex urban traffic environments.
With a total of 8,693 frames, the dataset is structured into 24 continuous sequences. Consistent with the data partitioning scheme of~\cite{4DRVONet}, we divide these sequences into 13 for training, 4 for validation, and 7 for testing. Since each radar frame contains only 300-500 points, VoD presents severe challenges for odometry estimation and map optimization.

\begin{figure}[t!]
\vspace{-0.1in}
\centering
\includegraphics[width=\linewidth]{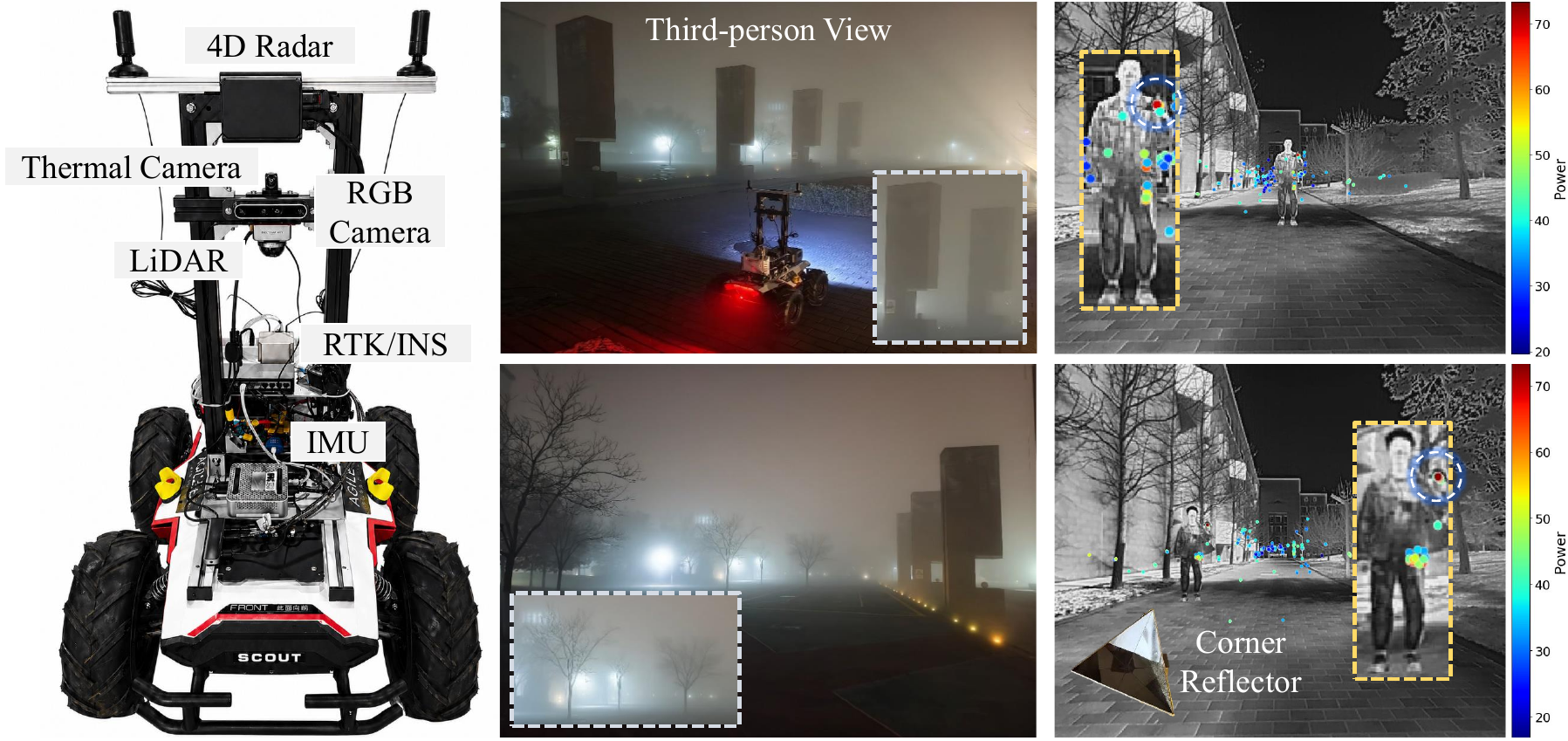}
\vspace{-0.28in}
\caption{Visualization of our mobile robotic platform, data collection in foggy environments, and radar-thermal extrinsic calibration using a corner reflector.}
\vspace{-0.2in}
\label{fig:FogPlatform}
\end{figure}

\subsubsection{NTU4DRadLM} The NTU4DRadLM dataset~\cite{ntu4dradlm} is collected using an Oculii Eagle 4D radar, Livox Horizon LiDAR, an RGB camera, and a thermal camera, which are not time-synchronized.
It captures six trajectories across NTU campus, including structured, semi-structured, and unstructured scenarios.
Data are collected from both a handcart platform moving at around 1$m/s$ and vehicle-mounted systems traveling at 25-30$km/h$. For odometry estimation, we partition the sequences into training (\textit{Garden}, \textit{Nyl}, \textit{Loop1}), validation (\textit{Loop2}), and testing (\textit{Loop3}, \textit{Cp}) sets. 
In particular, the radar point clouds in NTU4DRadLM exhibit a higher density relative to the VoD dataset, typically comprising thousands of points per frame.

\begin{table*}[t!]
\renewcommand\tabcolsep{4.2pt}
\renewcommand{\arraystretch}{1.00}
\caption{4D radar odometry experiment results on View-of-Delft (VoD) dataset.}
\vspace{-0.2in}
\centering
\begin{center}
\begin{tabular}{cc|cc|cc|cc|cc|cc|cc|cc|cc}
\toprule[.045cm]
\multicolumn{2}{c|}{\multirow{2}{*}{Method}} & \multicolumn{2}{c|}{03} & \multicolumn{2}{c|}{04} & \multicolumn{2}{c|}{09} & \multicolumn{2}{c|}{17} & \multicolumn{2}{c|}{19} & \multicolumn{2}{c|}{22} & \multicolumn{2}{c|}{24} & \multicolumn{2}{c}{Mean} \\ \cmidrule(l){3-18} 

& & $t_{rel}$ & $r_{rel}$ & $t_{rel}$ & $r_{rel}$ & $t_{rel}$ & $r_{rel}$ & $t_{rel}$ & $r_{rel}$ & $t_{rel}$   & $r_{rel}$ & $t_{rel}$ & $r_{rel}$ & $t_{rel}$ & $r_{rel}$ & $t_{rel}$ & $r_{rel}$ \\
\hline\hline
\noalign{\smallskip}

\multirow{4}{*}{Classical-based}
&ICP-po2po    
& 0.39  & 1.00    
& 0.21  & 1.14    
& 0.15  & 0.72    
& 0.16  & 0.53    
& 1.40  & 4.70    
& 0.44  & 0.76   
& 0.24  & 0.77  
& 0.43  & 1.37
\\ 

& ICP-po2pl     
& 0.42 & 2.19    
& 0.37 & 1.83    
& 0.50 & 1.32    
& 0.23 & 0.68     
& 3.04 & 5.62    
& 0.42 & 1.20 
& 0.35 & 0.67 
& 0.76 & 1.93
\\ 

& GICP    
& 0.46 & 0.68     
& 0.30 & 0.39     
& 0.51 & 0.32    
& 0.40 & 0.10    
& 0.51 & 1.23    
& 0.34 & 0.57 
& 0.15 & 0.30 
& 0.38 & 0.51
\\ 
& NDT    
& 0.55 & 1.60    
& 0.47 & 0.91     
& 0.46 & 0.56    
& 0.44 & 0.40    
& 1.33 & 2.58    
& 0.47 & 1.10 
& 0.36 & 1.84 
& 0.58 & 1.28 
\\ 
\cmidrule(r){1-18}
\multirow{4}{*}{LiDAR-based}
& Full A-LOAM    
& NA     & NA         
& 0.03  & 0.09        
& 0.04  & 0.19    
& 0.02  & 0.04  
& 0.38  & 1.35  
& 0.06  & 0.18    
& 0.06  & 0.20 
& 0.10  & 0.34
\\ 

& A-LOAM w/o mapping    
& NA     & NA         
& 0.14  & 0.35         
& 0.16  & 1.23    
& 0.09  & 0.26    
& 1.17  & 4.63    
& 0.27  & 0.92 
& 0.16  & 0.81 
& 0.33  & 1.37
\\ 

&LO-Net 
&1.05  & 1.78     
&0.26  & 0.49     
&0.30  & 0.36    
&0.57  & 0.14  
&3.29  & 3.07   
&1.00  & 1.12 
&0.77  & 1.45
&1.03  & 1.20
\\ 

&PWCLO-Net   
&0.26  & 0.37     
&0.31  & 0.40     
&0.38  & 0.55    
&0.27  & 0.39  
&1.23  & 0.91    
&0.23  & 0.35 
&0.46  & 0.82
&0.45  & 0.54
\\ 
\cmidrule(r){1-18}
\multirow{4}{*}{\shortstack{4D Radar-based\\ \textbf{Supervised}}}

&CMFlow
& 0.06 & 0.10     
& 0.05 & 0.09    
& 0.09 & 0.14    
& \underline{0.06} & 0.03    
& \underline{0.28} & 0.94     
& 0.14 & 0.29 
& \underline{0.12}  &0.58 
&0.11  &0.31
\\    
&4DRO-Net
&0.08  &0.10	  
&\underline{0.04} &0.07	 
&0.13 &0.38
&0.09 &0.10  
&0.91 &0.62
&0.23 &0.32
&0.28 &1.20
&0.25 &0.40
\\ 
&CAO-RONet
&\underline{0.05} &\underline{0.03}	  
&\underline{0.04} &\underline{0.04}	 
&\underline{0.06} &\underline{0.09}
&0.10 &\textbf{0.01}
&\textbf{0.02} &\underline{0.05}
&\underline{0.08} &\underline{0.06}
&0.14 &\underline{0.08}
&\underline{0.07} &\underline{0.05}
\\
&DNOI-4DRO 
&\textbf{0.02} &\textbf{0.02}	  
&\textbf{0.02} &\textbf{0.03} 
&\textbf{0.02} &\textbf{0.02}
&\textbf{0.02} &\underline{0.02}
&\textbf{0.02} &\textbf{0.04}
&\textbf{0.03} &\textbf{0.02}
&\textbf{0.03} &\textbf{0.05}
&\textbf{0.02} &\textbf{0.03}
\\

\cmidrule(r){1-18}
\multirow{2}{*}{\shortstack{4D Radar-based\\\textbf{Self-supervised}}}
&RaFlow    
& \underline{0.87}  & \underline{2.09}        
& \underline{0.07}  & \underline{0.44}         
& \underline{0.11}  & \underline{0.09}  
& \underline{0.13}  & \underline{0.03}    
& \underline{1.22}  & \underline{4.09}    
& \underline{0.72}  & \underline{1.34} 
& \underline{0.25} & \underline{1.14}
& \underline{0.48} & \underline{1.32}
\\
& \textbf{Ours}  
& \textbf{0.04} & \textbf{0.07}
& \textbf{0.04} & \textbf{0.03}
& \textbf{0.04} & \textbf{0.05}
& \textbf{0.03} & \textbf{0.02}
& \textbf{0.04} & \textbf{0.04}
& \textbf{0.06} & \textbf{0.10}
& \textbf{0.12} & \textbf{0.05}
& \textbf{0.05} & \textbf{0.05}
\\
\toprule[.045cm]
\vspace{-0.5in}
\end{tabular}
\end{center}		
\label{table:VoD1}
\end{table*}

\begin{table}[t!]
\renewcommand\tabcolsep{2.2pt}
\renewcommand{\arraystretch}{1.1}
\caption{4D radar odometry experiment results on VoD dataset using the same configuration as SelfRO. Blue and red are used to distinguish \textcolor[rgb]{0, 0, 0.75}{supervised} and \textcolor[rgb]{0.65, 0, 0}{self-supervised} methods.} 
\vspace{-0.05in}
\scalebox{0.92}{
\begin{tabular}{c|cc|cc|cc|cc|cc|cc}
\toprule[.045cm]
\multirow{2}{*}{Method} & \multicolumn{2}{c|}{00}                                        & \multicolumn{2}{c|}{03}                                        & \multicolumn{2}{c|}{04}                                        & \multicolumn{2}{c|}{07}                                        & \multicolumn{2}{c|}{23}
& \multicolumn{2}{c}{Mean} \\ \cline{2-13} 
& \multicolumn{1}{l}{$t_{rel}$} & \multicolumn{1}{l|}{$r_{rel}$} & \multicolumn{1}{l}{$t_{rel}$} & \multicolumn{1}{l|}{$r_{rel}$} & \multicolumn{1}{l}{$t_{rel}$} & \multicolumn{1}{l|}{$r_{rel}$} & \multicolumn{1}{l}{$t_{rel}$} & \multicolumn{1}{l|}{$r_{rel}$} & \multicolumn{1}{l}{$t_{rel}$} & \multicolumn{1}{l|}{$r_{rel}$} & \multicolumn{1}{l}{$t_{rel}$} & \multicolumn{1}{l}{$r_{rel}$} \\ \hline \hline
\textcolor[rgb]{0, 0, 0.75}{CMFlow} & \textbf{0.04} & \underline{0.05} & 0.07 & 0.09 & \underline{0.06} & 0.09 & \underline{0.03} & 0.04 & \underline{0.09} & \underline{0.14} & \underline{0.06} & 0.08 \\ 
\textcolor[rgb]{0, 0, 0.75}{4DRO-Net} & 0.08 & \textbf{0.03} & \underline{0.06} & \underline{0.05} & 0.08 & \underline{0.07} & 0.05 & \underline{0.03} & 0.10 & 0.15 & 0.07 & \underline{0.07} \\ 
\textcolor[rgb]{0, 0, 0.75}{CAO-RONet} & \underline{0.05} & \textbf{0.03} & \textbf{0.02} & \textbf{0.03} & \textbf{0.03} & \textbf{0.05} & \textbf{0.02} & \textbf{0.02} & \textbf{0.04} & \textbf{0.06} & \textbf{0.03} & \textbf{0.04} \\ \hline
\textcolor[rgb]{0.65, 0, 0}{RaFlow}  & 0.61 & 0.84 & 0.87 & 1.98 & 0.07 & 0.45 & \underline{0.07} & \underline{0.04} & 0.42 & 1.16 & 0.41 & 0.90 \\
\textcolor[rgb]{0.65, 0, 0}{SelfRO}  & \underline{0.07} & \underline{0.11} & \underline{0.10} & \underline{0.16} & \underline{0.05} & \underline{0.13} & \underline{0.07} & 0.14 & \underline{0.13} & \underline{0.14} & \underline{0.08} & \underline{0.14} \\
\textcolor[rgb]{0.65, 0, 0}{\textbf{Ours}}  & \textbf{0.03} & \textbf{0.04} & \textbf{0.04} & \textbf{0.07} & \textbf{0.03} & \textbf{0.03} & \textbf{0.01} & \textbf{0.02} & \textbf{0.05} & \textbf{0.04} & \textbf{0.03} & \textbf{0.04} \\
\textit{Impr. (\%)} & 57.1 & 63.6 & 60.0 & 56.3 & 40.0 & 76.9 & 85.7 & 50.0 & 61.5 & 71.4 & 62.5 & 71.4 \\
\toprule[.045cm]
\end{tabular}}
\vspace{-0.2in}
\label{table:VoD2}
\end{table}

\begin{table}[t!]
\renewcommand\tabcolsep{1.8pt}
\renewcommand{\arraystretch}{1.15}
\caption{4D radar odometry experiment results on NTU4DRadLM. ``$R$'' and ``$A$'' mean the relative pose error (RPE) and absolute trajectory error (ATE), respectively.}
\vspace{-0.28in}
\begin{center}
\scalebox{0.92}{
\begin{tabular}{c|ccc|ccc|ccc}
\toprule[.045cm]
\multirow{2}{*}{Method} & \multicolumn{3}{c|}{Loop2 (4.79$km$)} & \multicolumn{3}{c|}{Loop3 (4.23$km$)} & \multicolumn{3}{c}{Cp (0.25$km$)} \\ \cline{2-10} 
& $R$ ($m$) & $R$ (°) & $A$ ($m$) & $R$ ($m$) & $R$ (°) & $A$ ($m$) & $R$ ($m$) & $R$ (°) & $A$ ($m$) \\ \hline \hline
ICP & 12.357 & 8.003 & 613.073 & 13.772 & 3.818 & 304.793 & 0.543 & 1.274 & 3.639 \\
NDT & 3.406 & 5.775 & 652.073 & \underline{2.798} & 3.579 & 169.034 & 0.223 & 1.368 & 3.294\\
GICP & 2.448 & \underline{2.517} & \underline{58.155} & 3.161 & \underline{2.052} & 29.224  & \textbf{0.192} & \underline{1.184} & \underline{2.256} \\
APDGICP & 2.459 & 2.616 & 145.674 & 3.072 & 2.209 & \underline{27.871} & \underline{0.216} & 1.241 & 2.408\\ \hline 
\textcolor[rgb]{0, 0, 0.75}{CAO-RONet} & \textbf{1.504} & 4.215 & 265.864 & 4.228 & 2.604 & 481.916 & 0.342 & 2.471 & 15.557 \\  
\textcolor[rgb]{0.65, 0, 0}{RaFlow} & NA & NA & NA & NA & NA & NA & 1.310 & 9.367 & 44.228\\
\textbf{\textcolor[rgb]{0.65, 0, 0}{Ours}} & \underline{1.521} & \textbf{2.448} & \textbf{38.876} & \textbf{2.104} & \textbf{1.769} & \textbf{22.077} & 0.248 & \textbf{1.121} & \textbf{1.749} \\
\toprule[.045cm]
\end{tabular}}
\end{center} 
\vspace{-0.3in}
\label{table:NTU}
\end{table}

\subsubsection{Self-Collected Campus Dataset} To conduct more extensive experiments, we build a handheld data collection system that is more compact and portable compared with the handcart platform in NTU4DRadLM. Our system integrates a 16-beam LiDAR, GPAL 4D radar, RGB camera, thermal camera, and IMU. Here, the 4D radar operates at around 15 Hz and collects about 500 points per frame.
The configuration of our handheld equipment and the projection of the point cloud onto the image are illustrated in Fig.~\ref{fig:Hardware}.
Built on this platform, we collect eight daytime sequences and two nighttime sequences. 
The daytime sequences are leveraged for odometry evaluation, with the data (34,241 frames in total) separated into training, validation, and testing sets with a ratio of 6:1:3. The validation set is formed by \textit{Gym} sequence with 3,124 radar frames, whereas the testing set includes \textit{Building2}, \textit{Parking2}, and \textit{Straight} sequences with 9,701 frames. 
The nighttime data are then utilized to evaluate thermal-based map optimization.
To conduct evaluations under degraded conditions, we construct a mobile robotic platform, as displayed in Fig.~\ref{fig:FogPlatform}, and collect two multi-sensor sequences (\textit{Fog0} and \textit{Fog1}) in foggy environments.
Note that the extrinsic calibration between the 4D radar and the camera is performed following the procedure in~\cite{radarcalib}. The known sensor mounting configuration is used as an initial reference, and the calibration is further refined using OpenCalib~\cite{opencalib} with a corner reflector that produces high-intensity radar returns.

\begin{figure}[t!]
\centering
\includegraphics[width=\linewidth]{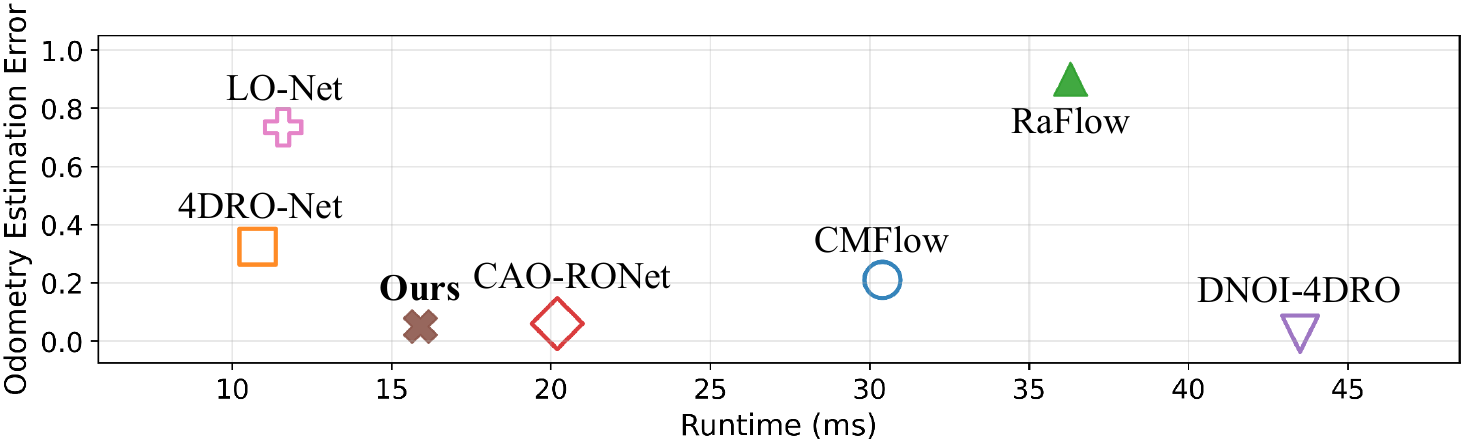}
\vspace{-0.32in}
\caption{Comparison of runtime and odometry estimation error across several methods on VoD dataset. The error is computed as the mean of $r_{rel}$ and $t_{rel}$. Our method achieves an optimal balance between performance and efficiency.}
\vspace{-0.1in}
\label{fig:runtime}
\end{figure} 

\begin{figure}[t!]
\centering
\includegraphics[width=\linewidth]{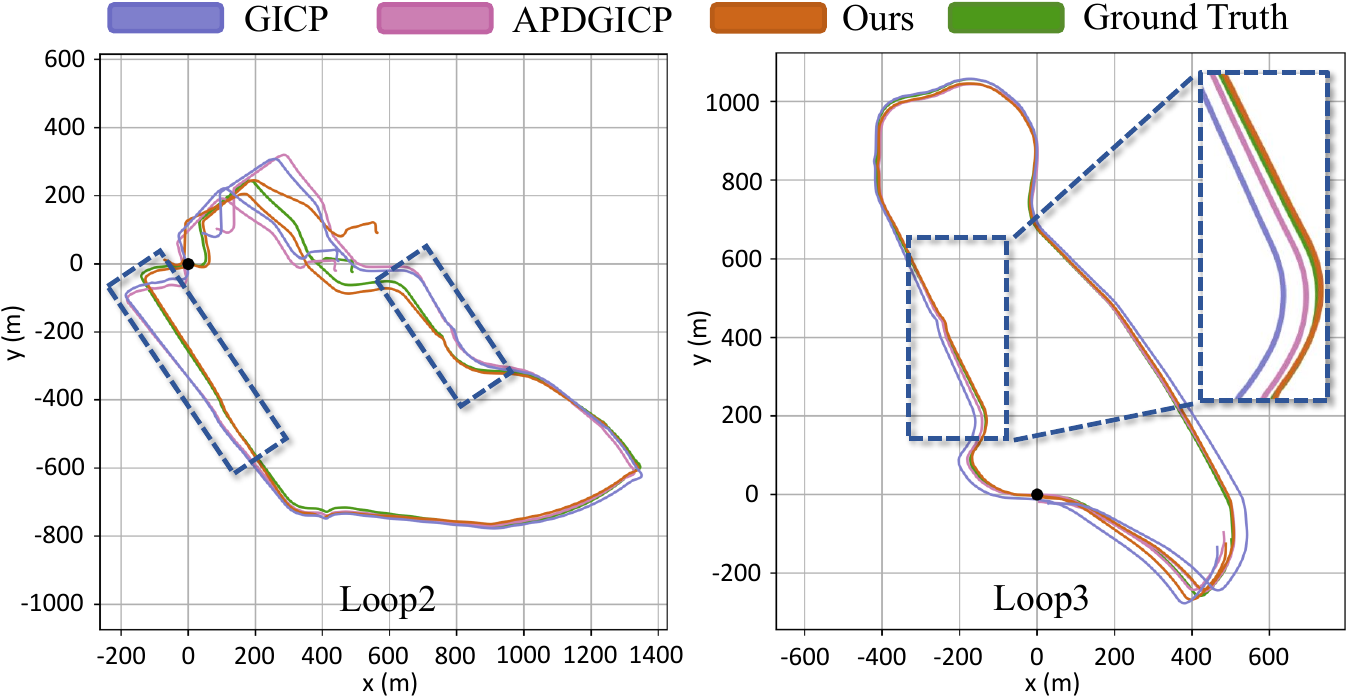}
\vspace{-0.3in}
\caption{The trajectory visualization on the long sequences of NTU4DRadLM.}
\vspace{-0.25in}
\label{fig:trac}
\end{figure} 

\subsection{Implementation Details}
\subsubsection{Data Processing} For odometry data preprocessing, we first apply a height-based filtering to the point clouds, retaining only those points within the vertical range of [-3$m$, 3$m$]. To augment the training data diversity in the VoD dataset, we flip training sequences to generate reversed-motion data. Random translations within ±1$m$ along each axis are also employed to point clouds during training. For DBSCAN-based clustering, we use a 3-meter search radius and a minimum cluster size of 3 points for VoD and self-collected datasets, while configuring a 1.5-meter radius and a 5-point minimum for NTU4DRadLM.
In the map optimization pipeline, we downsample the points of radar maps in the NTU4DRadLM to 25\% of their original density for gaussian initialization and select image keyframes at a 5-frame interval to optimize maps. For the VoD dataset, no sampling is applied to either images or point clouds.

\begin{figure*}[t!]
\centering
\includegraphics[width=\linewidth]{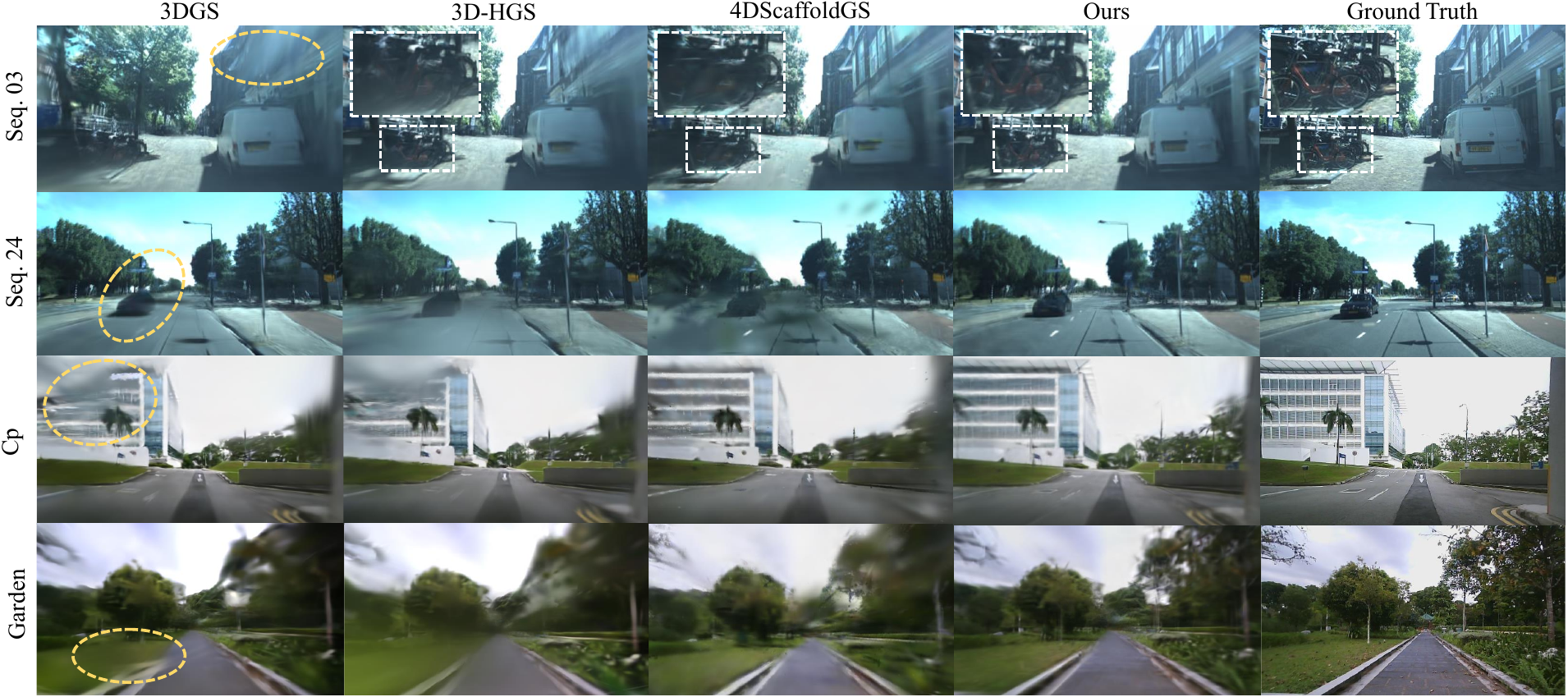}
\vspace{-0.3in}
\caption{Qualitative comparisons of rendering quality on the VoD and NTU4DRadLM datasets. Some representative regions are highlighted for comparison.}
\vspace{-0.15in}
\label{fig:render_image}
\end{figure*} 

\begin{table*}[t!]
\renewcommand\tabcolsep{2.0pt}
\renewcommand{\arraystretch}{1.1}
\caption{The performance of map optimization and image rendering on View-of-Delft (VoD) dataset. ``Odom'' is defined as using poses estimated by our self-supervised odometry to construct initial radar maps.}
\vspace{-0.15in}
\begin{center}
\begin{tabular}{c|ccc|ccc|ccc|ccc|ccc}
\toprule[.045cm]
\multirow{2}{*}{Method} & \multicolumn{3}{c|}{03} & \multicolumn{3}{c|}{04} & \multicolumn{3}{c|}{09} & \multicolumn{3}{c|}{22} & \multicolumn{3}{c}{24} \\ \cline{2-16} & F-Score↑ & CD↓   & MHD↓   & F-Score↑ & CD↓   & MHD↓   & F-Score↑ & CD↓   & MHD↓   & F-Score↑ & CD↓   & MHD↓ & F-Score↑ & CD↓   & MHD↓ \\ \hline \hline

3DGS & 0.339 & 1.171 & 1.642 & 0.351 & \underline{0.951} & 0.721 & 0.364 & 0.987 & 0.589 & 0.349 & \underline{1.171} & 0.831 & \underline{0.318} & \underline{1.728} & \underline{1.485} \\

2DGS  & 0.271 & 1.441 & 1.316  & 0.263 & 1.092 & {0.537} & 0.254 & 1.113 & 0.358 & 0.186 & 1.422 & 0.704 & 0.125 & 2.827 & 17.587 \\

GaussianPro & \underline{0.378} & 1.482 & \underline{1.148} & 0.297    & 1.881 & 1.486 & 0.253 & 2.514 & 1.801  & 0.357 & 1.481 & 0.975 & 0.241 & 3.085 & 6.817 \\

DASH & 0.345 & 1.370 & 1.513 & 0.283 & 1.008 & \underline{0.509} & 0.287 & 1.049 & \underline{0.301} & 0.205 & 1.401 & 0.871 & 0.291 & 1.747 & 2.022 \\

3D-HGS & 0.371 & \underline{1.168} & 1.472 & \underline{0.353} & 1.089 & 0.699 & \underline{0.367} & \underline{0.978} & 0.437 & 0.350 & 1.204 & 0.763 & 0.314 & 1.836 & 2.805 \\

4DScaffoldGS & {0.375} & 1.513 & 1.571 & 0.297 & 1.167 & 0.830 & 0.305 & 1.219 & 0.353 & \underline{0.371} & 1.240 & \underline{0.674} & 0.285 & 2.139 & 4.001 \\

\textbf{Ours} & \textbf{0.509} & \textbf{0.955} & \textbf{0.271} & \textbf{0.501} & \textbf{0.733} & \textbf{0.164} & \textbf{0.556} & \textbf{0.624} & \textbf{0.105} & \textbf{0.571} & \textbf{0.664} & \textbf{0.099} & \textbf{0.428} & \textbf{1.128} & \textbf{0.304} \\ \hline

3DGS (Odom) & 0.208 & 1.507 & 3.983 & 0.285 & 1.229 & 0.977 & 0.311 & 1.049 & 0.918 & 0.241 & 1.548 & 1.923 & 0.254 & 1.809 & 3.017 \\

3D-HGS (Odom) & 0.323 & 1.286 & 1.881 & 0.287 & 1.446 & 0.976 & 0.336 & 1.055 & 0.576 & 0.207 & 2.349 & 1.925 & 0.253 & 1.952 & 3.923 \\

\textbf{Ours} (Odom) & 0.394 & 1.196 & 0.670 & 0.438 & 0.731 & 0.188 & 0.389 & 0.892 & 0.267 & 0.433 & 1.083 & 0.473 & 0.402 & 1.161 & 0.352 \\ \hline \hline

\textbf{} & PSNR↑ & SSIM↑ & LPIPS↓ & PSNR↑ & SSIM↑ & LPIPS↓ & PSNR↑ & SSIM↑ & LPIPS↓ & PSNR↑ & SSIM↑ & LPIPS↓ & PSNR↑ & SSIM↑ & LPIPS↓ \\ \hline

3DGS & 15.965 & 0.505 & 0.511 & 19.184 & \underline{0.666} & \underline{0.389} & 18.271 & 0.696 & 0.408 & 16.727 & 0.609 & 0.476 & 21.858 & 0.762 & 0.321 \\

2DGS & 14.713 & 0.469 & 0.569 & 17.877 & 0.641 & 0.433 & 16.849 & 0.669 & 0.454 & 14.901 & 0.582 & 0.514 & 21.235 & 0.744 & 0.369 \\

GaussianPro & 15.783 & 0.494 & 0.488 & 18.721 & 0.647 & 0.391 & 16.805 & 0.664 & 0.434 & 17.171 & 0.615 & 0.466 & 20.551 & 0.733 & 0.339 \\

DASH & 17.464 & 0.524 & 0.518 & \underline{19.286} & 0.654 & 0.419 & \underline{18.684} & 0.694 & 0.424 & 17.631 & 0.614 & 0.491 & 22.003 & 0.756 & 0.354 \\

3D-HGS & \underline{17.823} & \underline{0.551} & \underline{0.477} & 18.962 & 0.665 & 0.391 & 18.529 & \underline{0.703} & \underline{0.402} & 17.248 & 0.618 & 0.470 & \underline{22.672} & \underline{0.781} & \textbf{0.309} \\

4DScaffoldGS & 17.253 & {0.529} & {0.479} & 19.127 & 0.645 & 0.399 & 17.784 & 0.665 & 0.429 & \underline{18.637} & \underline{0.637} & \underline{0.465} & 20.500 & 0.754 & 0.357 \\

\textbf{Ours} & \textbf{19.473} & \textbf{0.571} & \textbf{0.453} & \textbf{21.366} & \textbf{0.683} & \textbf{0.371} & \textbf{20.862} & \textbf{0.725} & \textbf{0.387} & \textbf{20.938} & \textbf{0.659} & \textbf{0.432} & \textbf{24.208} & \textbf{0.787} & \underline{0.316} \\ \hline

3DGS (Odom) & 15.702 & 0.504 & 0.516 & 19.002 & 0.661 & 0.397 & 17.956 & 0.693 & 0.413 & 16.631 & 0.605 & 0.488 & 21.845 & 0.760 & 0.323 \\

3D-HGS (Odom) & 17.784 & 0.548 & 0.484 & 18.874 & 0.652 & 0.393 & 18.469 & 0.701 & 0.403 & 17.212 & 0.615 & 0.481 & 22.597 & 0.775 & 0.310 \\

\textbf{Ours} (Odom) & 19.306 & 0.561 & 0.467 & 21.295 & 0.681 & 0.376 & 20.821 & 0.724 & 0.388 & 20.879 & 0.657 & 0.438 & 24.218 & 0.788 & 0.316 \\
\toprule[.045cm]
\end{tabular}
\end{center} 
\vspace{-0.3in}
\label{table:GSVod}
\end{table*}

\subsubsection{Network Training \& Gaussian Optimization} For odometry, our model is trained for 60 epochs on a single NVIDIA RTX 4090 GPU through the Adam optimizer with a batch size of 24. Meanwhile, we assign an initial learning rate of 1×10$^{-3}$ with a per-epoch decay rate of 0.9 and employ equal weights for all self-supervised loss terms.
For map optimization, the gaussian parameters are updated utilizing Adam optimizer for 15,000 iterations, with gaussian densification every 100 steps starting at iteration 500 with 5×10$^{-4}$ gradient threshold.

\subsubsection{Evaluation Metrics} For odometry evaluation on the VoD, we adopt the relative pose error (RPE) metric, following previous methods~\cite{CAO-RONet,4DRONet,4DRVONet}. With this metric, we calculate the root mean square error (RMSE) for rotation ($^{\circ}/m$) and translation ($m/m$), with 20$m$ intervals ranging from 20$m$ to 160$m$. 
For the NTU4DRadLM and self-collected datasets, we employ the same approach as 4DRadarSLAM~\cite{4dradarslam} to evaluate absolute trajectory error (ATE) and RPE. 
In our self-collected dataset, reference trajectories for the sequences acquired by the handheld platform are generated through FAST-LIO with loop closure~\cite{Fast-lio}. 
For the foggy sequences, RTK/INS measurements from the mobile robot are adopted as ground-truth poses, since LiDAR data are affected under such conditions.
To assess map quality, we construct LiDAR maps as ground truth and follow previous point cloud completion methods, such as~\cite{RDiffusion, RDiff}, to utilize the L1 Chamfer Distance (CD), Modified Hausdorff Distance (MHD), and F-Score (with a distance threshold of 0.3$m$) as metrics for comprehensive evaluation.
For rendering quality evaluation, we adopt three metrics, namely peak signal-to-noise ratio (PSNR), learned perceptual image patch similarity (LPIPS), and structural similarity index measure (SSIM), as utilized in 3DGS~\cite{3dgs}.

\begin{table}[t!]
\setlength{\tabcolsep}{0.01pt}
\renewcommand{\arraystretch}{1.1}
\caption{The evaluation of map optimization and image rendering on NTU4DRadLM. GSPro and 4DSGS are the abbreviation for GaussianPro and 4DScaffoldGS.}
\vspace{-0.25in}
\begin{center}
\scalebox{0.95}{
\begin{tabular}{c|ccc|ccc|ccc}
\toprule[.05cm]
\multirow{2}{*}{Method} & \multicolumn{3}{c|}{Cp (\textit{Structured})} & \multicolumn{3}{c|}{Garden (\textit{Unstructured})} & \multicolumn{3}{c}{Nyl (\textit{Semi-structured})} \\ \cline{2-10} 
& F-Score↑ & CD↓ & MHD↓ & F-Score↑ & CD↓ & MHD↓ & F-Score↑ & CD↓ & MHD↓ \\ \hline \hline

3DGS   & {0.382} & { 1.104} & {0.521} & {0.545} & {0.749}  & {0.132} & {0.228} & 2.592  & 2.039 \\

2DGS   & 0.222  & 1.109  & 1.219  & 0.287  & 1.037  & 0.268  & 0.061  & {2.073}  & 1.723 \\

GSPro  & 0.297  & 3.512  & 3.212  & 0.501  & 0.877  & 0.221  & 0.198  & 2.087 & {1.384}  \\

DASH & 0.255 & 1.297 & 0.957 & 0.240 & 1.754 & 0.476 & 0.026 & 4.493 & 5.256 \\

3D-HGS & 0.372 & 1.146 & 0.552 & 0.560 & \underline{0.687} & 0.109 & 0.207 & \underline{1.624} & 1.131\\

4DSGS &  \underline{0.444} &  \underline{1.056} &  \underline{0.332} & \underline{0.570} & 1.223 & \underline{0.094} & \underline{0.337} & 3.422 & \underline{0.584} \\

\textbf{Ours} & \textbf{0.615}  & \textbf{0.692} & \textbf{0.091} & \textbf{0.688}  & \textbf{0.467} & \textbf{0.041} & \textbf{0.372}  & \textbf{1.016} & \textbf{0.345}\\ \hline \hline 

\textbf{} & PSNR↑ & SSIM↑ & LPIPS↓ & PSNR↑ & SSIM↑ & LPIPS↓ & PSNR↑  & SSIM↑ & LPIPS↓ \\ \hline

3DGS & {18.385} & {0.578} & {0.466}    & {17.635} & {0.521} & {0.544}    & {15.319}    & {0.488} & {0.591}\\

2DGS  & 17.363 & 0.555 & 0.519 & 14.741 & 0.461 & 0.625 & 14.456 & 0.474 & 0.613 \\

GSPro & 17.411 & 0.546 & 0.506 & 15.908 & 0.473 & 0.547 & 14.793 & 0.466 & 0.604\\

DASH & 18.193 & 0.569 & 0.498 & 16.898 & 0.496 & 0.586 & 14.975 & 0.482 & 0.613 \\

3D-HGS & 18.709 & \underline{0.589} & \underline{0.461} & 17.218 & 0.514 & 0.554 & 15.258 &  0.490 & 0.595 \\

4DSGS & \underline{18.748} & 0.584 & 0.479 & \underline{18.641} & \underline{0.534} & \underline{0.515} & \underline{17.421} & \underline{0.516} & \underline{0.571} \\

\textbf{Ours} & \textbf{19.821} & \textbf{0.601} & \textbf{0.445} & \textbf{19.608} & \textbf{0.549} & \textbf{0.502} & \textbf{18.788} & \textbf{0.536} & \textbf{0.529} \\
\toprule[.05cm]
\end{tabular}}
\end{center} 
\vspace{-0.25in}
\label{table:GSNTU}
\end{table}

\begin{figure}[t!]
\centering
\includegraphics[width=\linewidth]{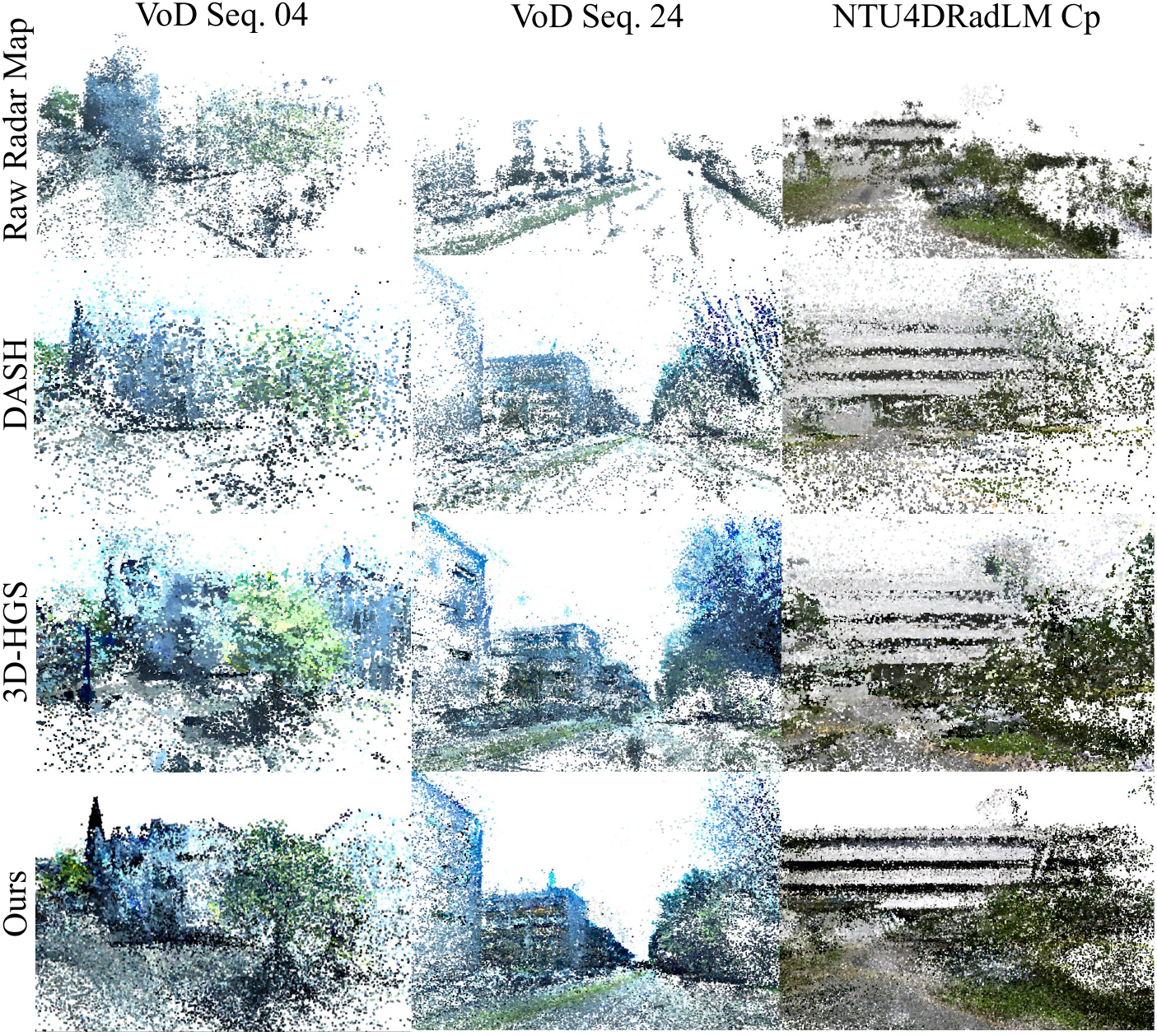}
\vspace{-0.27in}
\caption{Visualization of the optimized 4D radar maps. Our method generates more coherent and complete scenes compared with 3D-HGS and DASH.}
\vspace{-0.15in}
\label{fig:reconstruction}
\end{figure}

\subsection{Performance Comparisons on Public Datasets}
\label{sec:public datasets}
\subsubsection{4D Radar Odometry} As displayed in Tab.~\ref{table:VoD1}, since each radar frame of the VoD dataset only includes 300-500 points, classical approaches such as ICP-po2po, ICP-po2pl, GICP, and NDT are difficult to find sufficient geometric correspondences for pose estimation, resulting in unsatisfactory results. 
Besides, the differences in sensor characteristics make it challenging for the geometric-based method (A-LOAM~\cite{loam}) and learning-based algorithms (LO-Net~\cite{Lo-net} and PWCLO-Net~\cite{Pwclo}) designed for LiDAR to be applicable to 4D radar. 
Among radar-based learning works, our self-supervised framework surpasses RaFlow~\cite{RaFlow} notably based on constraints from various aspects and reduces the gap compared to supervised learning methods obviously in a label-free manner. 
Besides, as shown in Fig.~\ref{fig:runtime}, our self-supervised approach runs at 15.9$ms$, outperforming supervised CAO-RONet~\cite{CAO-RONet} (20.2$ms$) and DNOI-4DRO~\cite{DNOI-4DRO} (43.2$ms$).
In Tab.~\ref{table:VoD2}, since SelfRO~\cite{SelfRONet} is not open-sourced, we employ the same sequence division protocol as described in its paper to train our model for a fair comparison. The results validate the superiority of our method over SelfRO.

\begin{table}[t!]
\centering
\setlength{\tabcolsep}{3.0pt}
\caption{Average performance and computational cost on the VoD dataset. \#GS denote the final number of Gaussians.}
\begin{tabular}{c|ccccccc}
\toprule[.05cm]
Method  & 3DGS   & 2DGS   & GSPro & DASH   & 3D-HGS & 4DSGS & \textbf{Ours}   \\ \hline \hline
F-Score & 0.344  & 0.220  & 0.305       & 0.282  & \underline{0.351}  & 0.327        & \textbf{0.513}  \\
PSNR    & 18.401 & 17.115 & 17.806      & 19.014 & \underline{19.047} & 18.523       & \textbf{21.369} \\ \hline
\#GS    & 921K   & 527K   & 2525K       & 350K   & 933K   & 373K        & 507K   \\
Memory  & 11.3G  & 3.9G   & 18.7G       & 10.6G  & 11.4G  & 3.8G         & 11.8G  \\
FPS     & 103    & 145    & 163         & 248    & 156    & 139          & 175    \\
\toprule[.05cm]
\end{tabular}
\vspace{-0.2in}
\label{tab:efficiency}
\end{table}

\begin{figure}[t]
\centering
\includegraphics[width=\linewidth]{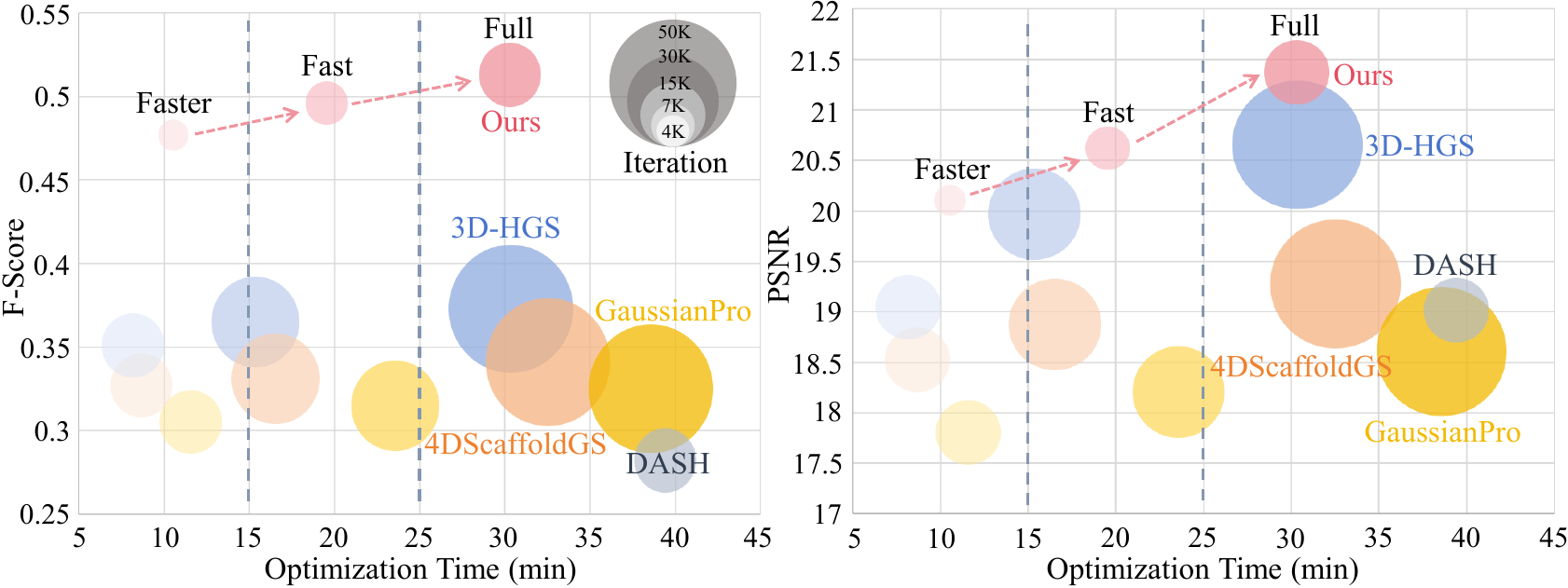}
\vspace{-0.30in}
\caption{Reconstruction performance as a function of optimization time on the VoD dataset. Different gaussian-based approaches are compared within similar optimization-time ranges by varying the number of optimization iterations.}
\vspace{-0.12in}
\label{fig:efficiency}
\end{figure} 

\begin{figure}[t!]
\centering
\includegraphics[width=\linewidth]{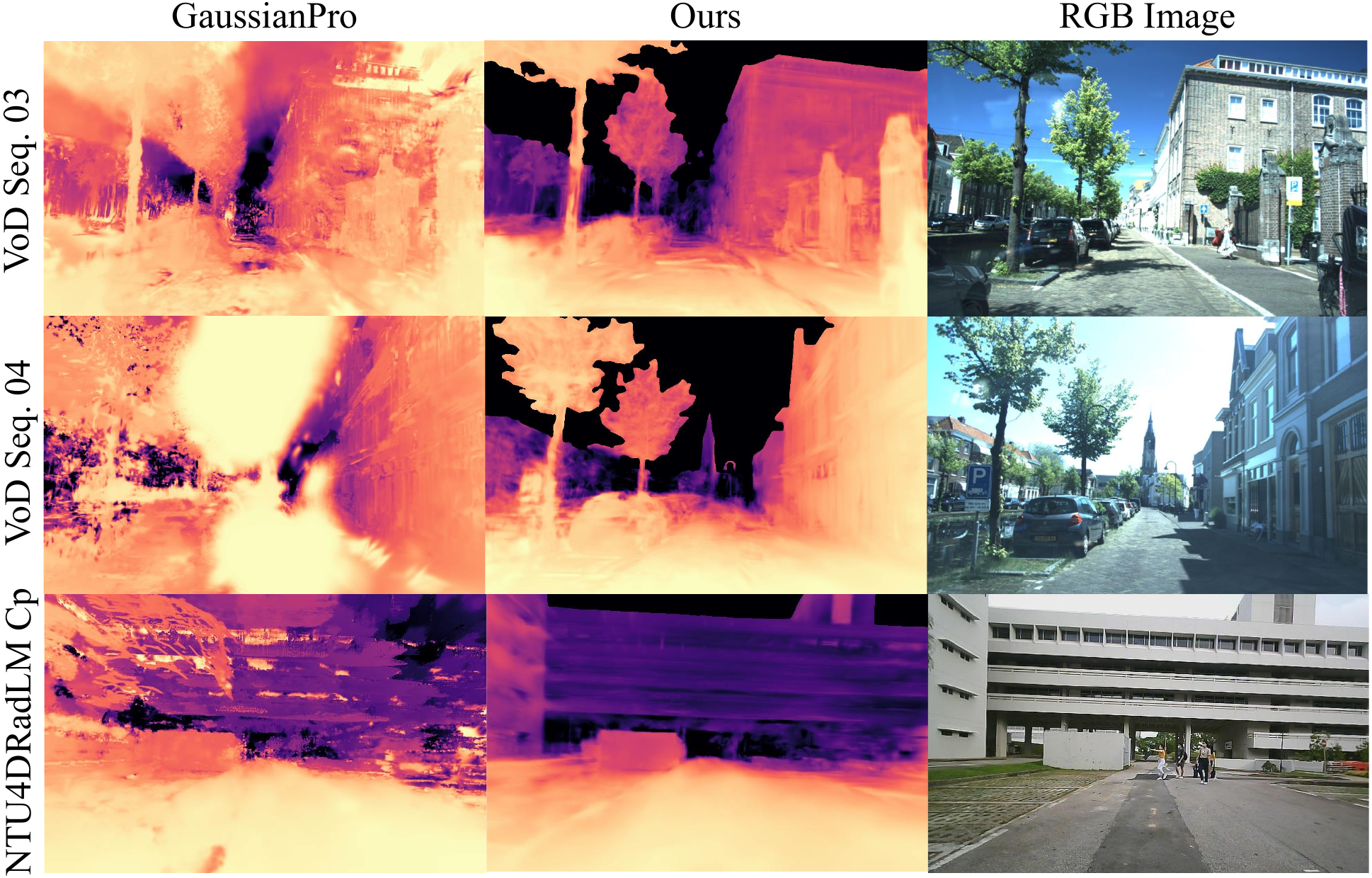}
\vspace{-0.28in}
\caption{The visualization of rendered depths. Our depth maps show smoother surfaces, proving the superior geometric quality of reconstructed maps.}
\vspace{-0.2in}
\label{fig:depth_image}
\end{figure}

Due to lacking time synchronization in NTU4DRadLM, we first perform an interpolation operation to compute the ground-truth poses corresponding to the 4D radar timestamps and then train CAO-RONet~\cite{CAO-RONet}. However, the results in Tab.~\ref{table:NTU} show that \cite{CAO-RONet} suffers from poor performance, implying that supervised algorithms heavily rely on precise time synchronization and accurate poses as supervision signals.
In contrast, our self-supervised method breaks free from these limitations and still works well.
Especially for long sequences (\textit{Loop2} and \textit{Loop3}) with 4.79 $km$ and 4.23$km$, we achieve ATE of 38.876$m$ and 22.077$m$, representing a reduction of 73\% and 21\% compared to APDGICP~\cite{4dradarslam} (145.674$m$ and 27.871$m$), respectively.

\begin{figure*}[t!]
\centering
\includegraphics[width=\linewidth]{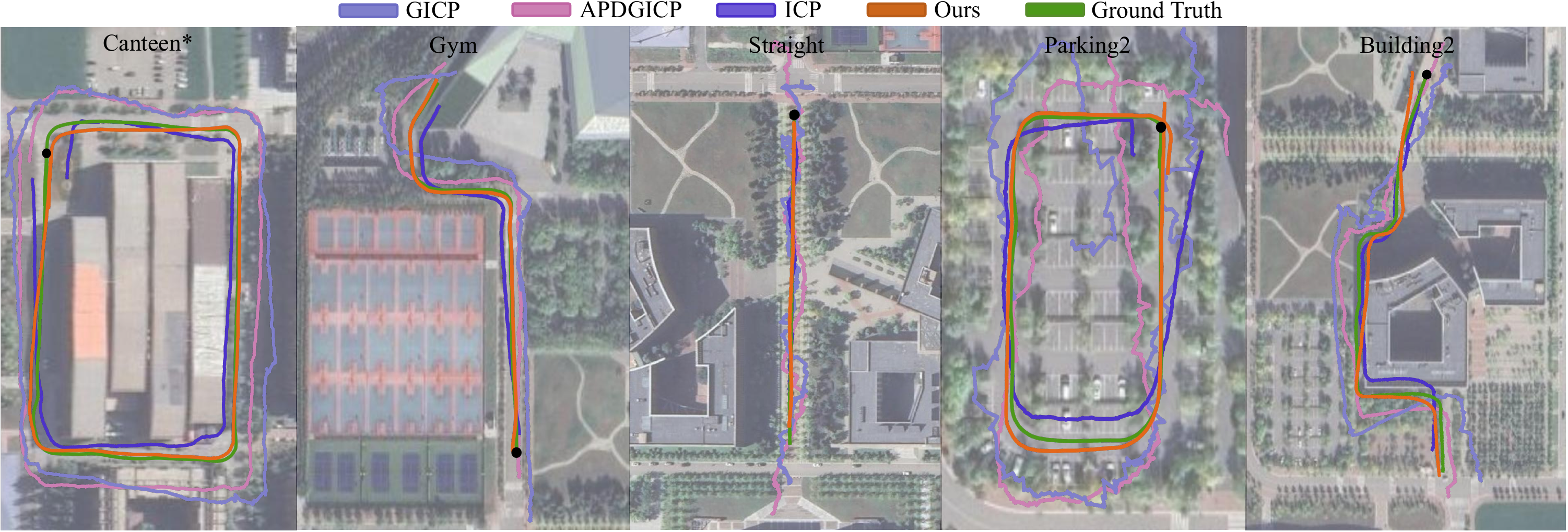}
\vspace{-0.3in}
\caption{Trajectory comparison of various methods using different colors on our self-collected campus dataset. * indicates the sequence from the training set.}
\vspace{-0.27in}
\label{fig:trac_neu}
\end{figure*}

In addition, Fig.~\ref{fig:trac} further demonstrates that our trajectories in kilometer-level sequences are closer to the ground truth than traditional GICP and APDGICP designed for 4D radar, especially in rotation. 
This improvement could be attributed to our method's ability to learn effective inter-frame correspondences through multi-level constraints, including noise-tolerant geometric consistency, teacher guidance, feature contrast, and constant-acceleration motion modeling. And the effectiveness of these constraints is validated in ablation studies (Sec.~\ref{sec:Self-supervised Signals}). We also show more visualizations of trajectories based on our self-collected dataset in Fig.~\ref{fig:trac_neu}. 

\begin{figure*}[t!]
\centering
\includegraphics[width=\linewidth]{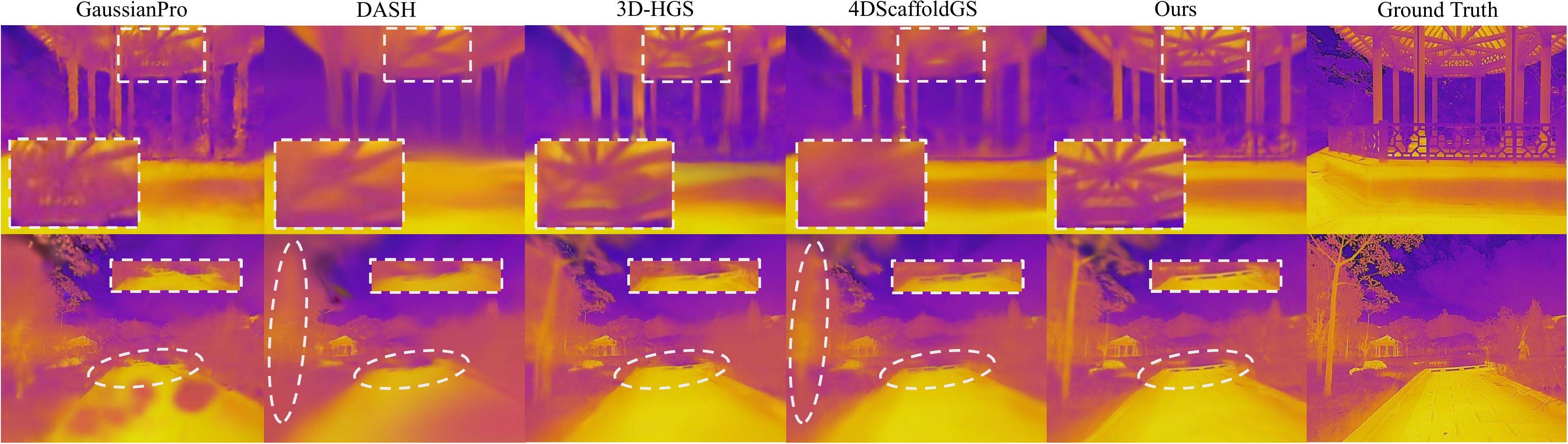}
\vspace{-0.3in}
\caption{Comparisons of thermal images rendered in the garden sequence of NTU4DRadLM. We highlight some representative regions for intuitive comparison.}
\vspace{-0.25in}
\label{fig:thermal_render}
\end{figure*} 

\begin{table}[t!]
\centering
\setlength{\tabcolsep}{4.0pt}
\renewcommand{\arraystretch}{1.1}
\caption{Geometric accuracy and rendering quality on the garden sequence of NTU4DRadLM based on thermal images.}
\vspace{-0.05in}
\begin{tabular}{c|ccc|ccc}
\toprule[.05cm]
Method & F-Score↑ & CD↓ & MHD↓ & PSNR↑ & SSIM↑ & LPIPS↓  \\ \hline \hline
3DGS & 0.572 & {0.561} & 0.102 & 21.984 & 0.676 & 0.491 \\
2DGS & 0.295 & 0.830 & 0.224 & 18.456 & 0.634 & 0.556 \\
GaussianPro & \underline{0.633} & 0.590 & \underline{0.080}& {22.355} & {0.680} & {0.478}  \\
DASH & 0.228 & 0.998 & 0.329 & 21.895 & 0.668 & 0.506\\ 
3D-HGS & 0.579 & \underline{0.549} & 0.089 & 22.572 & 0.679 & 0.487 \\
4DScaffoldGS & 0.607 & 0.599 & 0.084 & \underline{24.695} & \underline{0.701} & \underline{0.472} \\
\textbf{Ours} & \textbf{0.666} & \textbf{0.445} & \textbf{0.058} & \textbf{25.875} & \textbf{0.715} & \textbf{0.436} \\
\toprule[.05cm]
\end{tabular}
\vspace{-0.2in}
\label{tab:thermal}
\end{table}

\begin{table}[t!]
\centering
\setlength{\tabcolsep}{4.0pt}
\caption{The effects of different self-supervised loss terms.}
\vspace{-0.07in}
\begin{tabular}{@{}cccccc|c|cc@{}}
\toprule[.05cm]
$\mathcal{L}_{cd}$ & $\mathcal{L}_{co}$ & $\mathcal{L}_{ga}$ & $\mathcal{L}_{tg}$ & $\mathcal{L}_{fc}$ & $\mathcal{L}_{cm}$ & Cluster & \textit{Mean} $t_{rel}$ & \textit{Mean} $r_{rel}$ \\ 
\hline \hline
$\checkmark$ & & & & & & $\checkmark$ & 0.219 & 0.332 \\
$\checkmark$ & $\checkmark$ & & & & & $\checkmark$ & 0.190 & 0.235 \\
$\checkmark$ & $\checkmark$ & $\checkmark$ & & & & $\checkmark$ & 0.124 & 0.175 \\
$\checkmark$ & $\checkmark$ & $\checkmark$ & $\checkmark$ &  &  & $\checkmark$ & 0.078 & 0.091 \\
$\checkmark$ & $\checkmark$ & $\checkmark$ & $\checkmark$ & $\checkmark$ & & $\checkmark$ & 0.056 & 0.068 \\
$\checkmark$ & $\checkmark$ & $\checkmark$ & $\checkmark$ & $\checkmark$ & $\checkmark$ & & 0.068 & 0.108 \\
$\checkmark$ & $\checkmark$ & $\checkmark$ & $\checkmark$ & $\checkmark$ & $\checkmark$ & $\checkmark$ & \textbf{0.053} & \textbf{0.053} \\
\toprule[.05cm]
\end{tabular}
\label{tab:ablation_odm}
\vspace{-0.2in}
\end{table}

\subsubsection{Map Optimization} In this part, we select five sequences from test sets of the VoD dataset, then use ground-truth poses and predicted odometry to build initial radar maps respectively.
The former isolates map optimization from odometry errors, enabling a more focused and pure evaluation of our proposed method. The radar maps are further optimized by differentiable rasterization.
As illustrated in Tab.~\ref{table:GSVod}, most gaussian-based approaches suffer from limited geometric accuracy. 
In particular, 2DGS produces a large MHD on Seq. 24, mainly owing to the degeneration of elongated gaussians, causing over-splitting and many spatially misplaced primitives.
In contrast, our algorithm achieves superior geometric accuracy in the optimized maps, with average CD and MHD of 0.821 and 0.189, considerably lower than those of 3D-HGS~\cite{3d-hgs} (1.255 CD and 1.235 MHD).
We attribute these advantages to radar-specific growth strategy for improving structural completeness, selective separation for reducing floaters while preserving useful gaussians, and multi-view overlap regularization for alleviating spatial ambiguity.
Then, these well-regularized gaussians support accurate scene representation, resulting in better rendering quality compared to previous methods.
Furthermore, we conduct experiments on NTU4DRadLM across diverse scenarios, as reported in Tab.~\ref{table:GSNTU}. 
The results indicate that our approach maintains advantages in structural quality and rendering fidelity over other algorithms. 
For the NTU4DRadLM dataset, gaussian colors are initialized by temporally aligning 4D radar points and camera images via pose interpolation, projecting the radar points onto the images through calibrated camera-radar parameters, and sampling the corresponding pixel colors. 
The same process is also used for our self-collected dataset in Sec.~\ref{sec:our datasets}.

\begin{figure}[t!]
\centering
\includegraphics[width=\linewidth]{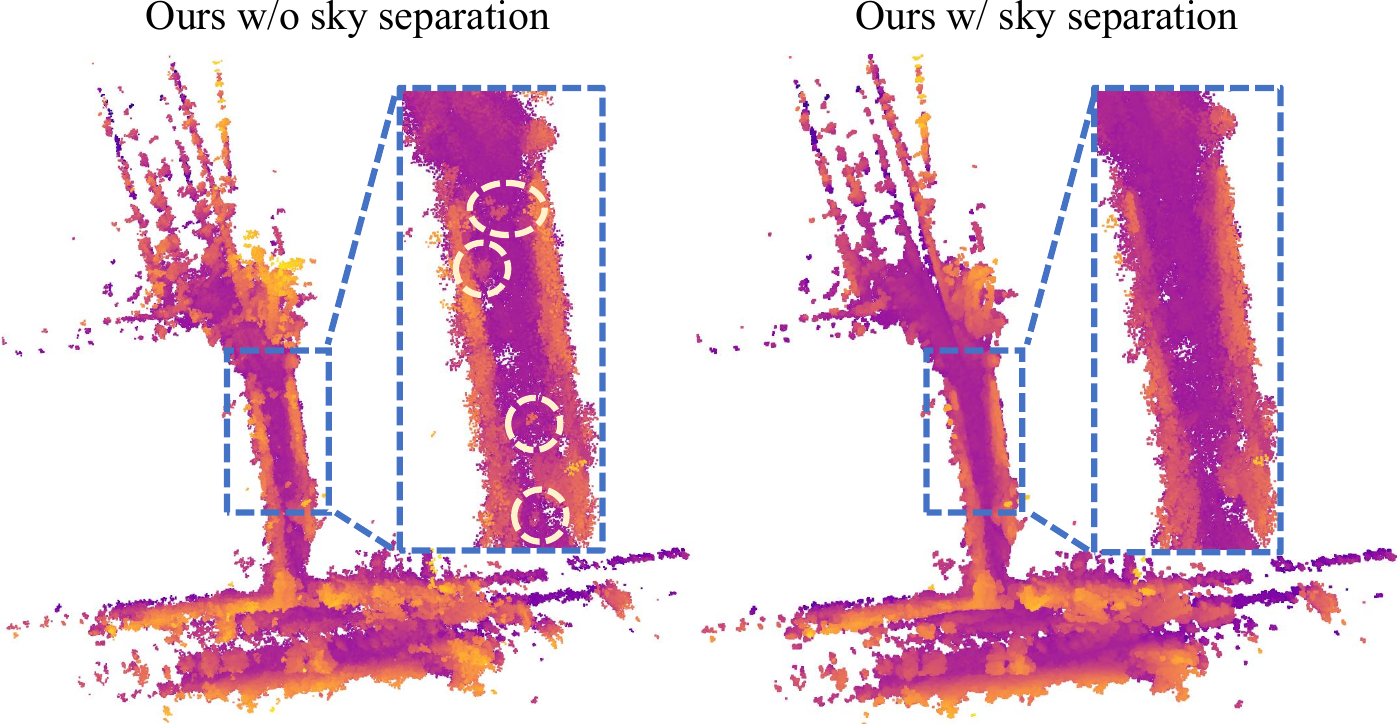}
\vspace{-0.3in}
\caption{The effect of sky separation. Floaters are marked using yellow circles.}
\vspace{-0.2in}
\label{fig:sky}
\end{figure} 

As shown in Fig.~\ref{fig:render_image}, our method achieves a higher-fidelity rendering of environmental details. Especially, it better reconstructs buildings extending beyond the radar's FOV and ground with weak initialization, primarily due to our gaussian growth strategy. 
To intuitively show the effect of map optimization, we present the original input and optimized radar maps in Fig.~\ref{fig:reconstruction}. 
Although 3D-HGS and DASH~\cite{DASH} recover missing structures to some extent, the reconstructed geometry still exhibits uneven distribution and contains blurred artifacts. 
In contrast, our approach restores ground surfaces via depth-assisted completion, with clearer and more coherent structures reconstructed using geometry-aware densification.
We also present rendered depth maps in Fig.~\ref{fig:depth_image}, further demonstrating smoother surfaces and better local consistency in the optimized scene representation.

Moreover, we report the average computational cost on the VoD dataset. As summarized in Tab.~\ref{tab:efficiency}, our approach achieves favorable map reconstruction results with a relatively compact representation of 507K Gaussians, while maintaining modest GPU memory usage and 175 FPS rendering speed.
In Fig.~\ref{fig:efficiency}, we compare different methods under comparable optimization-time ranges by varying the number of optimization iterations. 
Across the three time regimes (5-15 $min$, 15-25 $min$, and $\geq$25 $min$), our faster, fast, and full variants achieve better performance than competing methods, indicating the effectiveness of our radar-oriented design for map enhancement in improving both geometric quality and rendering fidelity.

\begin{figure*}[t!]
\centering
\includegraphics[width=\linewidth]{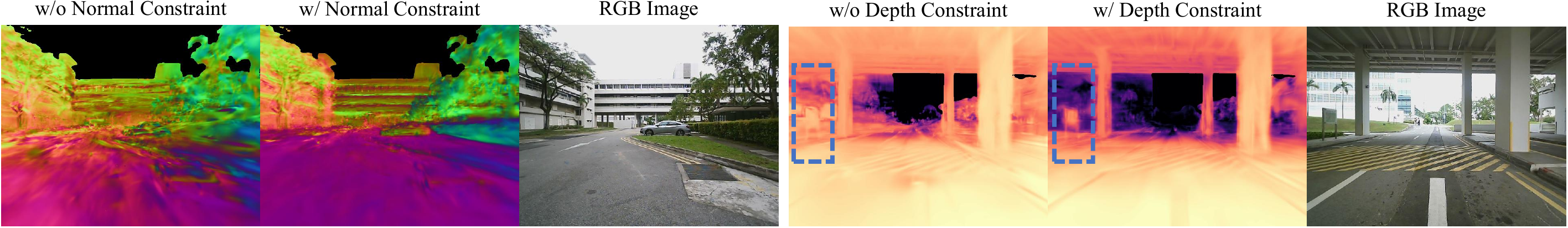}
\vspace{-0.32in}
\caption{Effects of normal and depth constraints. Results display that the former smooth surfaces, while the latter enhance the spatial accuracy of structures.}
\vspace{-0.1in}
\label{fig:normal_image}
\end{figure*} 

\begin{table}[t!]
\centering
\setlength{\tabcolsep}{3.5pt}
\renewcommand{\arraystretch}{1.1}
\caption{The effects of our self-supervised strategy selection.}
\vspace{-0.07in}
\begin{tabular}{l|cc}
\toprule[.05cm]
Strategy & \textit{Mean} $t_{rel}$ & \textit{Mean} $r_{rel}$  \\ \hline \hline
$\mathcal{L}_{cd}$: w/o Cluster Weighting & 0.063 & 0.085 \\
$\mathcal{L}_{co}$: Polar Coordinate $\rightarrow$ Cartesian Coordinate & 0.083 & 0.086 \\
$\mathcal{L}_{tg}$: w/o Selection Mechanism & 0.088 & 0.106 \\
$\mathcal{L}_{fc}$: w/ Neighbor as Negative Sample & 0.064 & 0.139 \\
$\mathcal{L}_{cm}$: Acc. Consistency $\rightarrow$ Vel. Consistency & 0.066 & 0.068 \\
\textbf{Ours} & \textbf{0.053} & \textbf{0.053} \\
\toprule[.05cm]
\end{tabular}
\label{tab:more_ablation_odm}
\vspace{-0.2in}
\end{table}

\begin{table}[t!]
\centering
\setlength{\tabcolsep}{1.3pt}
\caption{The effects of gaussian update and regularization strategy.}
\vspace{-0.05in}
\begin{tabular}{l|ccc|ccc}
\toprule[.05cm]
\multirow{2}{*}{Strategy} & \multicolumn{3}{c|}{Geometric Quality} & \multicolumn{3}{c}{Rendering Quality} \\
\cmidrule(lr){2-4} \cmidrule(lr){5-7} 
 & F-Score↑ & CD↓ & MHD↓ & PSNR↑ & SSIM↑ & LPIPS↓ \\ 
\hline \hline
Baseline & 0.401 & 1.063 & 0.583 & 17.546 & 0.525 & 0.547 \\ 

+ Ground Completion & 0.422 & 0.937 & 0.506 & 17.885 & 0.531 & 0.540 \\ 

+ NB Pruning & 0.465 & 0.898 & 0.377 & 18.064 & 0.538 & 0.535 \\ 

+ GA Densification & 0.511 & 0.837 & 0.270 & 18.565 & 0.542 & 0.527 \\ 

+ Sky Decoupling & 0.528 & 0.787 & 0.227 & 18.582 & 0.542 & 0.526 \\ 

+ Depth \& Normal Loss & 0.542 & 0.750 & 0.186 & 18.738 & 0.549 & 0.513  \\ 

+ Multi-view Constraint & \textbf{0.558} & \textbf{0.725} & \textbf{0.159} & \textbf{19.406} & \textbf{0.562} & \textbf{0.492} \\ 
\toprule[.05cm]
\end{tabular}
\label{tab:ablation_gaus}
\vspace{-0.28in}
\end{table} 

\subsubsection{Application Potential with Thermal Image} 
While RGB images are commonly used, their quality may deteriorate under low-light conditions, whereas thermal images remain relatively stable in such environments.
To reveal the potential of thermal sensing and the adaptability of our approach to different visual sensors, we adopt thermal images for map optimization in this part.
As NTU4DRadLM does not provide thermal data for the \textit{Cp} and \textit{Nyl} sequences, we conduct the evaluation on \textit{Garden}, with the results shown in Tab.~\ref{tab:thermal}. 
In this setting, we omit the normal constraint since Metric3D~\cite{Metric3D} does not support normal estimation from thermal images.
Nevertheless, our method still achieves a PSNR of 25.875 and a CD of 0.445, surpassing the performance of 4DScaffoldGS~\cite{4DScaffoldGS} (24.695 / 0.599).
Moreover, Fig.~\ref{fig:thermal_render} demonstrates that the thermal images rendered by our approach recover finer details, especially in regions with poor gaussian initialization, such as the ground and the upper parts of structures.
This advantage is consistent with RGB rendering results in Fig.~\ref{fig:render_image}, suggesting the effectiveness of our designs across visual modalities.
For a broader evaluation in real-world scenes, we collect multi-sensor data through a handheld device and a mobile robot (Figs.~\ref{fig:Hardware} and~\ref{fig:FogPlatform}) under different conditions.
More comprehensive experiments and detailed discussions are presented in Sec.~\ref{sec:our datasets}.

\subsection{Ablation Studies}
\label{sec:Ablation Studies}
In this section, we conduct ablation studies through progressively adding modules and replacing core strategies to evaluate their contributions and the rationality of our framework design.

\begin{table}[t!]
\centering
\setlength{\tabcolsep}{2.8pt}
\renewcommand{\arraystretch}{1.1}
\caption{The effects of view count on multi-view regularization.}
\vspace{-0.05in}
\begin{tabular}{l|ccc|ccc}
\toprule[.05cm]
Number of views & F-Score↑ & CD↓ & MHD↓ & PSNR↑ & SSIM↑ & LPIPS↓\\ \hline \hline
Two & 0.549 & 0.736 & 0.171 & 19.121 & 0.554 & 0.509 \\
Three & 0.553 & 0.729 & 0.166 & 19.291 & 0.559 & 0.499 \\
\textbf{Ours} (Five) & \textbf{0.558} & \textbf{0.725} & \textbf{0.159} & \textbf{19.406} & \textbf{0.562} & \textbf{0.492}\\
\toprule[.05cm]
\end{tabular}
\label{tab:view_number}
\vspace{-0.25in}
\end{table} 

\subsubsection{Self-supervised Signals}
\label{sec:Self-supervised Signals}
In Tab.~\ref{tab:ablation_odm}, we perform a set of experiments on the VoD with extremely sparse points. $\mathcal{L}_{cd}$ and $\mathcal{L}_{co}$ are cluster-based distance and column occupancy loss. $\mathcal{L}_{tg}$ and $\mathcal{L}_{ga}$ stand for teacher guidance and GMM alignment loss, while $\mathcal{L}_{fc}$ and $\mathcal{L}_{cm}$ represent feature-level and temporal constraints.
At first, the use of $\mathcal{L}_{co}$ leads to reduced odometry errors, especially in $r_{rel}$, as its larger-scale column comparison with occupancy representation is more sensitive to rotational discrepancies and can ease noise from point-to-point constraint $\mathcal{L}_{cd}$.
Then, by incorporating soft labels derived from geometry-based pose refinement, $\mathcal{L}_{tg}$ guides the network to predict more reasonable results and avoid getting stuck in local optima due to rigid geometric constraints.
Using $\mathcal{L}_{fc}$ further reduces $t_{rel}$ and $r_{rel}$, validating that the discriminative features derived by contrastive learning of farthest and nearest points make model easier to identify inter-frame relationships.
Besides, the results in the 5$^{th}$, 6$^{th}$ and 7$^{th}$ rows demonstrate that the acceleration constraint $\mathcal{L}_{cm}$ is crucial for suppressing abnormal poses, and that the cluster cues are critical for reducing feature correlation between unrelated instances.

\begin{table*}[t!]
\renewcommand\tabcolsep{2.5pt}
\renewcommand{\arraystretch}{1.1}
\caption{Comparison of 4D radar odometry results on the daytime sequence of our self-collected campus dataset.}
\vspace{-0.15in}
\begin{center}
\begin{tabular}{c|ccc|ccc|ccc|ccc|ccc}
\toprule[.045cm]
\multirow{2}{*}{Method} & \multicolumn{3}{c|}{Gym (300$m$)} & \multicolumn{3}{c|}{Building2 (370$m$)} & \multicolumn{3}{c|}{Straight (205$m$)} & \multicolumn{3}{c|}{Parking2 (234$m$)} & \multicolumn{3}{c}{Mean}\\ \cline{2-16} 
& RPE($m$) & RPE(°) & ATE($m$) & RPE($m$) & RPE(°) & ATE($m$) & RPE($m$) & RPE(°) & ATE($m$) & RPE($m$) & RPE(°) & ATE($m$) & RPE($m$) & RPE(°) & ATE($m$) \\ \hline \hline
ICP & 0.886 & \underline{1.913} & 7.677 & \underline{0.903} & \underline{2.149} & \underline{10.330} & \underline{1.414} & 2.612 & \underline{13.945} & \underline{0.770} & \textbf{1.939} & \underline{5.595} & \underline{0.993} & \underline{2.153} & \underline{9.387} \\
NDT & \underline{0.785} & 3.715 & \underline{4.751} & 2.161 & 8.463 & 24.089 & 5.617 & 32.763 & 26.836 & 2.046 & 9.028 & 20.961 & 2.652 & 13.492 & 19.159 \\
GICP & 3.337 & 5.031 & 26.357 & 3.721 & 6.532 & 22.384 & 5.754 & 10.316 & 16.307 & 3.983 & 5.893 & 16.490 & 4.199 & 6.943 & 20.384 \\
APDGICP & 1.608 & 2.914 & 11.619 & 2.196 & 3.932 & 17.340 & 5.049 & 6.047 & 29.351 & 2.359 & 3.082 & 12.517 & 2.803 & 3.994 & 17.707 \\ \hline 
RaFlow & 5.623 & 9.010 & 50.921 & 5.012 & 10.804 & 48.504 & 5.112 & \underline{2.569} & 46.497 & 4.666 & 13.803 & 32.208 & 5.103 & 9.047 & 44.533\\
\textbf{Ours} & \textbf{0.440} & \textbf{1.878} & \textbf{2.316} & \textbf{0.477} & \textbf{1.778} & \textbf{7.166} & \textbf{0.542} & \textbf{1.913} & \textbf{4.882} & \textbf{0.626} & \underline{2.728} & \textbf{2.663} & \textbf{0.521} & \textbf{2.074} & \textbf{4.257} \\
\toprule[.045cm]
\end{tabular}
\end{center} 
\vspace{-0.37in}
\label{table:NEU}
\end{table*}

\begin{figure}[t!]
\centering
\includegraphics[width=\linewidth]{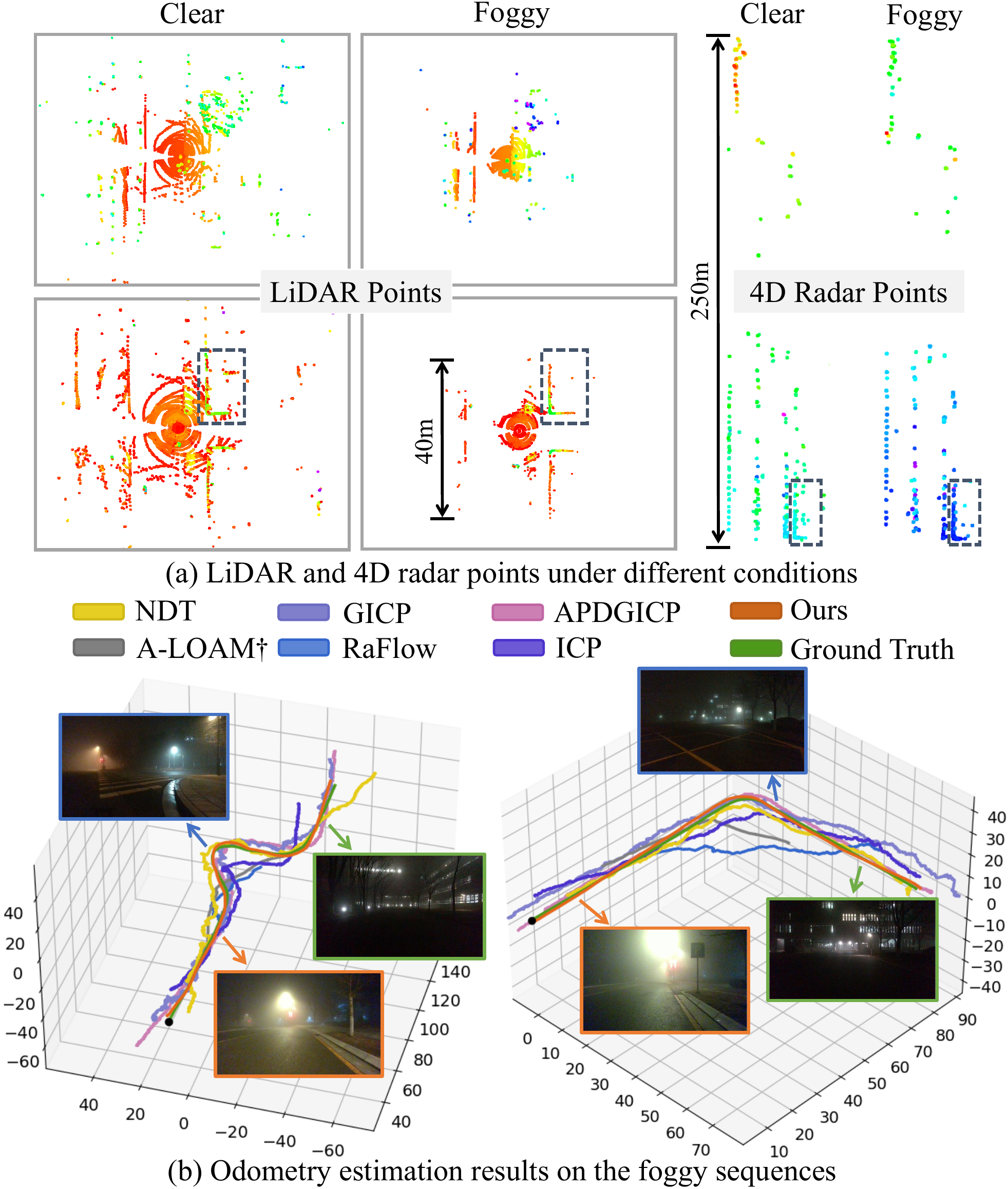}
\vspace{-0.28in}
\caption{Comparison of LiDAR and 4D radar sensing under clear and foggy conditions, and odometry estimation on foggy sequences.
In (a), fog degrades LiDAR point density, whereas the radar measurements stay relatively stable. To facilitate comparison of the detection range, the same region is highlighted.}
\vspace{-0.25in}
\label{fig:Trajectory_Fog}
\end{figure} 

\subsubsection{Self-supervised Strategy Selection} We show the impact of various self-supervised strategies in Tab.~\ref{tab:more_ablation_odm}. The 1$^{st}$ row shows that assigning higher weights to larger clusters is effective in $\mathcal{L}_{cd}$. 
Polar coordinates exhibit superiority over cartesian coordinates when transforming points into BEV for column comparison in $\mathcal{L}_{co}$. 
Besides, without a selection mechanism in $\mathcal{L}_{tg}$, suboptimal soft labels would misguide the model, causing performance degradation.
Using the farthest point rather than the neighbor with similar features as a negative sample in $\mathcal{L}_{fc}$ is more reasonable, as demonstrated in the 4$^{th}$ row.
Finally, the constant velocity assumption adversely affects the network learning, due to frequent speed variations in real-world driving.

\begin{table}[t!]
\renewcommand\tabcolsep{1.2pt}
\renewcommand{\arraystretch}{1.0}
\caption{4D radar odometry results on the foggy sequence of the self-collected campus dataset. ``L" means LiDAR points, and ``R" denotes 4D radar points.}
\vspace{-0.12in}
\begin{center}
\begin{tabular}{c|c|ccc|ccc}
\toprule[.045cm]
\multirow{2}{*}{Method} & \multirow{2}{*}{Input} & \multicolumn{3}{c|}{Fog0 (154$m$)} & \multicolumn{3}{c}{Fog1 (202$m$)} \\ \cline{3-8} 
& & RPE ($m$) & RPE (°) & ATE ($m$) & RPE ($m$) & RPE (°) & ATE ($m$) \\ \hline \hline
A-LOAM\dag & L & \underline{0.165} & 0.663 & 20.653 & \underline{0.180} & 1.113 & 29.384  \\ \hline
ICP & R & 0.217 & \underline{0.495} & 6.714 & {0.246} & \underline{0.627} & 13.075  \\
NDT & R & 0.224 & {0.566} & 6.565 & 0.263 & 0.709 & \underline{7.942}  \\
GICP & R & 0.352 & 0.629 & 8.252 & 0.372 & 0.749 & 10.362 \\
APDGICP & R & 0.230 & 0.641 & \underline{2.644} & 0.300 & 0.824 & 11.950  \\ \hline 
RaFlow & R & 0.317 & 0.669 & 23.747 & 0.297 & 0.937 & 38.178  \\
\textbf{Ours} & R & \textbf{0.146} & \textbf{0.352} & \textbf{1.378} & \textbf{0.079} & \textbf{0.403} & \textbf{2.592}  \\
\toprule[.045cm]
\end{tabular}
\end{center} 
\vspace{-0.3in}
\label{table:FogNEU}
\end{table}

\subsubsection{Gaussian Update and Regularization} 
\label{sec:Gaussian Update and Regularization}
We conduct ablation studies on the NTU4DRadLM dataset with various scenes to evaluate map optimization. As illustrated in Tab.~\ref{tab:ablation_gaus}, ground completion improves geometric quality because the radar map lacks ground structure initially. The improvement in ground rendering is also evident in 3$^{rd}$ and 4$^{th}$ rows of Fig.~\ref{fig:render_image}.
Since neighborhood-based (NB) pruning preserves useful gaussians for splitting and geometry-aware (GA) densification produces more gaussians in structurally appropriate positions to complete voids, all metrics further achieve a consistent boost.
The results in the 5$^{th}$ row of Tab.~\ref{tab:ablation_gaus} exhibit that the sky decoupling strategy improves map quality by using a sky mask to remove floaters. A qualitative example of this enhancement is shown in Fig.~\ref{fig:sky}.
The 6$^{th}$ row of Tab.~\ref{tab:ablation_gaus} and Fig.~\ref{fig:normal_image} demonstrate that geometric knowledge from visual foundation models serves as effective constraints. 
Finally, we mitigate the spatial ambiguity of a single view through multi-view collaborative constraints in overlap regions, resulting in the best performance.
In Tab.~\ref{tab:view_number}, we further explore the effect of view count. The results prove that using constraints from more perspectives facilitates proper convergence of gaussian attributes, leading to better results.

\subsection{More Experiments on Self-collected Campus Dataset}\label{sec:our datasets}
To validate our method’s effectiveness in real-world scenarios beyond public datasets, we conduct more experiments on a self-collected dataset captured using a handheld platform and a mobile robot under different weather and lighting conditions, including daytime, nighttime, and foggy environments.

\subsubsection{4D Radar Odometry}
As illustrated in Tab.~\ref{table:NEU}, we first assess 4D radar odometry on the daytime sequences collected using the handheld platform.
Here, our method achieves mean translation and rotation RPEs of 0.521 and 2.074, respectively, showing a clear advantage over learning-based RaFlow (5.103 / 9.047).
We attribute this advantage primarily to our cluster-aware matching and multi-level self-supervision, which jointly enforce constraints on noise-tolerant geometric consistency of warped points, feature discrimination on the extracted features, and temporal smoothness for the predicted poses.
Furthermore, we visualize odometry trajectories in Fig.~\ref{fig:trac_neu}. Owing to unstable inter-frame registration caused by low-quality radar points, most methods show noticeable fluctuations or deviations from reference trajectories.
In contrast, benefiting from the learnable association enhanced with object information and appropriate supervision signals, our method produces smoother trajectories across most sequences with lower accumulated error.

\begin{table*}[t!]
\renewcommand\tabcolsep{2.2pt}
\renewcommand{\arraystretch}{1.0}
\caption{Geometric and rendering quality on the daytime sequence of the self-collected dataset under RGB-based map optimization.}
\vspace{-0.15in}
\begin{center}
\begin{tabular}{c|ccc|ccc|ccc|ccc|ccc}
\toprule[.05cm]
\multirow{2}{*}{Method} & \multicolumn{3}{c|}{Gym} & \multicolumn{3}{c|}{Building2} & \multicolumn{3}{c|}{Straight} & \multicolumn{3}{c|}{Parking2} & \multicolumn{3}{c}{Mean} \\ \cline{2-16} & F-Score↑ & CD↓   & MHD↓   & F-Score↑ & CD↓   & MHD↓   & F-Score↑ & CD↓   & MHD↓   & F-Score↑ & CD↓   & MHD↓ & F-Score↑ & CD↓   & MHD↓ \\ \hline \hline

3DGS & 0.267 & \underline{1.210} & {0.422} & 0.198 & 3.531 & {1.012} & {0.298} & {0.898} & \textbf{0.319} & {0.389} & \underline{0.938} & {0.274} & {0.288} & 1.644 & {0.507} \\

2DGS  & 0.118 & 1.669 & 0.959 & 0.088 & {1.835} & 1.262 & 0.144 & 1.467 & 0.692 & 0.181 & 1.370 &1.824  & 0.133 & {1.585} & 1.184 \\

GaussianPro & {0.276} & 5.973 & 0.673 & {0.230} & 5.733 & 1.164 & 0.297 & 2.290 & 0.366 & 0.349 & 4.902 & 0.678 & {0.288} & 4.724 & 0.720\\

DASH & 0.108 & 1.923 & 1.078 & 0.066 & 4.179 & 2.259 & 0.180 & 1.089 & 0.604 & 0.315 & 1.011 & 0.229 & 0.167 & 2.051 & 1.043 \\

3D-HGS & 0.287 & 1.240 & 0.402 & 0.216 & \underline{2.306} & 0.848 & \underline{0.329} & \underline{0.878} & 0.337 & \underline{0.410} & 0.945 & \underline{0.218} & \underline{0.306} & \underline{1.342} & 0.451 \\

4DScaffoldGS & \underline{0.291} & 1.713 & \underline{0.392} & \underline{0.252} & 2.493 & \underline{0.704} & 0.291 & 1.812 & 0.333 & 0.379 & 1.709 & 0.285 & 0.303 & 1.932 & \underline{0.429} \\

\textbf{Ours} & \textbf{0.341} & \textbf{0.974} & \textbf{0.377} & \textbf{0.283} & \textbf{1.577} & \textbf{0.586} & \textbf{0.362} & \textbf{0.818} & \underline{0.328} & \textbf{0.484} & \textbf{0.854} & \textbf{0.173} & \textbf{0.367} & \textbf{1.056} & \textbf{0.366} \\ \hline\hline

\textbf{} & PSNR↑ & SSIM↑ & LPIPS↓ & PSNR↑ & SSIM↑ & LPIPS↓ & PSNR↑ & SSIM↑ & LPIPS↓ & PSNR↑ & SSIM↑ & LPIPS↓ & PSNR↑ & SSIM↑ & LPIPS↓ \\ \hline

3DGS & 18.467 & 0.557 & 0.516 & 18.911 & 0.640 & 0.474 & 19.131 & {0.509} & 0.547 & {21.141} & {0.705} & 0.383 & 19.412 & 0.602 & 0.480 \\

2DGS & 18.535 & 0.566 & 0.548 & 17.127 & 0.606 & 0.532 & 18.695 & 0.491 & 0.600 & 19.093 & 0.647 & 0.487 & 18.362 & 0.577 & 0.542 \\

GaussianPro & \underline{20.081} & {0.584} & \underline{0.459} & \underline{20.970} & \underline{0.671} & {0.437} & {19.461} & 0.498 & \textbf{0.512} & 21.122 & 0.683 & \textbf{0.346} & {20.408} & {0.609} & \textbf{0.439} \\

DASH & 19.749 & 0.582 & 0.528 & 20.105 & 0.650 & 0.503 & \underline{19.872} & \underline{0.510} & 0.562 & 21.144 & 0.691 & 0.430 & 20.218 & 0.608 & 0.506 \\

3D-HGS &  19.149 & 0.576 & 0.507 & 19.218 & 0.643 & 0.466 & 19.769 & 0.508 & 0.552 & \underline{22.058} & 0.708 & 0.378  & 20.049 & 0.609 & 0.475\\

4DScaffoldGS & 19.877 & \underline{0.586} & \textbf{0.458} & 20.594 & 0.670 & \underline{0.434} & 19.350 & 0.497 & 0.522 & 21.851 & \underline{0.711} & \underline{0.360} & \underline{20.418} & \underline{0.616} & 0.444 \\

\textbf{Ours} & \textbf{21.273} & \textbf{0.602} & {0.468} & \textbf{23.177} & \textbf{0.699} & \textbf{0.408} & \textbf{20.756} & \textbf{0.524} & \underline{0.519} & \textbf{22.853} & \textbf{0.724} & {0.366} & \textbf{22.015} & \textbf{0.637} & \underline{0.440} \\ \hline
\toprule[.035cm]
\end{tabular}
\end{center} 
\vspace{-0.25in}
\label{table:GSNEU}
\end{table*}

\begin{figure*}[t!]
\centering
\includegraphics[width=\linewidth]{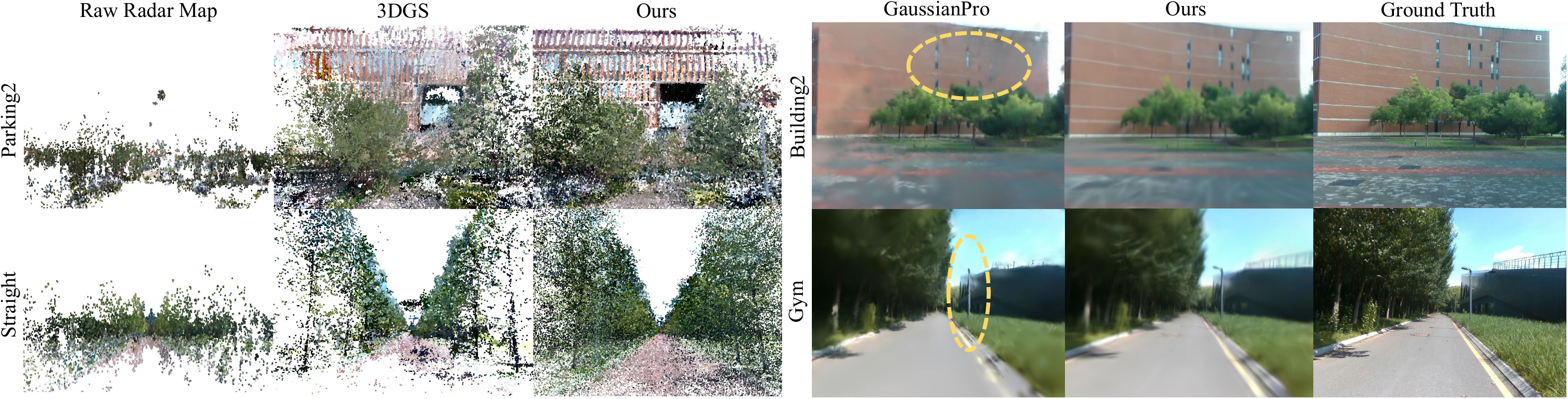}
\vspace{-0.3in}
\caption{Comparison of optimized radar maps and rendered images on the daytime sequence. Yellow circles mark the defective rendering areas in GaussianPro.}
\vspace{-0.15in}
\label{fig:reconstruction_neu}
\end{figure*}

To validate the advantage of radar in challenging conditions, we also conduct experiments on the foggy sequences collected by the mobile robot. 
As shown at the top of Fig.~\ref{fig:Trajectory_Fog}, the LiDAR points are degraded to some extent, with limited valid returns, whereas the 4D radar still provides stable measurements with a longer detection range.
As a result, A-LOAM\dag \ with LiDAR scan-to-scan matching exhibits inferior performance compared with the methods based on 4D radar, as illustrated in Tab.~\ref{table:FogNEU}.
For our method, we directly use the model trained on daytime sequences for evaluation under different platforms and weather conditions, without any fine-tuning. 
Even so, it still achieves a lower ATE of 1.985$m$ than learning-based RaFlow (30.963$m$), LiDAR-based A-LOAM (25.019$m$), and radar-based APDGICP (7.297$m$). The trajectory plots in Fig.~\ref{fig:Trajectory_Fog} also indicate that our method is closer to the ground truth than the others.

\subsubsection{Map Optimization with RGB images}
\label{sec:Map Optimization with RGB images}
Tab.~\ref{table:GSNEU} presents the geometric and novel-view rendering results on the daytime sequences under RGB-based optimization.
Compared with 3D-HGS and 4DScaffoldGS, our approach lowers the average CD from 1.342 and 1.932 to 1.056, respectively, while achieving a higher F-score.
These results demonstrate the effectiveness of our method in refining sparse radar maps into more complete scene representations with relatively accurate structures.
Then, such improved geometric quality is also accompanied by better image rendering performance. Specifically, our method obtains mean PSNR and SSIM values of 22.015 and 0.637, surpassing 4DScaffoldGS at 20.418 and 0.616.
We think this advantage mainly stems from the reasonable spatial distribution of Gaussian ellipsoids, which could better support the rendering of fine texture details.

\begin{figure*}[t!]
\centering
\includegraphics[width=\linewidth]{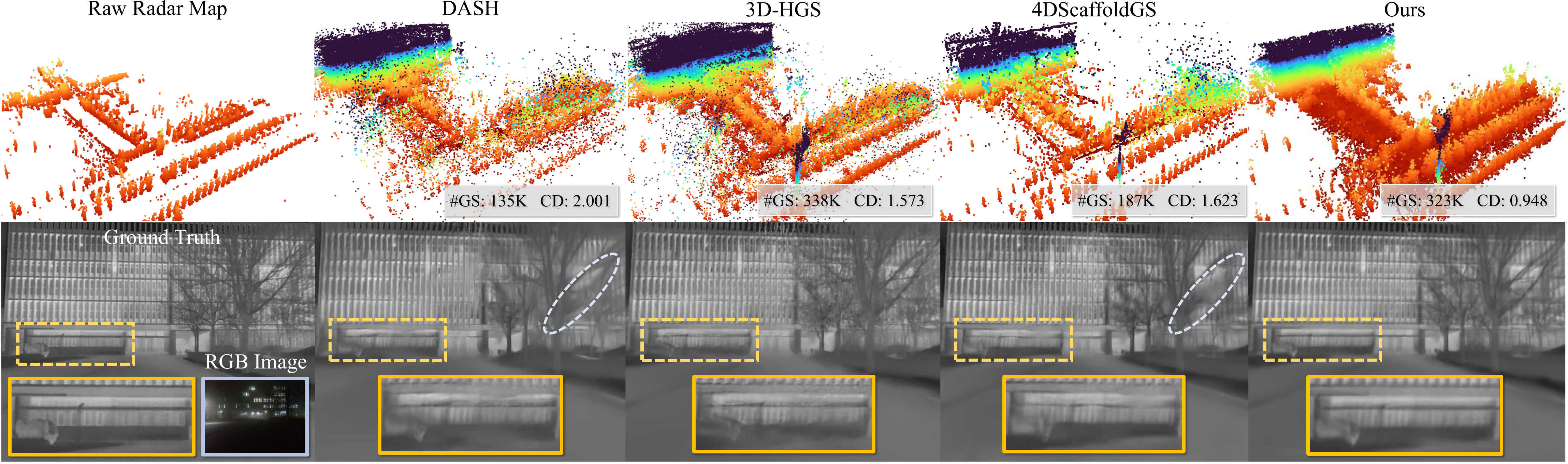}
\vspace{-0.3in}
\caption{Comparison of thermal-based map optimization on the Fog0 sequence of the self-collected dataset, along with the corresponding rendered images.}
\vspace{-0.15in}
\label{fig:thermal_mapping_neu}
\end{figure*} 

\begin{figure*}[t!]
\centering
\includegraphics[width=\linewidth]{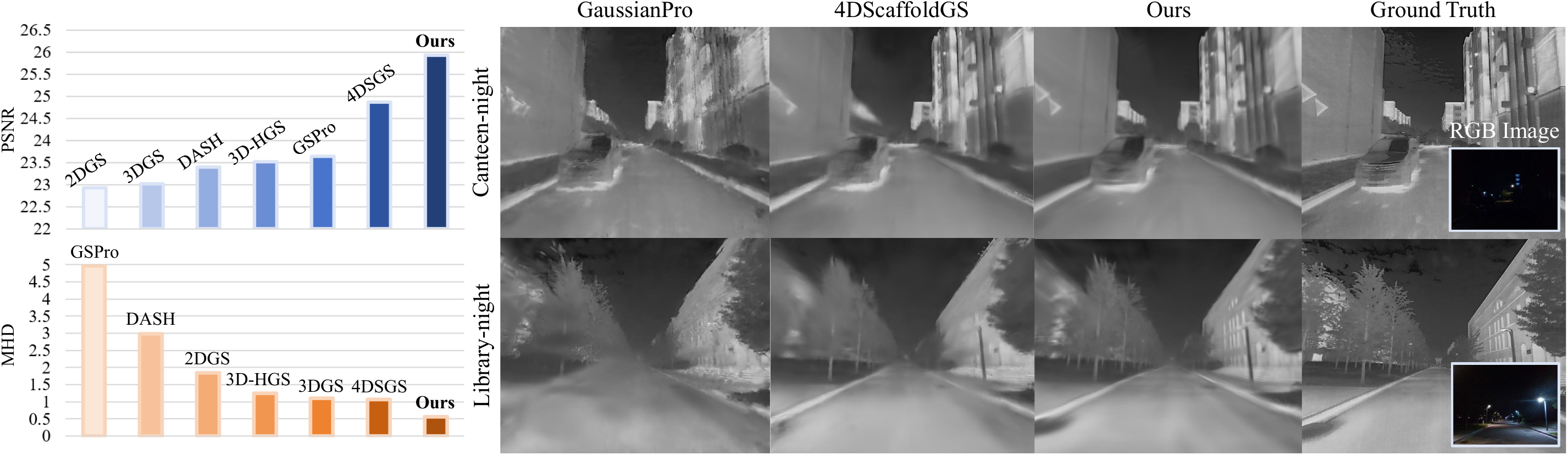}
\vspace{-0.3in}
\caption{Comparison of thermal-based optimization results on the nighttime sequences of the self-collected dataset. Left: quantitative results in terms of PSNR and MHD. Right: qualitative comparison of rendered thermal images, with paired RGB images shown for reference.}
\vspace{-0.2in}
\label{fig:thermal_render_neu}
\end{figure*} 

\begin{table}[t!]
\centering
\setlength{\tabcolsep}{2.0pt}
\renewcommand{\arraystretch}{1.0}
\caption{Reconstruction quality on the foggy sequences of the self-collected dataset using RGB and thermal images. ``T" means thermal image.}
\vspace{-0.05in}
\begin{tabular}{c|c|ccc|ccc}
\toprule[.05cm]
\multirow{2}{*}{Method} & \multirow{2}{*}{Mod.} & \multicolumn{3}{c|}{Fog0} & \multicolumn{3}{c}{Fog1} \\ \cline{3-8} 
& & F-Score↑ & CD↓ & MHD↓ & F-Score↑ & CD↓ & MHD↓ \\ \hline \hline

3D-HGS & RGB & 0.272 & 1.632 & 0.671 & 0.199 & 2.089 & 0.587  \\

Ours & RGB & 0.326 &1.093 & 0.422 & 0.323 & 1.504 & 0.341  \\ \hline

3DGS & T & 0.305 & \underline{1.469} & 0.379 & 0.297 & 2.120 & 0.326  \\

2DGS & T & 0.119 & 2.308 & 1.270 & 0.107 & 1.853 & 1.193 \\

GaussianPro & T & 0.245 & 5.826 & 5.241 & 0.169 & 6.285 & 9.826  \\

DASH & T & 0.149 & 2.001 & 1.074 & 0.102 & 2.299 & 0.976  \\

3D-HGS & T & \underline{0.328} & 1.573 & \underline{0.365} & \underline{0.326} & \underline{1.431} & \underline{0.286}  \\

4DScaffoldGS & T & 0.249 & 1.623 & 0.809 & 0.224 & 1.654 & 0.649  \\

\textbf{Ours} & T & \textbf{0.381} & \textbf{0.948} & \textbf{0.206} & \textbf{0.360} & \textbf{1.218} & \textbf{0.250}  \\ \hline \hline

\textbf{} & \textbf{} & PSNR↑ & SSIM↑ & LPIPS↓ & PSNR↑ & SSIM↑ & LPIPS↓ \\ \hline

3D-HGS & RGB & 25.456 &0.691 & 0.494 & 23.409 & 0.666 & 0.513  \\

Ours & RGB & 27.941 & 0.699 & 0.488 & 27.233 & 0.697 & 0.492  \\ \hline

3DGS & T & 29.762 & 0.906 & 0.241 & 27.278 & 0.857 & \underline{0.281}  \\

2DGS & T & 27.841 & 0.886 & 0.282 & 26.338 & 0.856 & 0.326   \\

GaussianPro & T & 28.789 & 0.884 & 0.259 & 27.216 & 0.859 & 0.286   \\

DASH & T & 29.157 & 0.896 & 0.270 & 27.365 & 0.861 & 0.309  \\

3D-HGS & T & 30.287 & \underline{0.909} & \underline{0.239} & 27.616 & \underline{0.870} & 0.286  \\

4DScaffoldGS & T & \underline{30.432} & 0.903 & 0.243 & \underline{28.514} & 0.869 & 0.284  \\

\textbf{Ours} & T & \textbf{31.599} & \textbf{0.911} & \textbf{0.231} & \textbf{29.828} & \textbf{0.878} & \textbf{0.267} \\

\toprule[.05cm]
\end{tabular}
\vspace{-0.2in}
\label{tab:thermal_fog_neu}
\end{table}

For a more intuitive comparison, we visualize the optimized local radar maps and rendered images in Fig.~\ref{fig:reconstruction_neu}.
Compared with 3DGS, our algorithm produces denser and more coherent structures. 
We speculate that this arises from our depth-assisted completion, which introduces synthetic gaussians using depth priors to compensate for fragmented ground, while geometry-aware densification resplits and interpolates gaussians based on local geometry to strengthen structural continuity and integrity in regions lacking radar returns.
Moreover, the multi-view regularization further exploits overlapping regions across adjacent views to suppress single-view ambiguity and guide gaussians toward more accurate states.
With better scene representations, our approach can also recover clearer appearance details in the rendered images, such as building windows and light poles, as highlighted on the right side of Fig.~\ref{fig:reconstruction_neu}.

\subsubsection{Radar-Thermal Reconstruction} 
As shown in Tab.~\ref{tab:thermal_fog_neu}, we compare map optimization results on the foggy sequences across different visual modalities. 
To evaluate geometric accuracy, we collect LiDAR data under normal weather conditions and construct reference maps aligned with the raw radar maps.
The results demonstrate that our method still achieves superior results, and also indicate that thermal-based optimization leads to better structural quality than using RGB images in degraded conditions.
For example, under our framework, replacing RGB with thermal images improves the F-score from 0.326 to 0.381 and decreases the MHD from 0.422 to 0.206 on \textit{Fog0}, with similar gains also observed for 3D-HGS.
This is likely because RGB images suffer from reduced visibility and loss of scene details, which weakens the appearance cues available for map optimization and limits the effective splitting of gaussians for recovering missing structures.
For image synthesis quality, our method also outperforms 4DScaffoldGS by 1.241 dB in mean PSNR for thermal-based rendering and 3D-HGS by 3.155 dB in the RGB-based setting, showing its effectiveness in not only geometric accuracy but also rendering fidelity.

We also exhibit the reconstructed map in Fig.~\ref{fig:thermal_mapping_neu}. 
Compared with other methods, we incorporate depth-assisted completion to recover ground regions and geometry-aware densification to strengthen structural details, resulting in a more complete and coherent reconstruction with 323K gaussians and a lower CD of 0.948.
Moreover, the resulting gaussian representation with a more reasonable spatial distribution further contributes to the improved local detail rendering, as displayed at the bottom row of Fig.~\ref{fig:thermal_mapping_neu}.
These advantages are also reflected in the thermal-based optimization results on nighttime sequences in Fig.~\ref{fig:thermal_render_neu}. 
The left side presents the average PSNR and MHD on the two sequences, with our method achieving notable improvements, while the qualitative visualization on the right shows rendered images with clearer details.
Overall, combined with the RGB-based results in Sec.~\ref{sec:Map Optimization with RGB images}, these findings 
demonstrate that our approach remains consistently superior to other algorithms for map optimization across different visual inputs.

\subsection{Limitation and Future Direction}\label{sec:limitation}
Although our framework employs self-supervised 4D radar odometry, learned without labeled data, to estimate continuous poses and construct raw radar maps, and further enhances map structural completeness through gaussian-based optimization, inevitable odometry drift may still impact the consistency and overall quality of the optimized radar map.
Therefore, in future work, we will extend Super4DR with radar-based loop closure detection and pose-graph optimization to correct large-scale accumulated drift. 
Subsequently, joint refinement of trajectory states and gaussian map parameters will leverage differentiable map-based feedback to reduce small residual pose errors and improve the global consistency of the final reconstruction.

\section{Conclusions}
In this article, we introduce the first 4D radar-based framework, called \textit{Super4DR}, which contains learning-based odometry estimation and map optimization while explicitly accounting for radar-specific characteristics.
It can predict poses based on noisy and sparse radar points, reconstruct dense structures from low-quality radar maps, and render multi-modal images with depth and normal maps.
\textit{For odometry estimation}, we first propose a cluster-aware network, which leverages object-level cues to perform robust inter-frame matching and is trained by multi-level self-supervised losses for label-free learning.
Under consideration of radar point distribution, cluster-weighted distance and column occupancy losses are adopted to constrain geometric consistency. 
A teacher guidance loss then uses soft labels as additional supervision to perform knowledge transfer.
The feature contrast and temporal constraint are also applied to enable effective matching and improve trajectory smoothness.
\textit{For map optimization}, we propose a gaussian-based optimizer that views 3D gaussians as an intermediate map representation. 
It contains a depth-guided completion module for the ground recovery, combined with a radar-specific densification module aimed at improving map integrity. The selective separation and multi-view regularization are further used to decrease floaters, avoid mistaken deletion and accelerate optimization.
Extensive experiments on public and self-collected datasets show that our method achieves superior results in various aspects, including odometry accuracy, map geometric quality and rendering detail fidelity. We further explore the potential of 4D radar-thermal reconstruction in poor illumination and verify the applicability of our method across different visual sensors.

\bibliographystyle{ieeetr}
\bibliography{ref}

\begin{thebibliography}{10}

\bibitem{4dradarslam}
J.~Zhang, H.~Zhuge, Z.~Wu, G.~Peng, M.~Wen, Y.~Liu, and D.~Wang, ``4dradarslam: A 4d imaging radar slam system for large-scale environments based on pose graph optimization,'' in {\em 2023 IEEE International Conference on Robotics and Automation (ICRA)}, pp.~8333--8340, IEEE, 2023.

\bibitem{EFEAR-4D}
X.~Wu, Y.~Chen, Z.~Li, Z.~Hong, and L.~Hu, ``Efear-4d: Ego-velocity filtering for efficient and accurate 4d radar odometry,'' {\em IEEE Robotics and Automation Letters}, vol.~9, no.~11, pp.~9828--9835, 2024.

\bibitem{Radar-ICP}
D.~C. Herraez, M.~Zeller, L.~Chang, I.~Vizzo, M.~Heidingsfeld, and C.~Stachniss, ``Radar-only odometry and mapping for autonomous vehicles,'' in {\em 2024 IEEE International Conference on Robotics and Automation (ICRA)}, pp.~10275--10282, 2024.

\bibitem{Radar4Motion}
S.~Kim, J.~Seok, J.~Lee, and K.~Jo, ``Radar4motion: Imu-free 4d radar odometry with robust dynamic filtering and rcs-weighted matching,'' {\em IEEE Transactions on Intelligent Vehicles}, pp.~1--11, 2024.

\bibitem{4DRONet}
S.~Lu, G.~Zhuo, L.~Xiong, X.~Zhu, L.~Zheng, Z.~He, M.~Zhou, X.~Lu, and J.~Bai, ``Efficient deep-learning 4d automotive radar odometry method,'' {\em IEEE Transactions on Intelligent Vehicles}, 2023.

\bibitem{4DRVONet}
G.~Zhuoins, S.~Lu, L.~Xiong, H.~Zhouins, L.~Zheng, and M.~Zhou, ``4drvo-net: Deep 4d radar--visual odometry using multi-modal and multi-scale adaptive fusion,'' {\em IEEE Transactions on Intelligent Vehicles}, 2023.

\bibitem{SelfRONet}
H.~Zhou, S.~Lu, and G.~Zhuo, ``Self-supervised 4-d radar odometry for autonomous vehicles,'' in {\em 2023 IEEE 26th International Conference on Intelligent Transportation Systems (ITSC)}, pp.~764--769, 2023.

\bibitem{RaFlow}
F.~Ding, Z.~Pan, Y.~Deng, J.~Deng, and C.~X. Lu, ``Self-supervised scene flow estimation with 4-d automotive radar,'' {\em IEEE Robotics and Automation Letters}, vol.~7, no.~3, pp.~8233--8240, 2022.

\bibitem{ntu4dradlm}
J.~Zhang, H.~Zhuge, Y.~Liu, G.~Peng, Z.~Wu, H.~Zhang, Q.~Lyu, H.~Li, C.~Zhao, D.~Kircali, {\em et~al.}, ``Ntu4dradlm: 4d radar-centric multi-modal dataset for localization and mapping,'' in {\em 2023 IEEE 26th International Conference on Intelligent Transportation Systems (ITSC)}, pp.~4291--4296, IEEE, 2023.

\bibitem{RDiffusion}
R.~Zhang, D.~Xue, Y.~Wang, R.~Geng, and F.~Gao, ``Towards dense and accurate radar perception via efficient cross-modal diffusion model,'' {\em IEEE Robotics and Automation Letters}, vol.~9, no.~9, pp.~7429--7436, 2024.

\bibitem{RDiff}
K.~Luan, C.~Shi, N.~Wang, Y.~Cheng, H.~Lu, and X.~Chen, ``Diffusion-based point cloud super-resolution for mmwave radar data,'' in {\em 2024 IEEE International Conference on Robotics and Automation (ICRA)}, pp.~11171--11177, 2024.

\bibitem{3dgs}
B.~Kerbl, G.~Kopanas, T.~Leimk{\"u}hler, and G.~Drettakis, ``3d gaussian splatting for real-time radiance field rendering.,'' {\em ACM Trans. Graph.}, vol.~42, no.~4, pp.~139--1, 2023.

\bibitem{P2P-ICP}
P.~J. Besl and N.~D. McKay, ``Method for registration of 3-d shapes,'' in {\em Sensor fusion IV: control paradigms and data structures}, vol.~1611, pp.~586--606, Spie, 1992.

\bibitem{GICP}
A.~Segal, D.~Haehnel, and S.~Thrun, ``Generalized-icp.,'' in {\em Robotics: science and systems}, vol.~2, p.~435, Seattle, WA, 2009.

\bibitem{Kiss-icp}
I.~Vizzo, T.~Guadagnino, B.~Mersch, L.~Wiesmann, J.~Behley, and C.~Stachniss, ``Kiss-icp: In defense of point-to-point icp--simple, accurate, and robust registration if done the right way,'' {\em IEEE Robotics and Automation Letters}, vol.~8, no.~2, pp.~1029--1036, 2023.

\bibitem{NDT}
P.~Biber and W.~Stra{\ss}er, ``The normal distributions transform: A new approach to laser scan matching,'' in {\em Proceedings 2003 IEEE/RSJ International Conference on Intelligent Robots and Systems (IROS 2003)(Cat. No. 03CH37453)}, vol.~3, pp.~2743--2748, IEEE, 2003.

\bibitem{loam}
J.~Zhang, S.~Singh, {\em et~al.}, ``Loam: Lidar odometry and mapping in real-time.,'' in {\em Robotics: Science and systems}, vol.~2, pp.~1--9, Berkeley, CA, 2014.

\bibitem{lego-loam}
T.~Shan and B.~Englot, ``Lego-loam: Lightweight and ground-optimized lidar odometry and mapping on variable terrain,'' in {\em 2018 IEEE/RSJ International Conference on Intelligent Robots and Systems (IROS)}, pp.~4758--4765, IEEE, 2018.

\bibitem{F-loam}
H.~Wang, C.~Wang, C.-L. Chen, and L.~Xie, ``F-loam: Fast lidar odometry and mapping,'' in {\em 2021 IEEE/RSJ International Conference on Intelligent Robots and Systems (IROS)}, pp.~4390--4396, IEEE, 2021.

\bibitem{Pwclo}
G.~Wang, X.~Wu, Z.~Liu, and H.~Wang, ``Pwclo-net: Deep lidar odometry in 3d point clouds using hierarchical embedding mask optimization,'' in {\em Proceedings of the IEEE/CVF conference on computer vision and pattern recognition}, pp.~15910--15919, 2021.

\bibitem{EfficientLO}
G.~Wang, X.~Wu, S.~Jiang, Z.~Liu, and H.~Wang, ``Efficient 3d deep lidar odometry,'' {\em IEEE Transactions on Pattern Analysis and Machine Intelligence}, vol.~45, no.~5, pp.~5749--5765, 2023.

\bibitem{Translo}
J.~Liu, G.~Wang, C.~Jiang, Z.~Liu, and H.~Wang, ``Translo: A window-based masked point transformer framework for large-scale lidar odometry,'' in {\em Proceedings of the AAAI Conference on Artificial Intelligence}, vol.~37, pp.~1683--1691, 2023.

\bibitem{DSLO}
H.~Zhang, G.~Wang, X.~Wu, C.~Xu, M.~Ding, M.~Tomizuka, W.~Zhan, and H.~Wang, ``Dslo: Deep sequence lidar odometry based on inconsistent spatio-temporal propagation,'' in {\em 2024 IEEE/RSJ International Conference on Intelligent Robots and Systems (IROS)}, pp.~10672--10677, 2024.

\bibitem{DeLORA}
J.~Nubert, S.~Khattak, and M.~Hutter, ``Self-supervised learning of lidar odometry for robotic applications,'' in {\em 2021 IEEE International Conference on Robotics and Automation (ICRA)}, pp.~9601--9607, 2021.

\bibitem{HPPLO-Net}
B.~Zhou, Y.~Tu, Z.~Jin, C.~Xu, and H.~Kong, ``Hpplo-net: Unsupervised lidar odometry using a hierarchical point-to-plane solver,'' {\em IEEE Transactions on Intelligent Vehicles}, vol.~9, no.~1, pp.~2727--2739, 2024.

\bibitem{RSLO}
Y.~Xu, J.~Lin, J.~Shi, G.~Zhang, X.~Wang, and H.~Li, ``Robust self-supervised lidar odometry via representative structure discovery and 3d inherent error modeling,'' {\em IEEE Robotics and Automation Letters}, vol.~7, no.~2, pp.~1651--1658, 2022.

\bibitem{CFEAR}
D.~Adolfsson, M.~Magnusson, A.~Alhashimi, A.~J. Lilienthal, and H.~Andreasson, ``Cfear radarodometry - conservative filtering for efficient and accurate radar odometry,'' in {\em 2021 IEEE/RSJ International Conference on Intelligent Robots and Systems (IROS)}, pp.~5462--5469, 2021.

\bibitem{ORORA}
H.~Lim, K.~Han, G.~Shin, G.~Kim, S.~Hong, and H.~Myung, ``Orora: Outlier-robust radar odometry,'' in {\em 2023 IEEE International Conference on Robotics and Automation (ICRA)}, pp.~2046--2053, 2023.

\bibitem{SDRP}
R.~Zhang, Y.~Zhang, D.~Fu, and K.~Liu, ``Scan denoising and normal distribution transform for accurate radar odometry and positioning,'' {\em IEEE Robotics and Automation Letters}, vol.~8, no.~3, pp.~1199--1206, 2023.

\bibitem{CMFlow}
F.~Ding, A.~Palffy, D.~M. Gavrila, and C.~X. Lu, ``Hidden gems: 4d radar scene flow learning using cross-modal supervision,'' in {\em Proceedings of the IEEE/CVF Conference on Computer Vision and Pattern Recognition}, pp.~9340--9349, 2023.

\bibitem{CAO-RONet}
Z.~Li, Y.~Cui, N.~Huang, C.~Pang, and Z.~Fang, ``Cao-ronet: A robust 4d radar odometry with exploring more information from low-quality points,'' in {\em 2025 IEEE International Conference on Robotics and Automation (ICRA)}, pp.~13014--13020, IEEE, 2025.

\bibitem{Nerf}
B.~Mildenhall, P.~P. Srinivasan, M.~Tancik, J.~T. Barron, R.~Ramamoorthi, and R.~Ng, ``Nerf: Representing scenes as neural radiance fields for view synthesis,'' {\em Communications of the ACM}, vol.~65, no.~1, pp.~99--106, 2021.

\bibitem{Gaussianpro}
K.~Cheng, X.~Long, K.~Yang, Y.~Yao, W.~Yin, Y.~Ma, W.~Wang, and X.~Chen, ``Gaussianpro: 3d gaussian splatting with progressive propagation,'' in {\em Proceedings of the 41st International Conference on Machine Learning}, pp.~8123--8140, 2024.

\bibitem{Dngaussian}
J.~Li, J.~Zhang, X.~Bai, J.~Zheng, X.~Ning, J.~Zhou, and L.~Gu, ``Dngaussian: Optimizing sparse-view 3d gaussian radiance fields with global-local depth normalization,'' in {\em Proceedings of the IEEE/CVF conference on computer vision and pattern recognition}, pp.~20775--20785, 2024.

\bibitem{2dgs}
B.~Huang, Z.~Yu, A.~Chen, A.~Geiger, and S.~Gao, ``2d gaussian splatting for geometrically accurate radiance fields,'' in {\em ACM SIGGRAPH 2024 conference papers}, pp.~1--11, 2024.

\bibitem{LiV-GS}
R.~Xiao, W.~Liu, Y.~Chen, and L.~Hu, ``Liv-gs: Lidar-vision integration for 3d gaussian splatting slam in outdoor environments,'' {\em IEEE Robotics and Automation Letters}, vol.~10, no.~1, pp.~421--428, 2025.

\bibitem{HGS-Mapping}
K.~Wu, K.~Zhang, Z.~Zhang, M.~Tie, S.~Yuan, J.~Zhao, Z.~Gan, and W.~Ding, ``Hgs-mapping: Online dense mapping using hybrid gaussian representation in urban scenes,'' {\em IEEE Robotics and Automation Letters}, vol.~9, no.~11, pp.~9573--9580, 2024.

\bibitem{Drivinggaussian}
X.~Zhou, Z.~Lin, X.~Shan, Y.~Wang, D.~Sun, and M.-H. Yang, ``Drivinggaussian: Composite gaussian splatting for surrounding dynamic autonomous driving scenes,'' in {\em Proceedings of the IEEE/CVF conference on computer vision and pattern recognition}, pp.~21634--21643, 2024.

\bibitem{kung2025radarsplat}
P.-C. Kung, S.~Harisha, R.~Vasudevan, A.~Eid, and K.~A. Skinner, ``Radarsplat: Radar gaussian splatting for high-fidelity data synthesis and 3d reconstruction of autonomous driving scenes,'' in {\em Proceedings of the IEEE/CVF International Conference on Computer Vision}, pp.~27596--27606, 2025.

\bibitem{DBSCAN}
M.~Ester, H.-P. Kriegel, J.~Sander, and X.~Xu, ``Density-based spatial clustering of applications with noise,'' in {\em Int. Conf. knowledge discovery and data mining}, vol.~240, 1996.

\bibitem{Flownet3d}
X.~Liu, C.~R. Qi, and L.~J. Guibas, ``Flownet3d: Learning scene flow in 3d point clouds,'' in {\em Proceedings of the IEEE/CVF conference on computer vision and pattern recognition}, pp.~529--537, 2019.

\bibitem{Pointpwc-net}
W.~Wu, Z.~Y. Wang, Z.~Li, W.~Liu, and L.~Fuxin, ``Pointpwc-net: Cost volume on point clouds for (self-) supervised scene flow estimation,'' in {\em Computer Vision--ECCV 2020: 16th European Conference, Glasgow, UK, August 23--28, 2020, Proceedings, Part V 16}, pp.~88--107, Springer, 2020.

\bibitem{gru}
K.~Cho, B.~Van~Merri{\"e}nboer, {\c{C}}.~Gul{\c{c}}ehre, D.~Bahdanau, F.~Bougares, H.~Schwenk, and Y.~Bengio, ``Learning phrase representations using rnn encoder--decoder for statistical machine translation,'' in {\em Proceedings of the 2014 conference on empirical methods in natural language processing (EMNLP)}, pp.~1724--1734, 2014.

\bibitem{HintonDistilling}
G.~Hinton, O.~Vinyals, and J.~Dean, ``Distilling the knowledge in a neural network,'' in {\em NIPS Deep Learning and Representation Learning Workshop}, 2015.

\bibitem{Apprentice}
A.~Mishra and D.~Marr, ``Apprentice: Using knowledge distillation techniques to improve low-precision network accuracy,'' in {\em International Conference on Learning Representations (ICLR)}, 2018.

\bibitem{Seed}
Z.~Fang, J.~Wang, L.~Wang, L.~Zhang, Y.~Yang, and Z.~Liu, ``Seed: Self-supervised distillation for visual representation,'' in {\em International Conference on Learning Representations (ICLR)}, 2021.

\bibitem{PDF-Flow}
P.~He, P.~Emami, S.~Ranka, and A.~Rangarajan, ``Self-supervised robust scene flow estimation via the alignment of probability density functions,'' in {\em Proceedings of the AAAI Conference on Artificial Intelligence}, vol.~36, pp.~861--869, 2022.

\bibitem{Depthanythingv2}
L.~Yang, B.~Kang, Z.~Huang, Z.~Zhao, X.~Xu, J.~Feng, and H.~Zhao, ``Depth anything v2,'' {\em Advances in Neural Information Processing Systems}, vol.~37, pp.~21875--21911, 2024.

\bibitem{Mask2Former}
B.~Cheng, I.~Misra, A.~G. Schwing, A.~Kirillov, and R.~Girdhar, ``Masked-attention mask transformer for universal image segmentation,'' in {\em Proceedings of the IEEE/CVF conference on computer vision and pattern recognition}, pp.~1290--1299, 2022.

\bibitem{Metric3D}
M.~Hu, W.~Yin, C.~Zhang, Z.~Cai, X.~Long, H.~Chen, K.~Wang, G.~Yu, C.~Shen, and S.~Shen, ``Metric3d v2: A versatile monocular geometric foundation model for zero-shot metric depth and surface normal estimation,'' {\em IEEE Transactions on Pattern Analysis and Machine Intelligence}, vol.~46, no.~12, pp.~10579--10596, 2024.

\bibitem{VoD}
A.~Palffy, E.~Pool, S.~Baratam, J.~F. Kooij, and D.~M. Gavrila, ``Multi-class road user detection with 3+ 1d radar in the view-of-delft dataset,'' {\em IEEE Robotics and Automation Letters}, vol.~7, no.~2, pp.~4961--4968, 2022.

\bibitem{radarcalib}
C.~Cao, X.~Wang, W.~Xi, H.~Zhang, W.~Chen, and J.~Wang, ``A 4d radar camera extrinsic calibration tool based on 3d uncertainty perspective n points,'' in {\em 2025 IEEE/RSJ International Conference on Intelligent Robots and Systems (IROS)}, pp.~8735--8740, IEEE, 2025.

\bibitem{opencalib}
G.~Yan, Z.~Liu, C.~Wang, C.~Shi, P.~Wei, X.~Cai, T.~Ma, Z.~Liu, Z.~Zhong, Y.~Liu, {\em et~al.}, ``Opencalib: A multi-sensor calibration toolbox for autonomous driving,'' {\em Software Impacts}, vol.~14, p.~100393, 2022.

\bibitem{Fast-lio}
W.~Xu and F.~Zhang, ``Fast-lio: A fast, robust lidar-inertial odometry package by tightly-coupled iterated kalman filter,'' {\em IEEE Robotics and Automation Letters}, vol.~6, no.~2, pp.~3317--3324, 2021.

\bibitem{Lo-net}
Q.~Li, S.~Chen, C.~Wang, X.~Li, C.~Wen, M.~Cheng, and J.~Li, ``Lo-net: Deep real-time lidar odometry,'' in {\em Proceedings of the IEEE/CVF Conference on Computer Vision and Pattern Recognition}, pp.~8473--8482, 2019.

\bibitem{DNOI-4DRO}
S.~Lu, H.~Zhou, G.~Zhuo, and X.~Tang, ``Dnoi-4dro: Deep 4d radar odometry with differentiable neural-optimization iterations,'' in {\em Proceedings of the AAAI Conference on Artificial Intelligence}, vol.~40, pp.~18487--18495, 2026.

\bibitem{3d-hgs}
H.~Li, J.~Liu, M.~Sznaier, and O.~Camps, ``3d-hgs: 3d half-gaussian splatting,'' in {\em Proceedings of the Computer Vision and Pattern Recognition Conference}, pp.~10996--11005, 2025.

\bibitem{DASH}
J.~Chen, Z.~Hu, P.~Wu, H.~Zhu, H.~Li, and X.~Sun, ``Dash: 4d hash encoding with self-supervised decomposition for real-time dynamic scene rendering,'' in {\em Proceedings of the IEEE/CVF International Conference on Computer Vision}, pp.~26349--26359, 2025.

\bibitem{4DScaffoldGS}
W.~O. Cho, I.~Cho, S.~Kim, J.~Bae, Y.~Uh, and S.~J. Kim, ``4d scaffold gaussian splatting with dynamic-aware anchor growing for efficient and high-fidelity dynamic scene reconstruction,'' in {\em Proceedings of the AAAI Conference on Artificial Intelligence}, vol.~40, pp.~3363--3371, 2026.

\end{thebibliography}

\end{document}